\documentclass{article}
\usepackage{ijcai24}

\usepackage{times}
\usepackage{soul}
\usepackage{url}
\usepackage[hidelinks]{hyperref}
\usepackage[utf8]{inputenc}
\usepackage[small]{caption}
\usepackage{graphicx}
\usepackage{amsmath}
\usepackage{amsthm}
\usepackage{booktabs}
\usepackage{algorithm}
\usepackage{algorithmic}
\usepackage[switch]{lineno}

\usepackage{multirow}
\usepackage{float}
\usepackage{colortbl}
\usepackage[table]{xcolor}
\usepackage{subfigure}
\usepackage{amssymb}
\usepackage{arydshln}
\numberwithin{table}{section}
\numberwithin{figure}{section}

\definecolor{mygreen}{RGB}{0, 128, 0}
\definecolor{cell1}{RGB}{244, 164, 96} % SandyBrown
\definecolor{cell2}{RGB}{135,206, 235} % Skyblue
\definecolor{cell3}{RGB}{198, 224, 180}

\title{Self-Supervised Monocular Depth Estimation in the Dark: \\
Towards Data Distribution Compensation}

\author{
Haolin Yang$^1$
\and
Chaoqiang Zhao$^1$\and
Lu Sheng$^2$\And
Yang Tang$^1$\\
\affiliations
$^1$East China University of Science and Technology\\
$^2$Beihang University\\
\emails
\{haolinyang, zhaocq\}@mail.ecust.edu.cn,
lsheng@buaa.edu.cn,
yangtang@ecust.edu.cn
}
\begin{document}

\maketitle

%%%%%%%%% ABSTRACT
\begin{abstract}
    Nighttime self-supervised monocular depth estimation has received increasing attention in recent years.
    However, using night images for self-supervision is unreliable because the photometric consistency assumption is usually violated in the videos taken under complex lighting conditions. 
    Even with domain adaptation or photometric loss repair, performance is still limited by the poor supervision of night images on trainable networks.
    In this paper, we propose a self-supervised nighttime monocular depth estimation method that does not use any night images during training.
    Our framework utilizes day images as a stable source for self-supervision and applies physical priors ~(e.g., wave optics, reflection model and read-shot noise model) to compensate for some key day-night differences.
    With day-to-night data distribution compensation, our framework can be trained in an efficient one-stage self-supervised manner.
    Though no nighttime images are considered during training, qualitative and quantitative results demonstrate that our method achieves SoTA depth estimating results on the challenging nuScenes-Night and RobotCar-Night compared with existing methods.
    
\end{abstract}
%%%%%%%%% BODY TEXT
\section{Introduction} \label{Introduction}
Monocular depth estimation plays a key role in several applications, 
such as augmented reality~\cite{AR2}, autonomous driving~\cite{obj_sf} and 
robotic manipulation~\cite{robotic_manipulation}.
With the usage and development of neural networks, like Convolutional Neural Network~\cite{resnet} and 
Vision Transformer~\cite{ViT}, deep-learning approaches present impressive results in this task~\cite{monocular_review_2020}.
Since the self-supervised framework does not 
require numerous costly depth-image pairs during training, it has achieved increasing attention in recent years.
Instead of using ground truth depth labels, the spatial and temporal geometric constraints from images are constructed in the self-supervised framework to supervise the training process. 
The photometric consistency assumption is used to build up the main constraint~(i.e., the photometric loss) of self-supervision. 
Results in \cite{monodepth2,monovit,packnet,hr-depth} show the effectiveness of classic self-supervised training on day scenes.

\begin{figure}[t]
    \centering
    \includegraphics[width=0.95\linewidth]{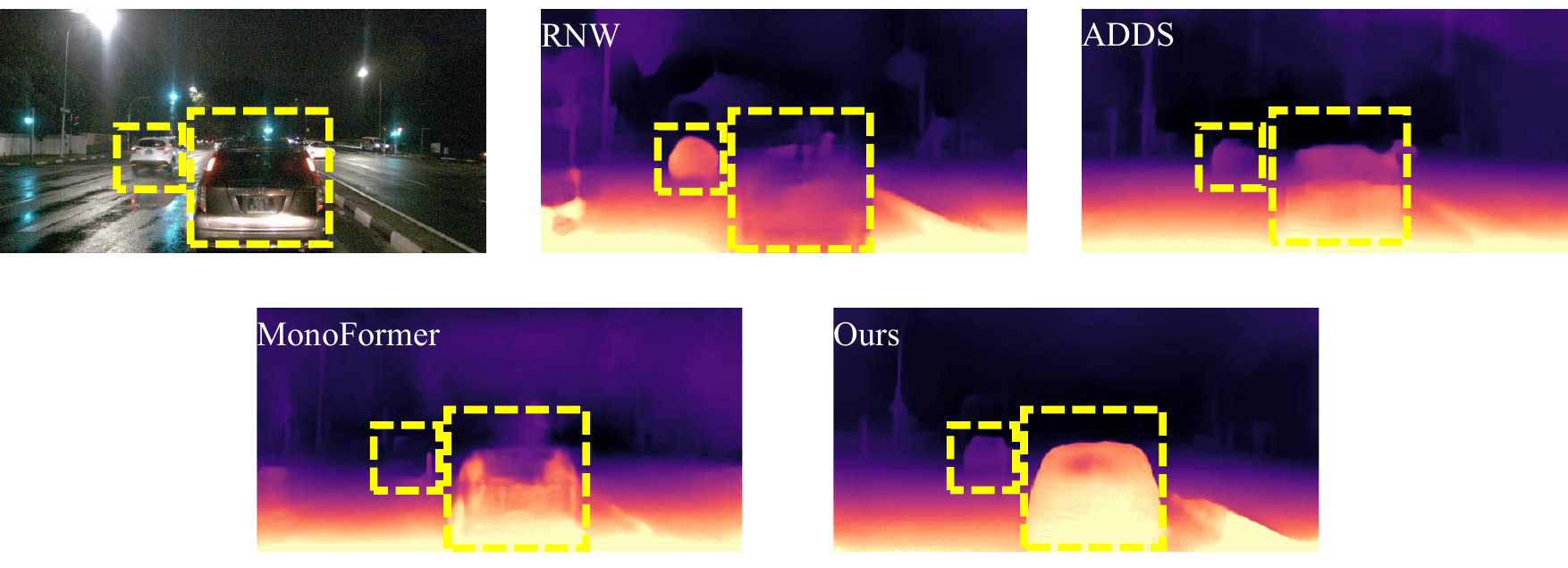}
    \caption{{\textbf{Nighttime monocular estimation results of different self-supervised frameworks.}} Compared with existing domain adaptation-based methods RNW~\protect\cite{rnw} and ADDS~\protect\cite{adds} and recent large model MonoFormer~\protect\cite{monoformer}, our result shows superior performance.}
    \label{Fig:1}
    \vspace{-2mm}
\end{figure}

While for nighttime scenes, the complex lighting conditions lead to significant photometric inconsistency within night-image sequences. And it causes divergence of training.  
To alleviate this problem, domain adaptation-based methods and photometric loss repair-based methods are proposed recently.
The domain adaptation-based methods~\cite{adds,rnw,ITDFA,adfa} apply extra adversarial loss or domain transfer network to make the depth model learn to decouple 
structure information and interfering elements.
Meanwhile, the photometric loss repair-based method~\cite{alldaydemono} utilizes additional trainable and trained parts to complement the daytime photometric loss.
Although these methods show improved results compared to the classical self-supervised framework~\cite{monodepth2}, their performance is still limited by the poor transfer quality of the image transfer model or the incomplete elimination of lighting effects in the photometric loss.

In this paper, we suspend the direct application of nighttime images during training since they are an inappropriate source of self-supervision.
Our goal is to train a night depth network that does not see night images during training, but only day images, in a self-supervised manner.
We achieve such a design by extracting some principal day-night dissimilarities and using physical priors to compensate for the day-image distribution.
Focusing on the difference in lighting conditions, the dissimilarities in photometric and noise distribution are located as two key components.
Then, we build up corresponding modules for simulation.
Considering the wave optics/diffraction effect of light sources as well as reflections, we propose Brightness Peak Generator~(BPG) to model the difference in photometric distribution.  
Based on the shot-read noise model, we build up Imaging Noise Generator~(ING) to model nighttime noise distribution. 
The joint application of BPG and ING together with the day-image distribution results in a fused distribution that has a photometric and noise distribution close to the night images.
The fused distribution is randomly sampled and used for the training of the depth network.
As shown in Fig.~\ref{Fig:1}, though \textbf{no night image} is seen during training, our method still achieves superior performance compared to existing nighttime methods. 

Our main contributions are summarized as follows:
    \textbf{(I)}
    We propose a nighttime monocular depth training framework that use day image pairs as the stable source of self-supervision.
    \textbf{(II)}
    Our training framework requires no night images during training.
    We accomplish this design by day-to-night data distribution compensation.
    \textbf{(III)}
    Photometric and noise distributions are located as two key day-night differences. 
    Using physical priors, we propose two simple but effective modules to model the differences in these distributions accordingly.
    Our presented method achieves SoTA performance on the challenging nuScenes-Night and RobotCar-Night dataset, though no nighttime images are used in our training framework.

%-------------------------------------------------------------------------
\section{Related work}
Estimating depth from a single image is an ill-posed problem, and deep learning-based frameworks address this challenge in an end-to-end manner and show promising performance~\cite{monocular_review_2020}. Considering the availability of precise depth labels in the real world, self-supervised solutions~\cite{sfmlearner,monodepth2} propose to use geometric constraints between image sequences instead of depth labels during training and received increasing attention recently~\cite{monocular_review_2020}.
\subsubsection{Daytime Self-supervised Framework} Considering the geometric constraints between temporal images, 
SfMLearner~\cite{sfmlearner} is proposed to use monocular image sequences to train the monocular depth network.
During training, a depth network and a pose network are designed to construct the geometric relationship between images, and a view reconstruction loss is used to supervise the training process.
Based on this framework, many works are proposed to handle the challenges caused by occlusions~\cite{monodepth2}, dynamic objects~\cite{dynamicdepth1,dynamicdepth2} and scale ambiguity~\cite{depth_scale_recovery,depth_scale_recovery2,scale_recovery_pseudo_label_depth}.
Meanwhile, some methods~\cite{segdepth1,segdepth2,segdepth3,segdepth4,segmentmonodepth} introduce and fuse semantic information into training to improve the depth estimation performance.
Many methods~\cite{monovit,hr-depth,packnet,cadepth} try to improve the depth estimation from the aspect of network architectures.
Most recently, MonoViT~\cite{monovit,MDC} and MonoFormer~\cite{monoformer} combine both CNN and Transformer blocks to learn both local and global features of images, and their results on unseen scenes show the convincing generalization ability of their network.

\subsubsection{Nighttime Self-supervised Framework}
Although the above methods show satisfactory depth estimation results, for nighttime scenes, such methods usually failed during training and testing because of significant day-night distribution differences.
To estimate in the night, many methods~\cite{adds,rnw,ITDFA,adfa} are proposed to address the challenge through domain adaptation or photometric loss restoration.
ADFA~\cite{adfa} succeeds in nighttime training by domain adaptation on feature space. They train a nighttime DepthNet encoder to learn daytime-like features with adversarial loss.
ITDFA~\cite{ITDFA} and ADDS~\cite{adds} utilize the image transfer-based domain adaptation framework. Based on the transferred image pairs, ITDFA~\cite{ITDFA} constrains the training process on both feature and output spaces, while
ADDS~\cite{adds} tries to learn a better decoupling of the private and invariant domains.
However, the performance of ITDFA and ADDS is restricted by the image transfer quality. Defects in the texture can cause the prediction missing structural details. 
In RNW~\cite{rnw}, an improved priors-based domain adaptive on output space is used to regularize nighttime photometric outliers. 
However, the significant illumination inconsistency within poor-quality image sequences is still out of regularization and causes performance degradation.
WSGD~\cite{alldaydemono} is the first one-stage method that directly trains their proposed framework on nighttime image splits. 
It introduces an Illuminating Change Net, a Residual Flow Net and a frozen Denoise Net~\cite{n2n_denoise1,n2n_denoise2} into monocular training. 
Instead of applying domain adaptation, they utilize extra trainable and trained parts to refine the photometric loss.

Though these methods show nighttime monocular depth estimation ability, low light and complex light in nighttime scenes still affect the photometric-based self-supervised training process. Meanwhile, the benefits of the image transfer model are limited. 
In contrast, our method performs data distribution compensation on day images and does not suffer from the poor self-supervision of night images.
%-------------------------------------------------------------------------
\begin{figure*}[htbp]
    \hsize=\textwidth
    \centering
    \includegraphics[width=0.75\linewidth]{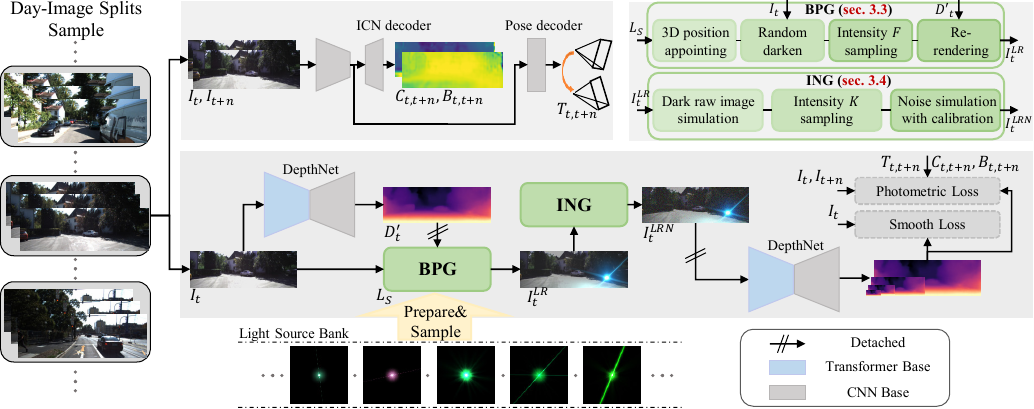}
%\vspace{-0.4cm}
    \caption{
    {\textbf{Overview of our data distribution compensation training framework}.}
    The proposed BPG and ING form our compensation stage, whose simple processes are also visualized in the top right. 
    Note that BPG and ING will not participate in the backward propagation. 
    Their input and output are detached.
    The Transformer-CNN hybrid DepthNet, CNN-based PoseNet and CNN-based Illuminating Change Net~(ICN) constitute the trainable part of our framework. 
    The two DepthNets share the same weights during training and the left one are frozen during the whole training. The images inputting of the pose network part and the loss part will not be pre-processed by BPG and ING, we discuss this setting in supplementary. 
    }
    \label{Fig:full_pipeline}
    \vspace{-2mm}
\end{figure*}

\section{Approach}
 % show out the very key point in the first place
\subsection{Data Distribution Compensation Framework}
\subsubsection{Compensation and sample}
To enable the depth network that sees no night image but generalizes well to different night scenes, 
and to take advantage of the stable self-supervision of day images, we make a compensation on the day-image distribution with day-night differences:
\begin{equation}
    \begin{split}
        & {\cal P}_n = {\cal P}_d + {\cal P}_{shift}, \\
        & I_t^{LRN} \sim {\cal P}_d + \tilde{\cal P}_{shift},
    \end{split}
\end{equation} 
where ${\cal P}_d$ and ${\cal P}_n$ represent the day-image and night-image distributions, $ {\cal P}_{shift} $ donates the distribution difference between them, $\tilde{\cal P}_{shift}$ refers to the simulation of some key differences, and $I_t^{LRN}$ indicates one sample from the fused distribution.

\subsubsection{Physical Priors or Image Transfer}
Either physical priors or image transfer technology can be used to build the fused distribution, $ {\cal P}_d + \tilde{\cal P}_{shift} $.
Though vivid fake night images could be transferred with solid backbones and properly partitioned training sets, some key differences~(e.g., differences in photometric and noise distribution) between day and night could be underestimated in the image transfer network because no corresponding constraint is made during their training.

Different from image transfer technology, the use of physical priors is more controllable, directional and explainable. 
It's also efficient because it doesn't require additional training stages or excessive memory costs. 
Thus, we use physical priors to model $\tilde{\cal P}_{shift}$. 
In addition, focusing on lighting conditions, we analyze key day-night differences in Section~\ref{sec:dnida} and describe their modeling in Section~\ref{sec:pdm} and Section~\ref{sec:inm}.

\subsubsection{Self-supervised Training}
The proposed framework is trained in a self-supervised manner. Following~\cite{monodepth2,monovit}, the combination of photometric loss and edge-aware smooth loss is used as the main supervision during training, and the framework is shown in Fig.~\ref{Fig:full_pipeline}.
We use the same Transformer-CNN hybrid network in MonoViT~\cite{monovit} as our DepthNet because of its good performance.
Besides, inspired by~\cite{alldaydemono}, we further consider the potential illuminating changes between images, and the Illuminating Change Net~(ICN) is introduced to predict the linear per-pixel illuminating change $ C_{t,t+n} $ and $ B_{t,t+n} $.
In addition, to avoid poor self-supervision, we make a simple decoupling at the input stage, and the sampled $I_t^{LRN}$ is only applied to the input of the depth network.
\subsection{Day-Night Image Differences Analysis}
\label{sec:dnida}
Objects~(cars, trees, pedestrians, streets, sky, etc.) in day and night scenes share the same internal properties and {similar} distributions, but when captured objects on images by camera sensors, the reactions of objects and sensors in different lighting conditions result in dissimilarities in photometric and noise distribution.

\subsubsection{Photometric distribution} For daytime scenes, the parallel white light from the sun illuminates the scene and reflects the color of objects.
The color, intensity and direction of the light are almost exactly the same between images, and the evenly distributed brightness matches the texture gradient of the objects well so that the photometric consistency assumption can be built on daytime images~\cite{monodepth2,ITDFA,alldaydemono}.
While for nighttime scenes, dynamic/static light sources together with different light colors result in a highly complex and nonuniform photometric distribution on the scenes.
Besides, on the image coordinates, the correspondence between the brightness gradient and the textures fails because the significant brightness peaks caused by the light sources overwhelm the texture gradient of scenes.
\subsubsection{Imaging noise}  
Since cameras are not perfect imaging device, it has a limited dynamic range and introduces noise at most stages of the imaging process~\cite{SID}.
At night, the camera adjusts the system gain to compensate for the lower number of input photons, so noise is amplified and appears more pronounced in the raw image~\cite{denoise_ELD,denoise_ELD_v2,denoise_rethinking}.
When the noise intensity exceeds the denoising capability of the Image Sensor Processor~(ISP)~\cite{SID,google_raw_denoise}, it will not be negligible in the output image.
Heavy noise distorts the local distribution and creates fake textures on both foreground and background, and such fake textures confuse the daytime depth model during testing. 

\subsection{Photometric distribution modeling}
\label{sec:pdm}
In this part, we model the nonuniform photometric distribution at night by adding additional light sources. 
The wave optics/diffraction effect of light source imaging and reflections are considered in the sampling. 
Since it makes peaks of brightness in the original image plane, we call it Brightness Peak Generator~(BPG).

\subsubsection{Light Source Sample}Due to the diffraction/wave optics effects~\cite{wave_optics_theory1,wave_optics_theory2}, 
when a point light is viewed through an aperture that is not an ideal circle or a lens that is stained or scratched,
the imaging results of the light source will be a combination of glare, streak, 
and shimmer, instead of a dot~\cite{flare7k,old_glare_maker}. 
In computer graphics, some methods~\cite{old_glare_maker,gazeflare} 
approximate this optical phenomenon by using the 2D Fourier transform, but the output is uncontrollable and the cost is expensive.
Therefore, instead of simulating the light source image directly, we construct a light source bank based on Flare7K~\cite{flare7k} dataset, which is the only light source dataset with 7000 samples.

When applied, BPG randomly samples a single image from the light source bank. 
Then, the image will be scaled to the standard size which is determined by the long side of $ I_t $~(A square whose sides are equal to the long side of $ I_t $ ). 
Simple image augmentation is further applied to expand the number of light source images. 
The scaled and augmented light source image is used as the standard light source image $ L_S $.

\subsubsection{3D Position of Light Source} 
% The sampled light source is assumed to be randomly distributed in the scene.
{
The sampled light source is assumed to be randomly distributed in the scene.
We do not model the location because it is costly to include precise segmentation information in our training which requires large trained models.
}

Firstly, a 2D coordinate $ p_{i} $ is randomly appointed as the location of the light source. 
We limit the depth of it, $z_i$, with the predicted scale depth map~(discussed latter in this section).
A up range with higher priority was manually set to 25m as farther light sources produce almost invisible reflection images.
Then, the appointed 3D position will be:
\begin{equation}
    \label{equ:P_i}
    P_i=z_i K_I^{-1}\dot{p_i}.
\end{equation}

\subsubsection{Random Darken}
The above light source imaging $ L_S $ cannot produce significant peaks of brightness on the image because the daytime image is usually well-lit with high brightness. To efficiently improve the role of $ L_S $ on the photometric distribution, a simple random darkening operation is applied.
The illumination scale rate is set to follow uniform distribution, i.e. $ s_d \sim {\cal{U}}(0.4,1) $. 

\subsubsection{BPG Intensity $ F $}We use the product of the light sources number $ N_F $ and a resize scale rate $ s_F $ as the BPG Intensity $ F $. 
To enrich outputs while avoiding numerous aggressive results, a 
log uniform distribution is applied to select $ N_F $ and $ F $:
\begin{equation}
    \label{equ:flare_intensity}
    \begin{split}
        & \log s_F \sim {\cal{U}}(\log s_F^{\min},\log s_F^{\max}),\\
        & \log {F} \sim {\cal{U}}(\log F^{\min},\log F^{\max}),\\
        & N_F=\max (\lfloor\frac{F}{s_F}+\frac{1}{2}\rfloor,1).
    \end{split}         
\end{equation}

Following Flare7K~\cite{flare7k}, BPG adds the light source within the gamma range of $ g_f \sim {\cal{U}}(1.8,2.2) $.
The sampled image on current stage, $ I_t^{L} $,  will be 
\begin{equation}
    I_{t}^L=\left( \left(s_dI_t\right)^{g_f}+\sum_{i=1}^{N_F} \mathtt{ss} \left(L_S, s_F, p_i \right)^{g_f} \right)^{1/g_f}.
\end{equation}
$ N_{F} $ is the total number of light sources, and $ \mathtt{ss}(.,.,.) $  
represents scaling and shifting the sampled $ L_S $ 
by the scale rate $ s_F $ and the 2D coordinate $ p_i $. 

\begin{figure}[htbp]
    \centering
    \includegraphics[]{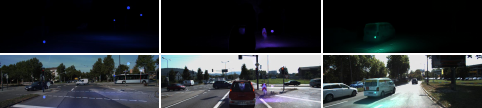}
    \caption{
    \textbf{Example visualizations of Re-rendering.} 
    Top: Reflection images. Bottom: Re-rendered images. 
    % Dot circles are used to mark locations.
    \vspace{-1mm}
    }
    \label{Fig:re_rendering_samples}
\end{figure}

\subsubsection{Re-rendering}
Reflections are also one of the important representations of light sources in the scene, affecting the color and texture of objects in the image plane.
Thus, we build up a Re-rendering submodule within BPG for compensation.

With limited and biased information (3D structure, surface normal, material, etc) about the scene, we find the popular used Phong illumination model~\cite{phong} effective.
Note that only the predicted unscaled depth map $ D'_t $, the color image $ I_t $, and the camera intrinsic $ K_I $ are used in the Re-rendering submodule.

Since the predicted depth map $ D'_t $ differs from the real-world values by a scale factor $ s $, following 
\cite{depth_scale_recovery,scale_recovery_pseudo_label_depth}, we use the unscaled camera height to predict the scale factor. 
By predicting surface normal from $ D'_{t} $, the unscaled camera height $ H_{c}' $ is estimated thereafter. Thus, the scale factor will be
\begin{equation}
    s = H_{c}'/{H_c},
\end{equation}
where $ H_c $ is the actual camera height. 
With the predicted scale factor $ s $,  the structured point cloud $ P $  can be calculated with
\begin{equation}
    P(u)=sD'_t(u)K_I^{-1}\dot{u}.
\end{equation}
We then use predicted depth map $D'_{t}$ to calculate our surface normal~\cite{depth2nv} and directly use the color image $ I_t $ to extract coarse material information.
Using the structured point cloud $ P $, surface normal, coarse material of objects, the color of light source together with the appointed position, $ P_i $~(Eq.~\ref{equ:P_i}), we can calculate the reflection image $ I_i^{R} $ with simple but effective Phong illumination model~\cite{phong}. 
The sampled image on current stage, $ I_{t}^{LR} $, will be:
\begin{equation}
    I_{t}^{LR}=I_t^{L}+\sum_{i=1}^{N_F} I_i^{R}.
\end{equation}
The detailed calculation of $ I_i^{R} $ is shown in the supplementary, and we give some examples of re-rendering in Fig.~\ref{Fig:re_rendering_samples}.

\subsection{Imaging noise modeling}
\label{sec:inm}
Heavy noise at night corrupts the local distribution and makes fake textures.
Based on the shot-read noise model~\cite{denoise_ELD,denoise_ELD_v2}, we build our Imaging Noise Generator~(ING) using color images as input.
 
\subsubsection{Noise Model}
For a camera imaging system, if without Image Signal Processor~(ISP), a linear model can generally formulate the digit sensor raw image~\cite{denoise_ELD,pmn_raw_denoise}:
\begin{equation}
    R^{*}=KC+N=R+N,
\end{equation}
where $ C $ is the number of photon, $ K $ donates the overall system gain, and $ N $
refers to the summation of all physical noise. In the physical model base image 
denoising~\cite{denoise_ELD,denoise_ELD_v2,denoise_rethinking},
the overall noise $ N $, is roughly separated into shot noise $ N_p $ and read noise
$ N_{read} $:
\begin{equation}
    \label{equ:noise model}
    N = KN_p + N_{read}.
\end{equation}

Here, Shot noise is caused by the collection uncertainty of photons, which follows
\begin{equation}
    (C+N_p) \sim {\cal{P}} (C).
    \label{equ:position}
\end{equation}
with $ \cal P $ indicating Poisson distribution.

Meanwhile, the composition of read noise is more complex, and $ N_{read} $ is usually
assumed to follow a bell-shape distribution e.g. Gaussian distribution, 
$ {\cal N}(0,\sigma_N) $ and Tukey lambda distribution~\cite{Tukey_lambda}, $ TL(\lambda_{TL};0,\sigma_{TL}) $. 
$ TL $  is a distribution family that can fit many bell-shaped distributions, and $ \lambda_{TL} $ is used to control the shape. 
Applying the linear least squares method, the system gains $ K $ and the variance of the distribution are considered to be linear in the logarithmic domain~\cite{denoise_ELD}:
 \begin{align}
     \label{equ:logKlogsigma_N}
     & \log {(\sigma_N)} | \log K \sim {\cal{N}} (a_{N} \log (K) + b_{N}, \hat{\sigma}_{N}),\\
     \label{equ:logKlogsigma_TL}
     & \log {(\sigma_{TL})} | \log K \sim {\cal{N}} (a_{TL} \log (K) + b_{TL}, \hat{\sigma}_{TL}), 
 \end{align}
with $ a_N $($ a_{TL} $) and $ b_N $($ b_{TL} $) being the approximating linear parameters and 
$ \hat{\sigma}_{N} $($ \hat{\sigma}_{TL} $) donating the standard deviation of the unbiased estimation. 
In ELD~\cite{denoise_ELD}, different kinds of digit sensors are calibrated with Gaussian distribution and Tukey lambda
distribution respectively and ING randomly samples the calibrations to generate
read noise $ N_{read} $. 
  
\subsubsection{From RGB to simulated dark raw image $ R_t^{LR} $}In ELD, shot noise $ N_p $ is applied
on the number of photons $ R/K $ while read noise $ N_{read} $ on raw image $ R $.
However, raw images or the exact inverse ISP are not available in most daytime datasets.
The simplest ISP includes white balance, binning, denoising, color collection and 
gamma compression~\cite{SID,google_raw_denoise}.
Fortunately, except for denoising and gamma compression, rest operations can be viewed as approximately linear processes. 
We consider the gamma compression in ING as it's strongly nonlinear. 
Following ELD, we set $ g_n = 1/2.2 $. 

To simulate the low photon count $ C $  in the dark, 
a light scale factor $ s_n $ is proposed follows $ {\cal{U}}(100,300) $~\cite{denoise_ELD}.  
Then, the simulated raw image will be
\begin{equation}
    R_t^{LR}=\frac{s_{bit}(I_t^{LR})^{1/g_n}}{s_n},
\end{equation}
where $ s_{bit} $ denotes quantization factor equals to $ 2^{bit} - 1$.

\subsubsection{ING Intensity $ K $}According to Eq.~\ref{equ:noise model}, Eq.~\ref{equ:logKlogsigma_N}
and Eq.~\ref{equ:logKlogsigma_TL}, system gain $ K $ has a positive relationship with the noise
intensity. 
% Besides, the system gain $ K $  will be promoted at night to fit the low photon count.  
Therefore, we appoint $ K $ as ING Intensity. Similar in
BPG, we make ING Intensity following the log uniform:
\begin{equation}
    \label{noise_intensity}
    \log K \sim {\cal{U}} (\log K^{\min},\log K^{\max}).
\end{equation}

\begin{figure}[htbp]
    \centering
    \newcommand{\turnheightnew}{0.31\columnwidth}
    \begin{tabular}{@{}c@{\hskip 0.3mm}c@{\hskip 0.3mm}c@{\hskip 0.3mm}c@{}}
    \vspace{-1mm}
    {\rotatebox{90}{\hspace{3mm}\scriptsize{$I_t$}}} &
    \includegraphics[width=\turnheightnew]{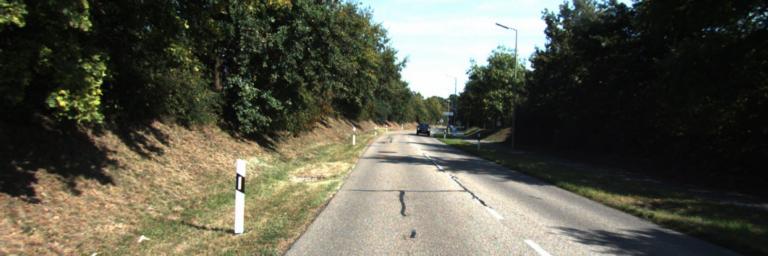} &
    \includegraphics[width=\turnheightnew]{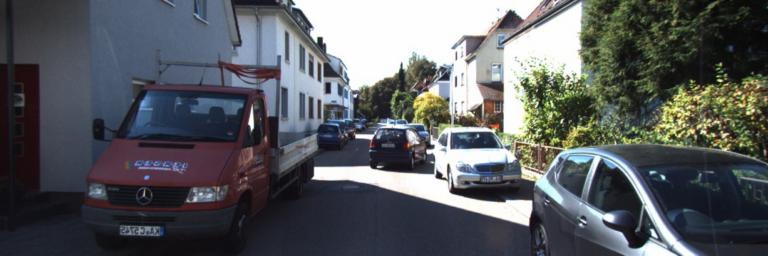} &
    \includegraphics[width=\turnheightnew]{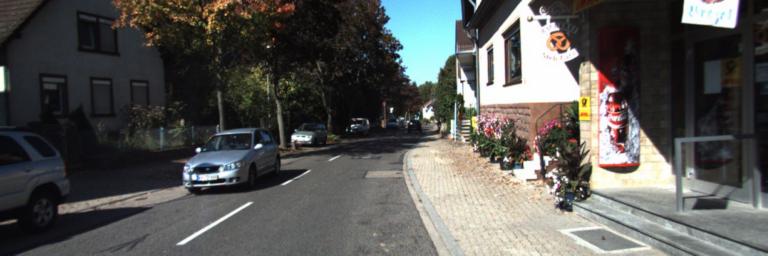} \\
    \vspace{-1mm}
    {\rotatebox{90}{\hspace{1.2mm}\scriptsize{$I_t^{LR}$}}} &
    \includegraphics[width=\turnheightnew]{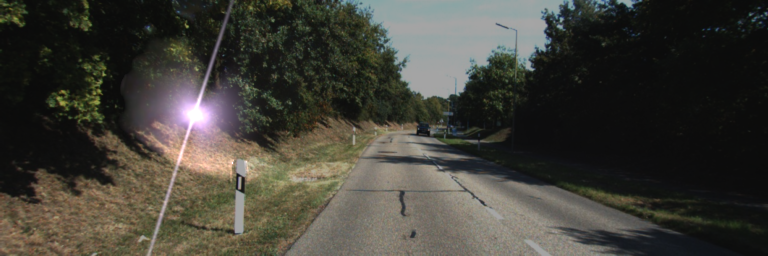} &
    \includegraphics[width=\turnheightnew]{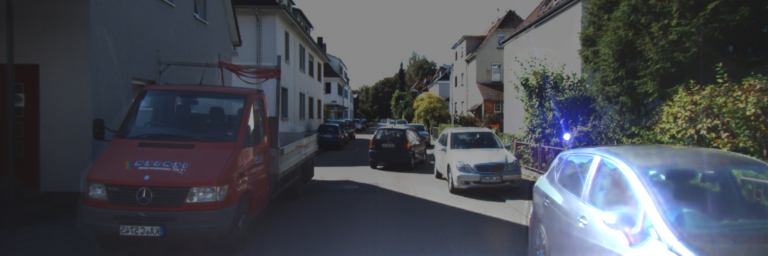} &
    \includegraphics[width=\turnheightnew]{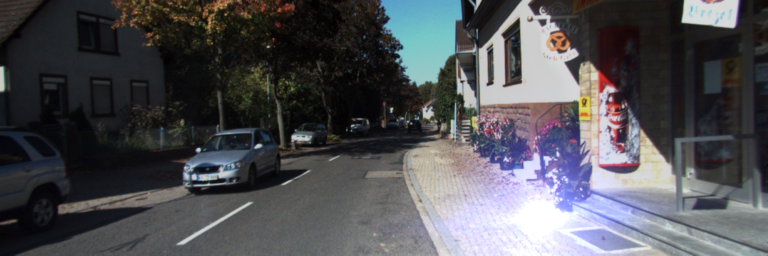} \\
    \vspace{-1mm}
    {\rotatebox{90}{\hspace{0.2mm}\scriptsize{$I_t^{LRN}$}}} &
    \includegraphics[width=\turnheightnew]{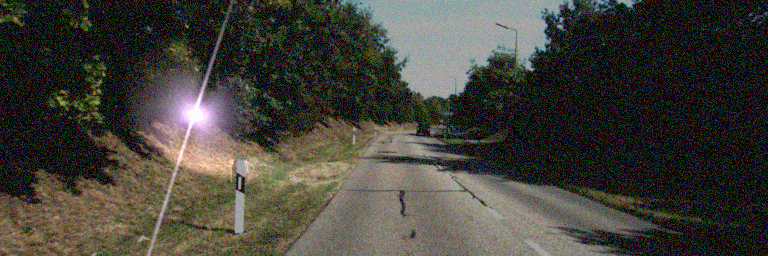} &
    \includegraphics[width=\turnheightnew]{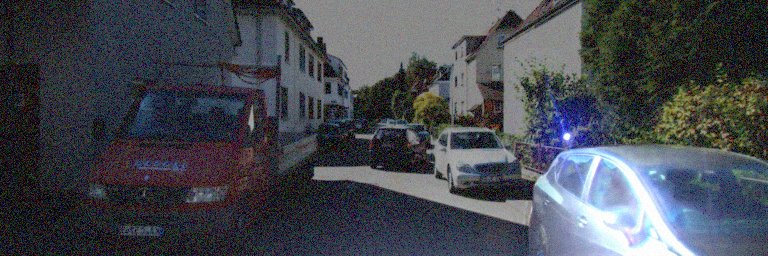} &
    \includegraphics[width=\turnheightnew]{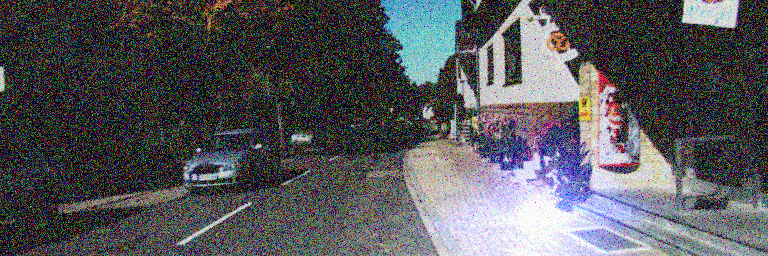} \\

    \end{tabular}
    \vspace{1mm}
    \caption{\textbf{Paired visual examples}. (Best view with zoom.) 
    % We leave the FID test to the \emph{supplementary material}.
    } 
    \label{Fig:FN_samples}
    \vspace{-1mm}
\end{figure}

Finally, with the simulated raw image $ R_t^{LR} $, the sampled shot noise $ N_p $ and 
read noise $ N_{read} $, the output of ING will be 
\begin{equation}
    I_t^{LRN}=\left(\frac{s_{n}R_t^{LR}+KN_p+N_{read}}{s_{bit}}\right)^{g_n}.
\end{equation}
Fig.~\ref{Fig:FN_samples} visualizes some examples of $ I_t^{LRN} $.

%-------------------------------------------------------------------------
\section{Experiments}

\begin{table*}[htbp]
    % \footnotesize
	\begin{center}
\scalebox{0.8}{
	\scriptsize
        \begin{tabular}{c| l | c | c | c | c c c c | ccc}
            \hline
            Type & Method & Train on & Train Res.  & Max depth & \cellcolor{cell1} ABS rel$\downarrow$ & \cellcolor{cell1} Sq rel$\downarrow$  & \cellcolor{cell1} RMSE$\downarrow$ & \cellcolor{cell1} RMSE log$\downarrow$  & \cellcolor{cell2} $\delta_{1} \uparrow$ & \cellcolor{cell2} $\delta_{2} \uparrow$ & \cellcolor{cell2} $\delta_{3} \uparrow$\\
            \hline
            \multicolumn{12}{c}{\cellcolor{cell3} Test on nuScenes-Night} \\

            \multirow{2}{*}{DT}
            & MonoViT\cite{monovit}           & N $ d \& n $ &  640 $\times$ 320   & 60     & 1.726     	& 93.031    	& 30.321    	& 2.183     	& 0.143     	& 0.291     	& 0.437  \\
            & WSGD\cite{alldaydemono}& N $ d \& n $ &  640 $\times$ 320   & 60     & 0.663     	& 9.573     	& 15.200    	& 0.755     	& 0.199     	& 0.388     	& 0.567  \\
            \hline
            \multirow{4}{*}{DA}
            & ITDFA\cite{ITDFA}               & N $ d \& n $ &  640 $\times$ 320   & 60       & 0.337     	 & 4.511     	& 10.118    	& 0.403     	& 0.515     	& 0.767     	& 0.890  \\        
            & RNW\cite{rnw}\dag               & N $ d \& n $ &  768 $\times$ 384   & 60       & 0.315		  & 3.792       & \underline{9.641}		    & 0.403		    & 0.508		    & 0.778		    & 0.896 \\
            & RNW\cite{rnw}                   & N $ d \& n $ &  640 $\times$ 320   & 60       & 0.341          & 5.516     	& 11.152    	& 0.406     	& 0.531     	& 0.789     	& 0.902 \\
            & ADDS\cite{adds}                 & N $ d \& n $ &  640 $\times$ 320   & 60       & \underline{0.299}     	& 4.790     	& 10.372    	& \underline{0.371}     	& \underline{0.620}     	& 
            \underline{0.814}     	& \underline{0.907}  \\
            \hline
            \multirow{6}{*}{G}
            & WSGD\cite{alldaydemono} & R $ d \& n  $     &  640 $\times$ 320   & 60 & 0.314     	& \underline{3.567}     	& 10.058    	& 0.408     	& 0.520     	& 0.758     	& 0.881 \\
            & ITDFA\cite{ITDFA}               & R $ d \& n $ &  640 $\times$ 320   & 60       & 0.362           & 3.760     	& 10.252    	& 0.441     	& 0.418     	& 0.702     	& 0.867  \\
            & RNW\cite{rnw}                   & R $ d \& n $ &  640 $\times$ 320   & 60       & 0.376           & 4.732     	& 11.193    	& 0.506     	& 0.451     	& 0.712     	& 0.835 \\
            & ADDS\cite{adds}                 & R $ d \& n $ &  640 $\times$ 320   & 60       & 0.322           & 4.401     	& 10.584    	& 0.397     	& 0.527     	& 0.786     	& 0.892 \\
            & MonoFormer\cite{monoformer}\ddag& K            &  768 $\times$ 256   & 60       & 0.307     	    & 3.591     	& 10.162    	& 0.413     	& 0.521     	& 0.762     	& 0.872  \\
            & MonoViT\cite{monovit}           & K            &  768 $\times$ 256   & 60       & 0.348          & 4.144          & 11.086    	& 0.473     	& 0.417     	& 0.708     	& 0.843 \\
            & \textbf{Ours}                   & K            &  768 $\times$ 256   & 60       & \textbf{0.259} & \textbf{3.147}& \textbf{8.547}& \textbf{0.344}& \textbf{0.641}& \textbf{0.850}& \textbf{0.928 } \\
            \hline
            % \hline
            % Type & Method & Train on & Test on & ABS rel & Sq rel  & RMSE & RMSE log  & $\delta_{1}$ & $\delta_{2}$ & $\delta_{3}$\\

            \multicolumn{12}{c}{\cellcolor{cell3} Test on RobotCar-Night} \\
            \multirow{2}{*}{DT}
            & MonoViT\cite{monovit}           & R $ d \& n $ &  640 $\times$ 320  & 40       & 0.513     	    & 13.558    	& 9.867     	& 0.479     	& 0.588     	& 0.846     	& 0.918  \\
            & WSGD\cite{alldaydemono}& R $ d \& n $ &  640 $\times$ 320  & 40       & \underline{0.202}& 1.835     	& 5.985     	& \textbf{0.231}& \underline{0.737}& \underline{0.934}& \underline{0.977}   \\
            \hline
            \multirow{4}{*}{DA}
            & ITDFA\cite{ITDFA}               & R $ d \& n $ &  640 $\times$ 320  & 40       & 0.266     	    & 3.010     	& 8.293     	& 0.287     	& 0.567     	& 0.888     	& 0.962  \\
            & ADDS\cite{adds}\dag             & R $ d \& n $ &  512 $\times$ 256  & 40       & 0.233          & 2.344         & 6.859         & 0.270         & 0.631         & 0.908         & 0.962  \\
            & ADDS\cite{adds}                 & R $ d \& n $ &  640 $\times$ 320  & 40       & 0.209     	& 2.179     	& 6.808     	& 0.254     	& 0.704     	& 0.918     	& 0.965 \\
            & RNW\cite{rnw}                   & R $ d \& n $ &  640 $\times$ 320  & 40       & \textbf{0.197} & \underline{1.789}     	& \underline{5.896}& \underline{0.234}& \textbf{0.742}& 0.930     	& 0.972 \\
            \hline     
            \multirow{6}{*}{G}     
            & ITDFA\cite{ITDFA}               & N $ d \& n $ &  640 $\times$ 320  & 40       & 0.302     	    & 3.692     	& 8.642     	& 0.327     	& 0.548     	& 0.852     	& 0.938   \\
            & ADDS\cite{adds}                 & N $ d \& n $ &  640 $\times$ 320  & 40       & 0.265     	    & 3.651     	& 8.700     	& 0.309     	& 0.640     	& 0.870     	& 0.945 \\
            & RNW\cite{rnw}                   & N $ d \& n $ &  640 $\times$ 320  & 40       & 0.237     	    & 2.958     	& 8.187     	& 0.298     	& 0.683     	& 0.885     	& 0.948 \\
            & MonoFormer\cite{monoformer}\ddag& K            &  768 $\times$ 256  & 40       & 0.289     	    & 2.893     	& 7.468     	& 0.302     	& 0.543     	& 0.873     	& 0.964  \\
            & MonoViT\cite{monovit}           & K            &  768 $\times$ 256  & 40       & 0.253     	    & 2.044     	& 6.208     	& 0.269     	& 0.572     	& 0.908     	& \underline{0.977} \\
            & \textbf{Ours}                   & K            &  768 $\times$ 256  & 40       & 0.210     	    & \textbf{1.515}& \textbf{5.386}& 0.238     	& 0.676     	& \textbf{0.936}& \textbf{0.980}  \\
            \hline
        \end{tabular}}
        \caption{\textbf{Effective test on nuScenes-Night}~\protect\cite{rnw,NuScenes} \textbf{and Robotcar-Night}~\protect\cite{adds,robotcar}.
     All methods use the \emph{same} DepthNet backbone unless marked.
    % The best results are in \textbf{bold}, while the second-best scores are \underline{underline}. 
    K, N and R indicate KITTI
    % ~\cite{kitti}
    , nuScenes
    % ~\cite{NuScenes} 
    and Oxford Robotcar
    % ~\cite{robotcar}
    dataset. 
    $ d $ and $ n $ are daytime and nighttime training splits proposed by RNW~\protect\cite{rnw} or ADDS~\protect\cite{adds}.
    Max depth here indicates the up range of ground truth depth.
    Note that \emph{no image} from the nuScenes or Oxford RobotCar datasets is used during our training and the applied resolution 768 $\times$ 256 is a little smaller than 640 $\times$ 320.
    \dag~means the original reported result in the paper with pure CNN backbone and Max depth as the clipping up range for the predicted depth (.i.e, a more relaxed approach to evaluation).
    \ddag~donates that the method applies a much larger Transformer-CNN hybrid backbone for DepthNet (about $\times$12 of parameters).
    \vspace{-1.5mm}
    }
    \label{Table:1}
    \end{center}
    \vspace{-2mm}
\end{table*}

\begin{table}[htbp]
    \centering
        \resizebox{0.4\textwidth}{!}{
        \begin{tabular}{c | c c c c | ccc}
            \hline
            Method & \cellcolor{cell1} ABS rel $\downarrow$ & \cellcolor{cell1} Sq rel $\downarrow$  & \cellcolor{cell1} RMSE $\downarrow$ & \cellcolor{cell1} RMSE log $\downarrow$  & \cellcolor{cell2} $\delta_{1} \uparrow$ & \cellcolor{cell2} $\delta_{2} \uparrow$ & \cellcolor{cell2} $\delta_{3} \uparrow$\\
            \hline
            \multicolumn{8}{c}{\cellcolor{cell3} Test on nuScenes-Night} \\
            Baseline  & 0.327     	& 3.740     	& 10.703    	& 0.448     	& 0.451     	& 0.733     	& 0.855 \\
            BPG Only & \underline{0.264} & \textbf{2.956}     	& \underline{9.209}     	& \underline{0.371}     	& 0.581     	& \underline{0.813}     	& \underline{0.914} \\
            ING Only  & 0.268     	& 3.177     	& 9.397     	& \underline{0.371}     	& \underline{0.588}     	& 0.807     	& 0.909 \\
            Full Method  & \textbf{0.259}& \underline{3.147}& \textbf{8.547}& \textbf{0.344}& \textbf{0.641}& \textbf{0.850}& \textbf{0.928}  \\
            \hline
            \multicolumn{8}{c}{\cellcolor{cell3} Test on RobotCar-Night} \\
            Baseline & 0.242     	& 1.882     	& 5.962     	& 0.261     	& 0.600     	& 0.915     	& \underline{0.979} \\
            BPG Only  & \underline{0.216}& 1.827     	& 5.986     	& \underline{0.248}& \textbf{0.686}& \underline{0.920}     	& 0.976 \\
            ING Only & 0.238     	& \underline{1.817}     	& \underline{5.882}     	& 0.257     	& 0.620     	& \underline{0.920}     	& \textbf{0.980} \\
            Full Method & \textbf{0.210}& \textbf{1.515}& \textbf{5.386}& \textbf{0.238}& \underline{0.676}     	& \textbf{0.936}& \textbf{0.980} \\
            \hline
        \end{tabular}
        }
        \caption{\textbf{Quantitative results of pre-processing ablation test.} 
        The baseline in this table is G:~MonoViT w. ICN.
        \vspace{-1mm}
        }
    \label{Talble:ablation}
    \vspace{-4mm}
\end{table}

\subsection{Dataset}
We construct the self-supervised training process on the widely used KITTI dataset, and following \cite{monodepth2,monovit}, the KITTI Eigen training split~\cite{kitti_eigen} is used as the training set 
due to its high-quality images, and we also regard it as our basic day-image distribution.

For evaluation, we use the nuScenes-Night test split from RNW~\cite{rnw} and the RobotCar-Night test split proposed in ADDS~\cite{adds}. And we also follow the same pre-cropping method proposed in their work.

\subsection{Effective Test} \label{Effective Test}
\subsubsection{Comparisons}
We mainly compare our results with four SoTA methods i.e. DA~(Domain adaptation):~RNW~\cite{rnw}, 
DA: ADDS~\cite{adds}, DA: ITDFA~\cite{ITDFA} and 
DT~(Direct training): WSGD~\cite{alldaydemono} on nuScenes-Night and RobotCar-Night,
respectively. 
To make a fair comparison, all methods apply the \emph{same} 
Transformer-CNN hybrid backbone~\cite{monovit} if no further explanation is made. 
We also provide the 
% G~(Generalization) 
{G~(Generalize to test set)}
version result of each DA method for 
further comparisons.
In addition, we will release the code upon acceptance.

\subsubsection{Evaluation on nuScenes-Night}
\begin{figure*}[t]
    \centering
    \newcommand{\turnheightnew}{0.25\columnwidth}
    \begin{tabular}{@{\hskip 0mm}c@{\hskip 1mm}c@{\hskip 1mm}c@{\hskip 1mm}c@{\hskip 1mm}c@{\hskip 1mm}@{\hskip 0mm}c@{\hskip 1mm}c@{\hskip 1mm}c@{\hskip 1mm}c@{\hskip 1mm}c@{\hskip 1mm}c@{}}
    {
    \vspace{-0.5mm}
    \rotatebox{90}{\hspace{0mm}\scriptsize{N-N \& R-N}}} &
    \includegraphics[width=\turnheightnew]{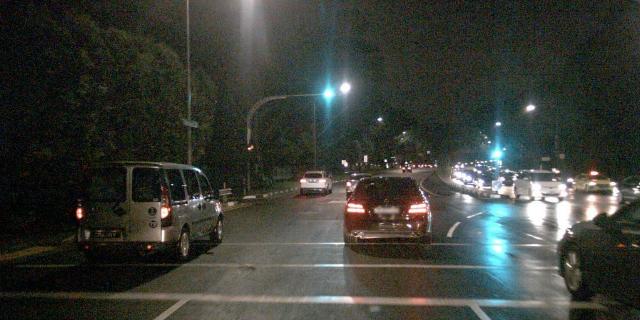} &
    \includegraphics[width=\turnheightnew]{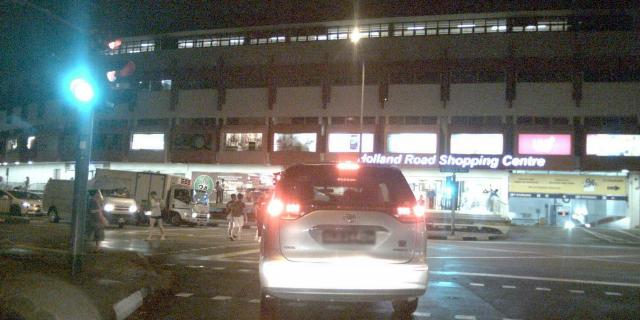} &
    \includegraphics[width=\turnheightnew]{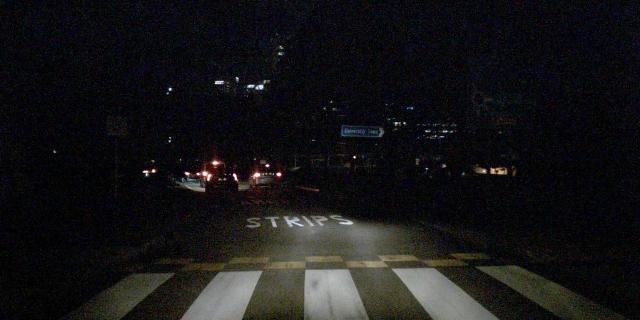} &
    \includegraphics[width=\turnheightnew]{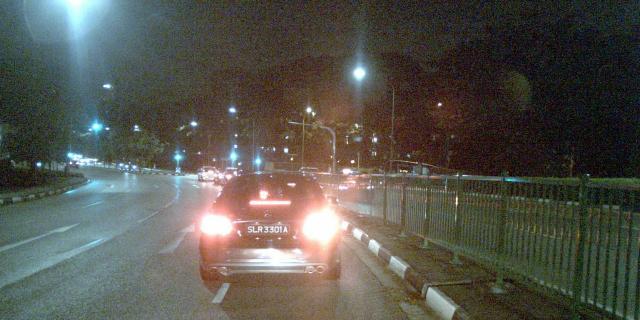} &
    
    \includegraphics[width=\turnheightnew]{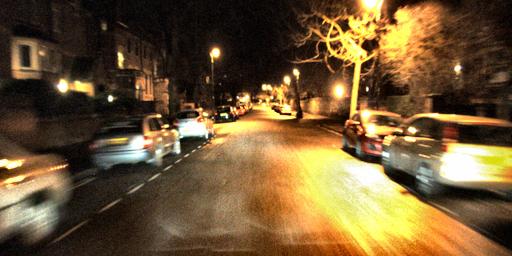} &
    \includegraphics[width=\turnheightnew]{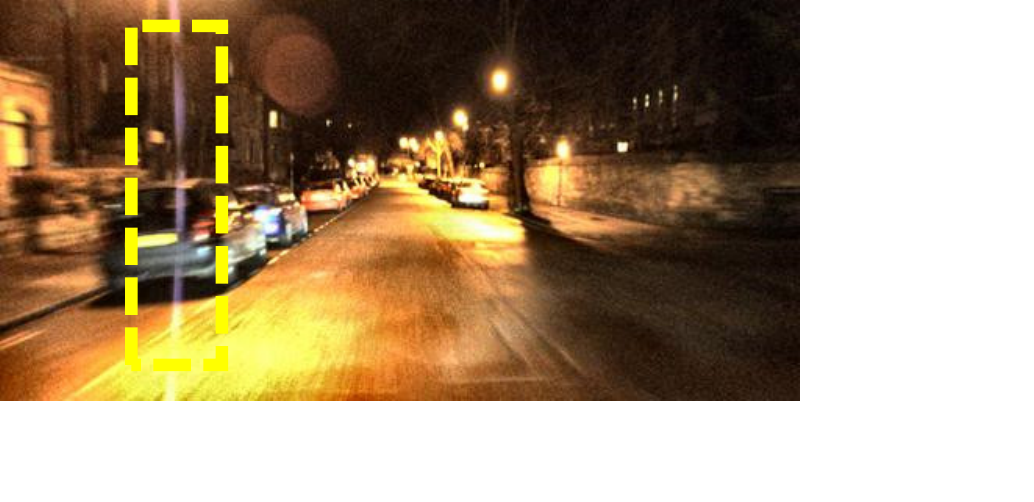} &
    \includegraphics[width=\turnheightnew]{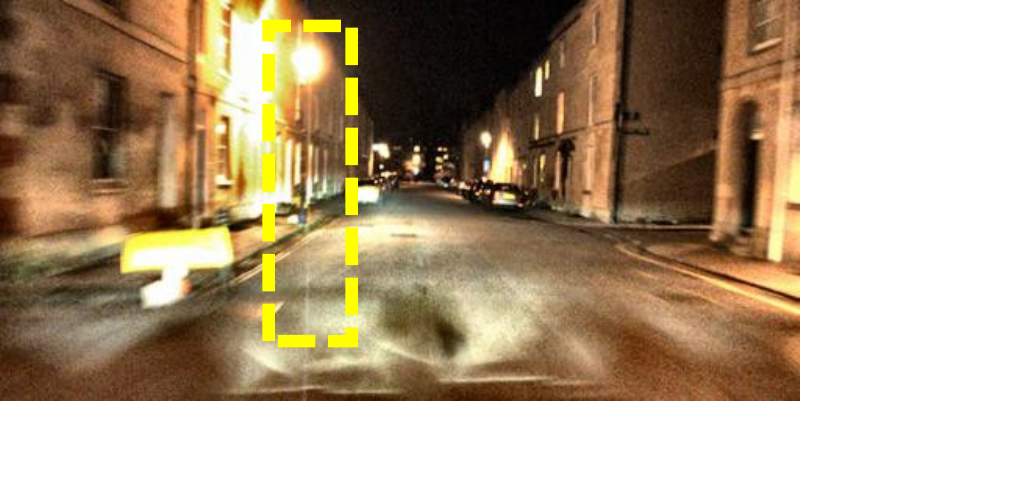} &
    \includegraphics[width=\turnheightnew]{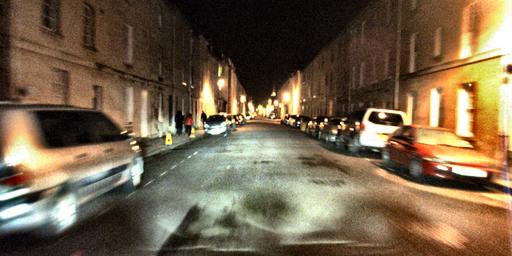} 
    \\
    
    \vspace{-0.5mm}
    \rotatebox{90}{\hspace{0mm}\scriptsize{DA: RNW}
    % \cite{rnw}
    }&
    \includegraphics[width=\turnheightnew]{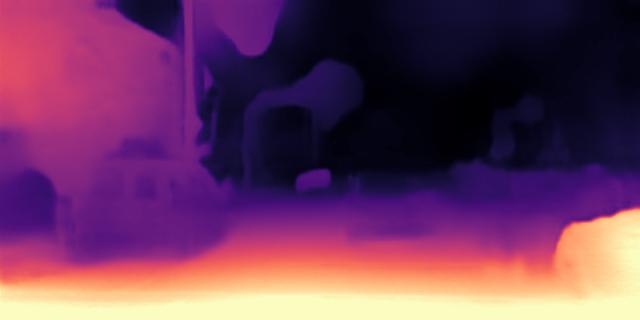} &
    \includegraphics[width=\turnheightnew]{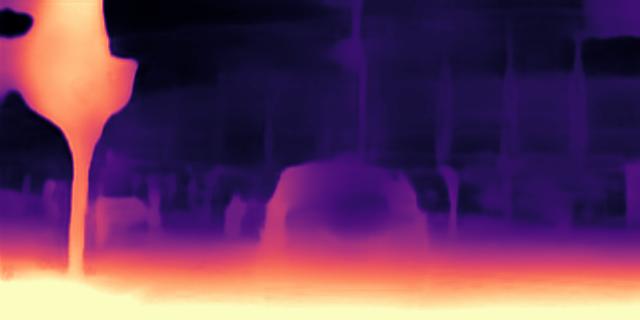} &
    \includegraphics[width=\turnheightnew]{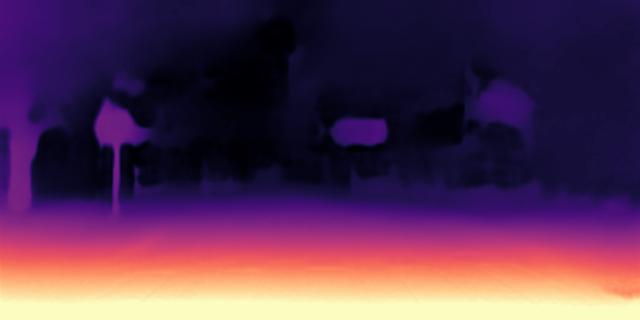} &
    \includegraphics[width=\turnheightnew]{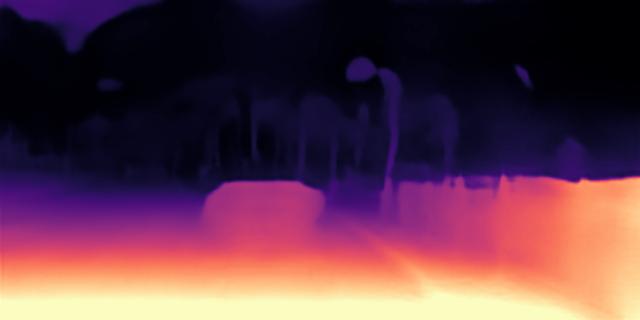} &

    \includegraphics[width=\turnheightnew]{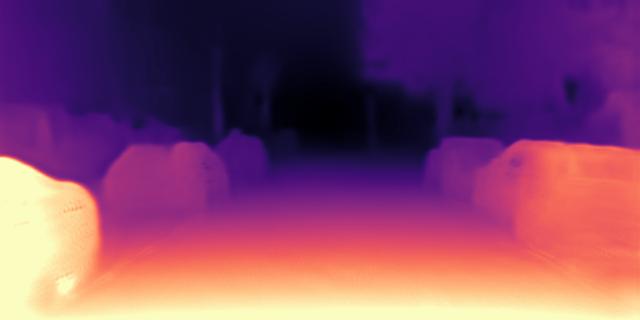} &
    \includegraphics[width=\turnheightnew]{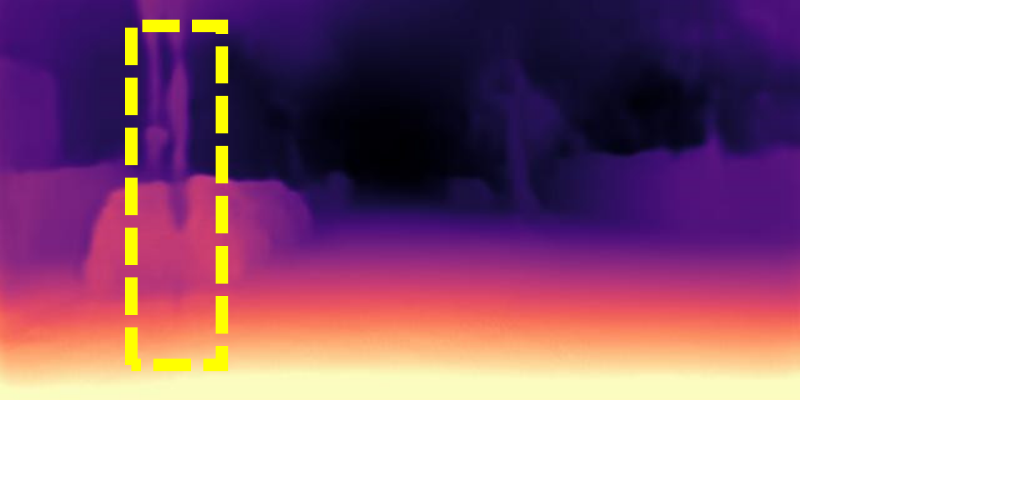} &
    \includegraphics[width=\turnheightnew]{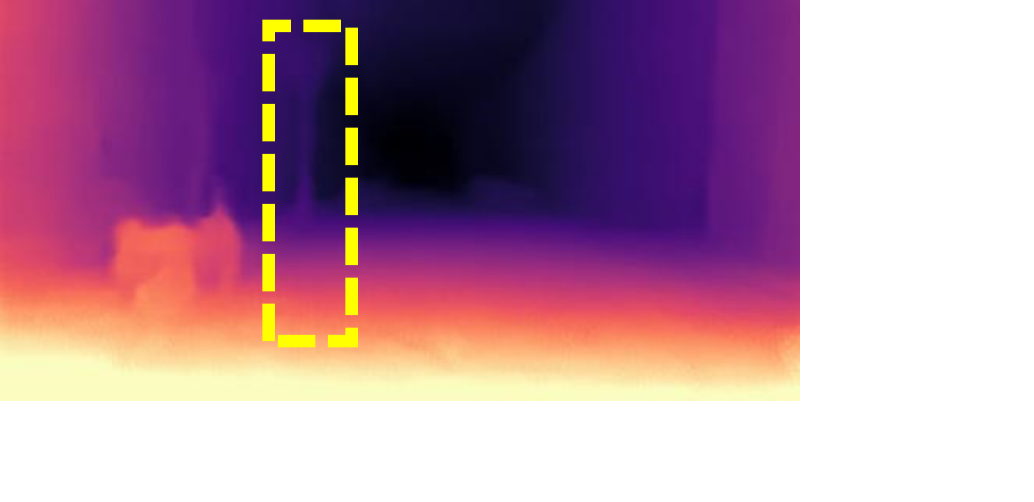} &
    \includegraphics[width=\turnheightnew]{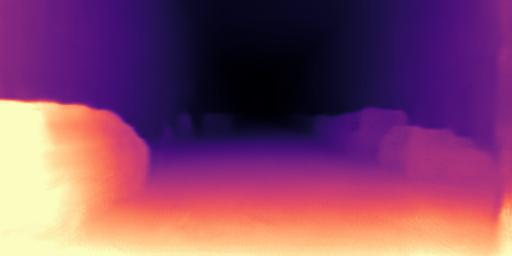} 
    \\
    
    \vspace{-0.5mm}
    \rotatebox{90}{\hspace{0mm}\scriptsize{DA: ADDS}
    % \cite{adds}
    }&
    \includegraphics[width=\turnheightnew]{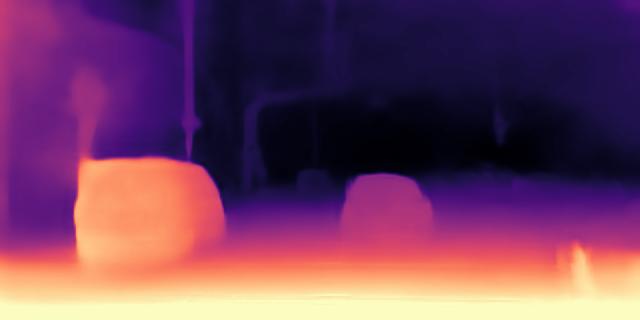} &
    \includegraphics[width=\turnheightnew]{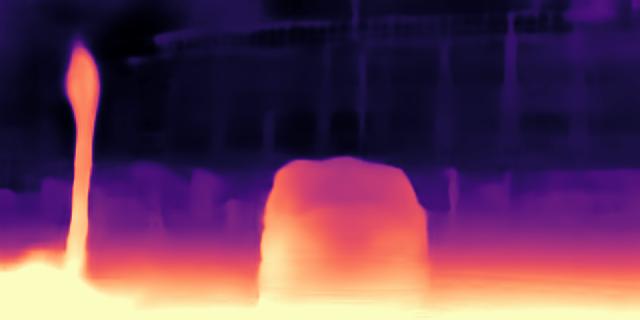} &
    \includegraphics[width=\turnheightnew]{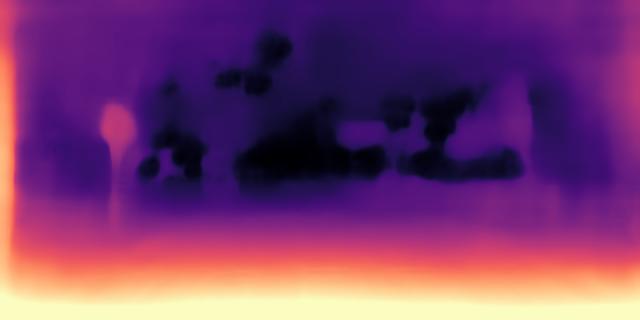} & 
    \includegraphics[width=\turnheightnew]{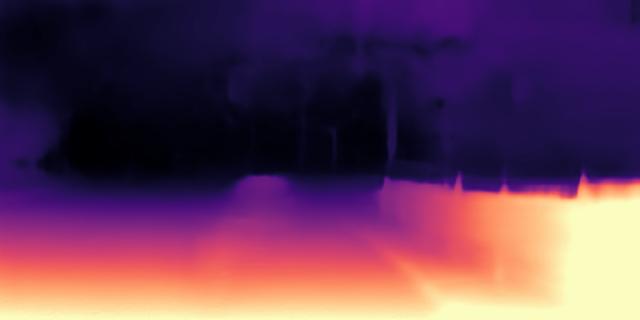} &

    \includegraphics[width=\turnheightnew]{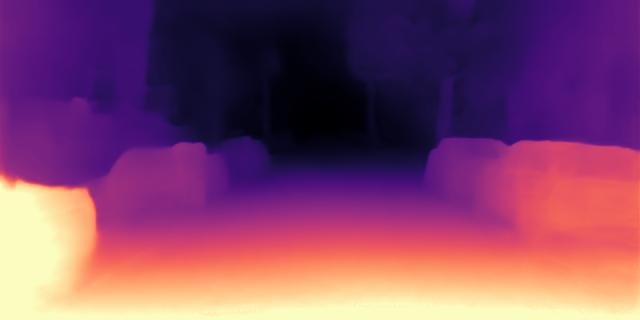} &
    \includegraphics[width=\turnheightnew]{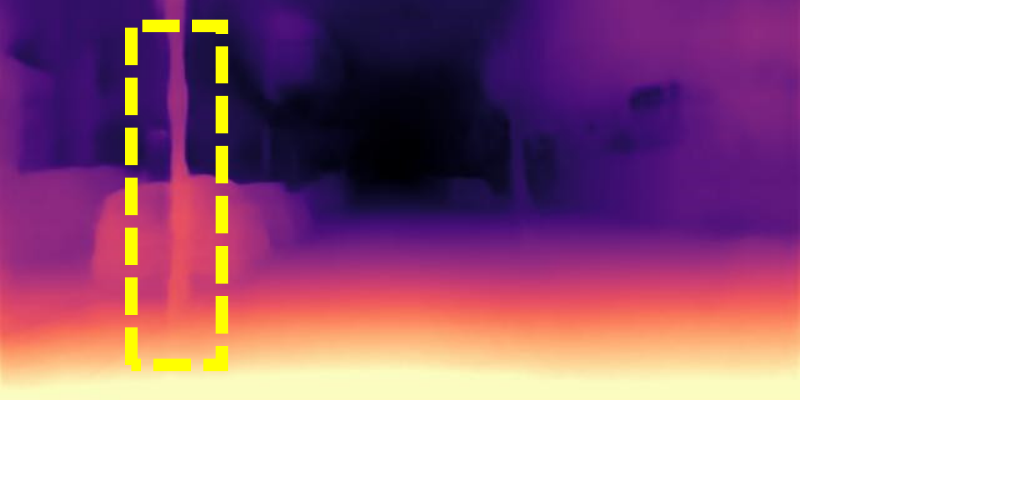} &
    \includegraphics[width=\turnheightnew]{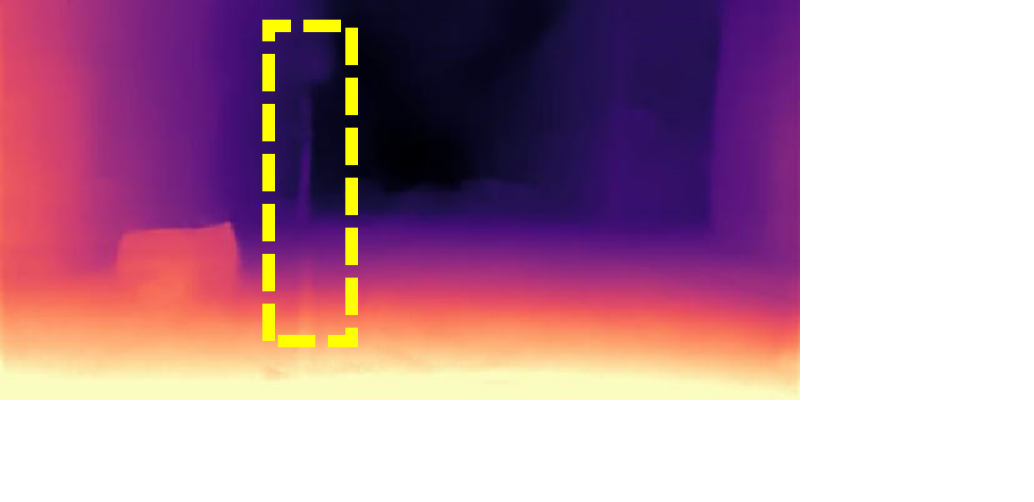} &
    \includegraphics[width=\turnheightnew]{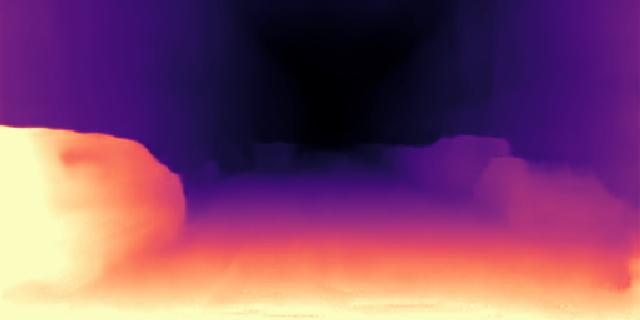} 
    \\

    \vspace{-0.5mm}
    \rotatebox{90}{\hspace{0mm}\scriptsize{G: Ours}}&
    \includegraphics[width=\turnheightnew]{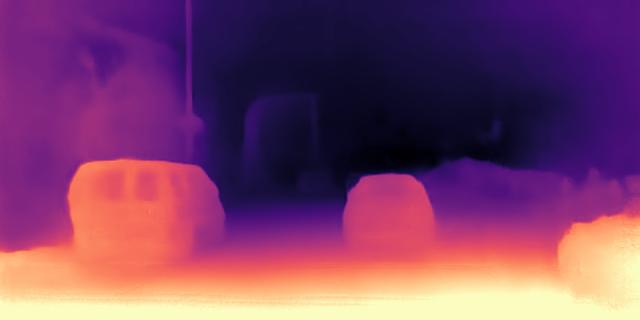} &
    \includegraphics[width=\turnheightnew]{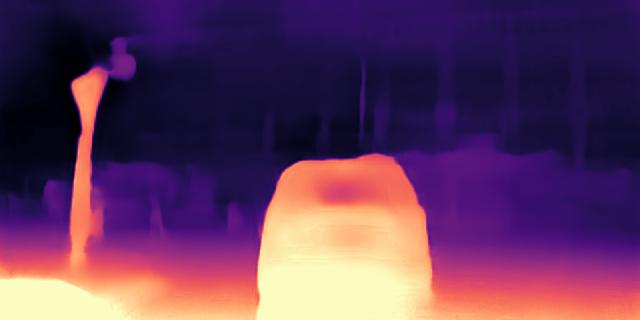} &
    \includegraphics[width=\turnheightnew]{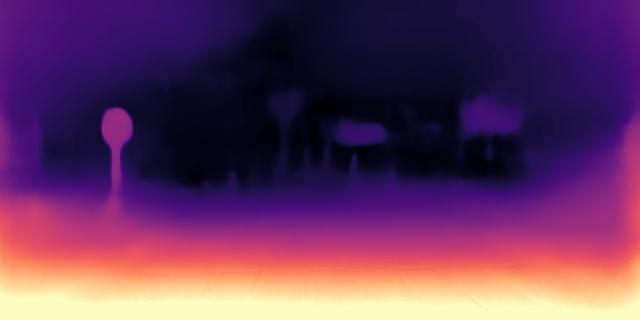} &
    \includegraphics[width=\turnheightnew]{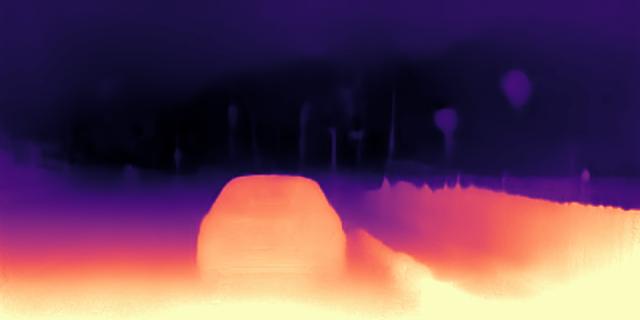} &
    
    \includegraphics[width=\turnheightnew]{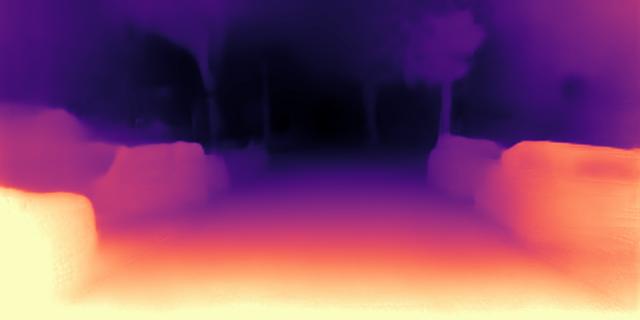} &
    \includegraphics[width=\turnheightnew]{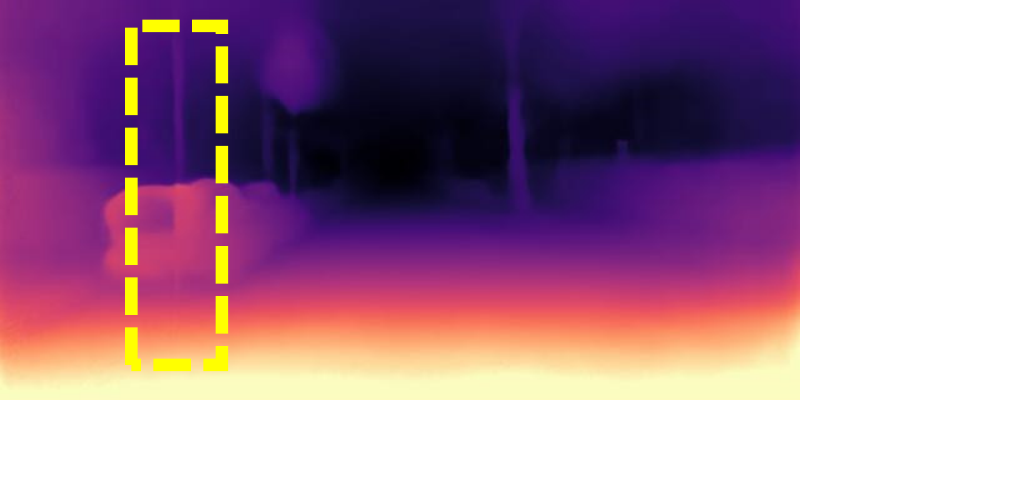} &
    \includegraphics[width=\turnheightnew]{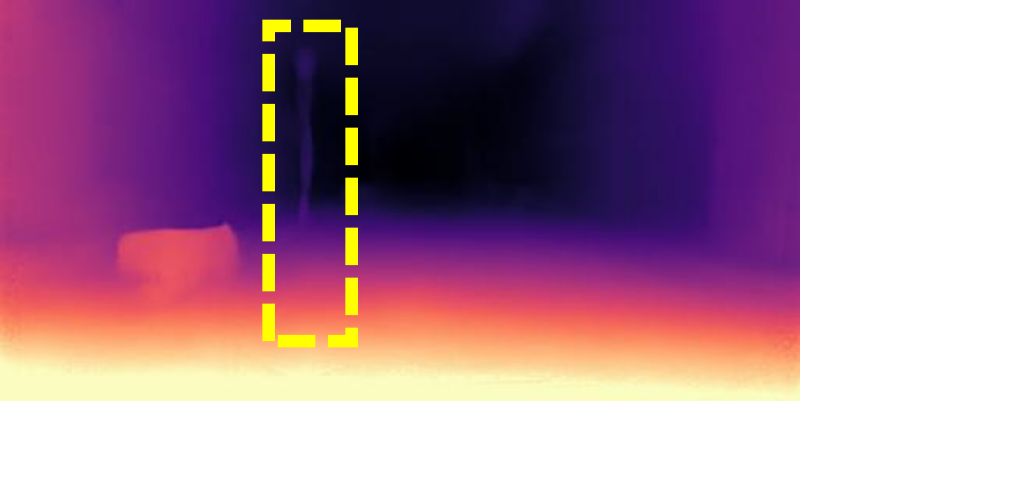} &
    \includegraphics[width=\turnheightnew]{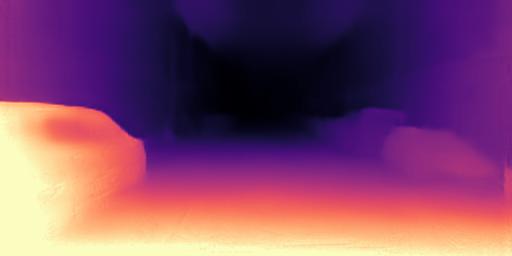}
    \\
    
    \end{tabular}
    \vspace{1mm}
    \caption{\textbf{Qualitative results on nuScenes-Night~(First four columns) and RobotCar-Night~(Last four columns).} 
    % We pick up the most competitive three methods for visualization due to limited space.
    We leave more visual comparisons to the \emph{supplementary material}.
    Compare to DA types methods, our training applies \emph{no images} from the 
    % target datasets.
    {nuScenes or Oxford RobotCar datasets.}
    } 
    \label{fig:vNR}
    \vspace{-2mm}
\end{figure*}

As shown in Fig.~\ref{fig:vNR}, due to the weak interference of ISP on the digit raw, wave optics/diffraction effects are evident, and the imaging noise is also visible. 
Though it provides more realistic imaging compared to that in RobotCar-Night, it introduces further disadvantages in the photometric loss.
Thus, two DT methods strongly diverge on this dataset.

Meanwhile, our method generalizes well to these unseen nighttime scenes and outperforms all DA methods though \emph{no image} from nuScenes dataset is used in our training.
Besides, our method makes a \textbf{13.3\%} improvement on ABS rel and \textbf{17.6\%} improvement on RMSE compared to the second-best approach (DA:ADDS).

As shown in the upper portion of Fig.~\ref{fig:vNR}, our method outperforms the second-best and third-best method by a margin with much fewer prediction outliers. 
In addition, our method provides reasonable depth prediction for both dim scenes and bright scenes. 

\subsubsection{Evaluation on RobotCar-Night}
As shown in Fig.~\ref{fig:vNR}, the strong corrective effect of ISP in RobotCar-Night results in a cleaner image plane compared to that in nuScnce-Night.
The light sources do not image streak and there is little noise in the sky.
Here, we use RobotCar-Night to test the effectiveness of our method when the wave optics effect and noise are not significant. 

The bottom half of Table~\ref{Table:1} shows quantitative results on RobotCar-Night.
Our result still outperforms the second-best method (DA: RNW) by \textbf{15.3\%} in Sq rel and \textbf{8.7\%} in RMSE, despite the reduction in ABS rel.
Given that Sq rel and RMSE is sensitive to outlier points, the comparison suggests that the DepthNet trained by our framework prefers to make fewer prediction outliers rather than take the risk of making potentially more accurate but radical predictions.

In the bottom portion of Fig.~\ref{fig:vNR}, we mark the smearing regions~\cite{smear} that are caused by the CCD~(Charge-coupled-Device) imaging errors instead of the wave optics/diffraction effect.
Our method gives a more accurate prediction in these regions compared to ADDS and RNW, which shows the robustness of our method.

In addition, the comparisons between the top and bottom halves of Fig.~\ref{fig:vNR} and Table~\ref{Table:1} suggest that the performance of DA: ADDS, DA: RNW and DT: WSGD degrades on nuScenes-Night.
It also reveals that domain adaptation-based and photometric loss repair-based methods are still limited by the quality of night images.
Meanwhile, using the same backbone of DepthNet, our method maintains convincing results on both nuScenes-Night and Robotcar-Night, although no image from these two datasets is used during training.

\subsubsection{Comparison of G Results}
In Table~\ref{Table:1}, we also provide the G results of domain adaptation-based methods.
Their decrease in accuracy presents that these methods generalize poorly to unseen night scenes, while our approach does not suffer from such a drawback, and note that no specific nighttime dataset is used as the target dataset during our training.

\begin{figure}[t]
 
    \centering
    \subfigure[ABS rel]{
    \label{Fig:fdat_a}
    \begin{minipage}[c]{0.32\linewidth}
    % \centering
    \includegraphics[width=1in]{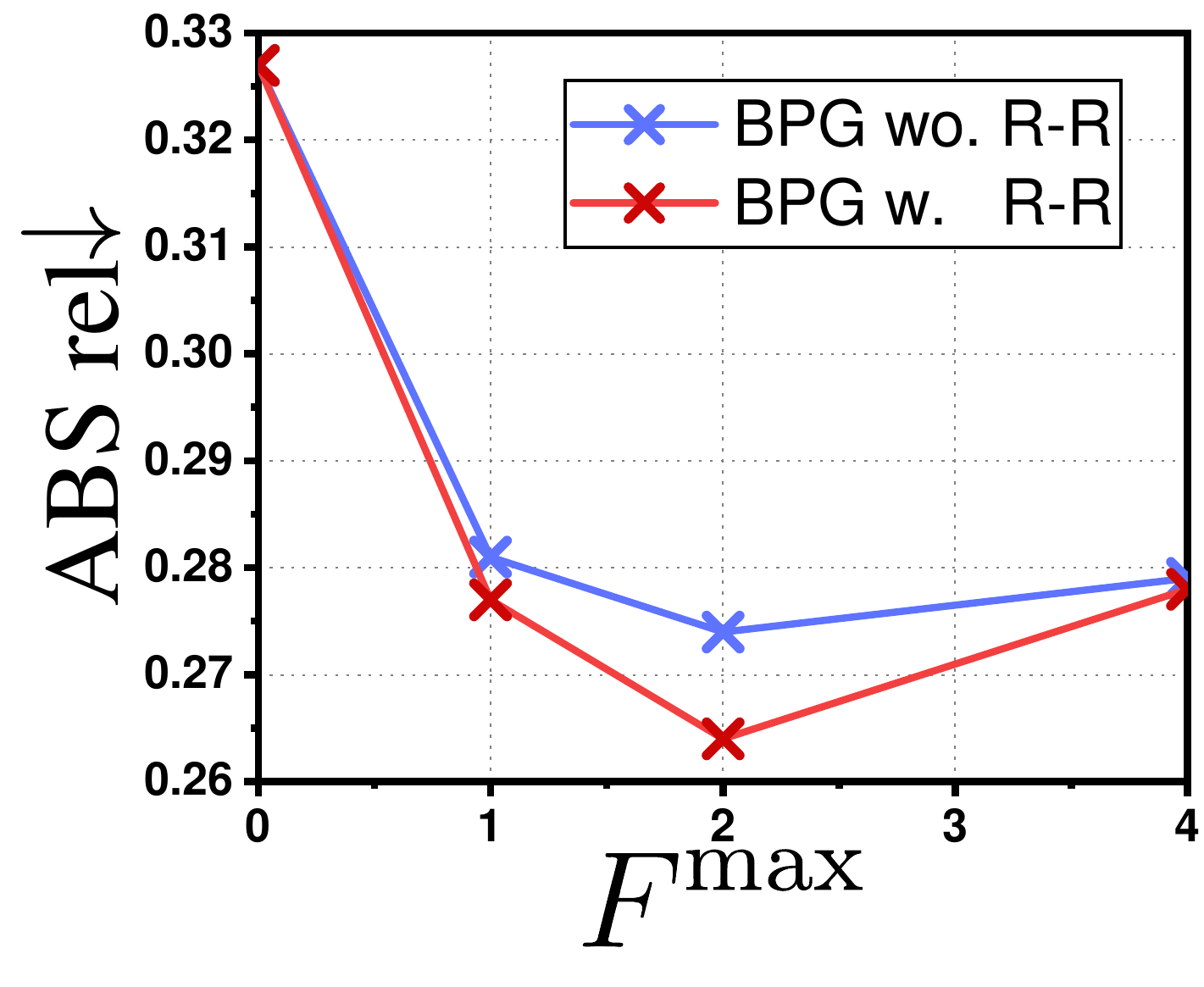}
    
    \end{minipage}%
    }%
    \subfigure[RMSE]{
    \label{Fig:fdat_b}
    \begin{minipage}[c]{0.32\linewidth}
    % \centering
    \includegraphics[width=1in]{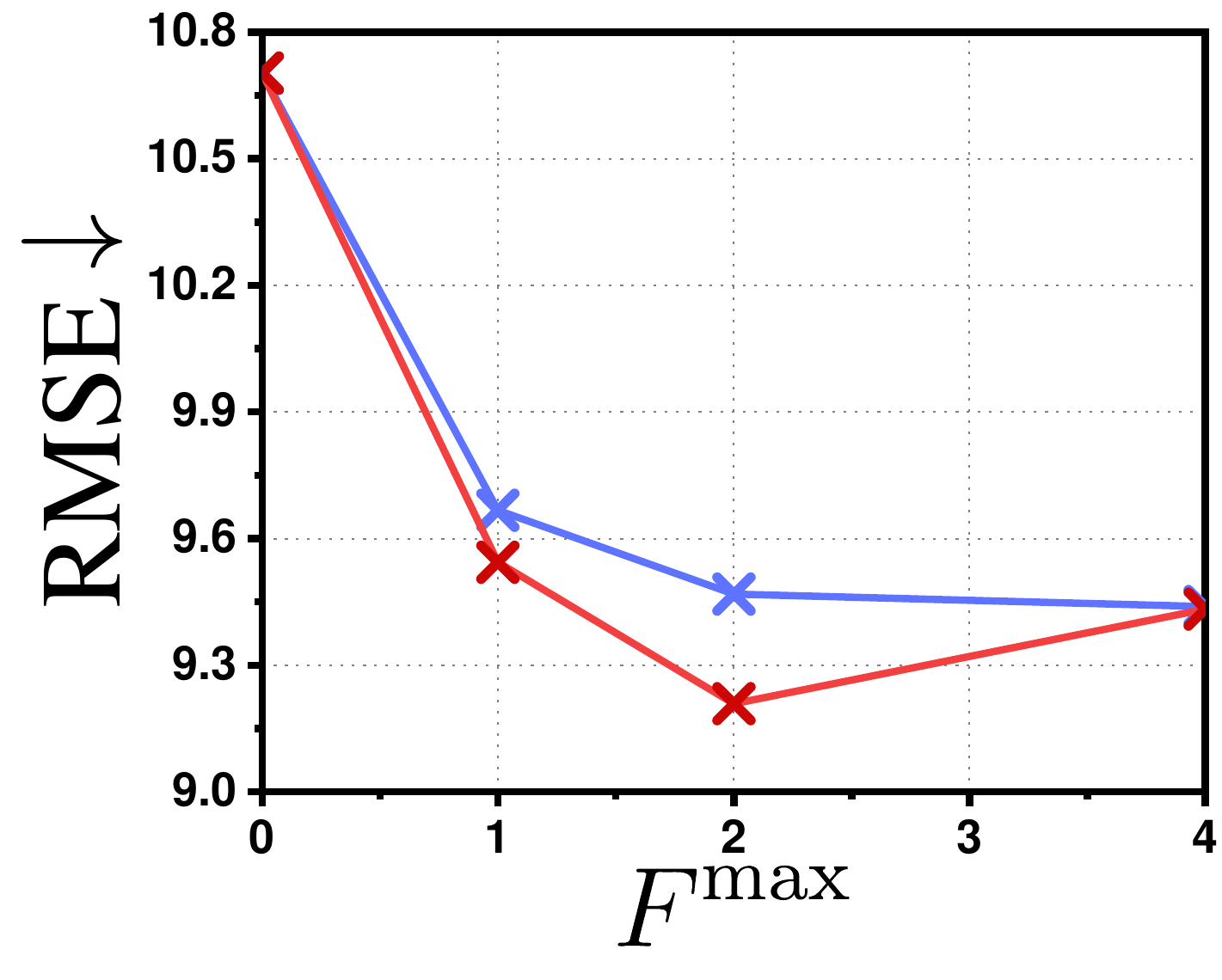}
    \end{minipage}%
    }%
    \subfigure[$ \delta_1 $]{
    \label{Fig:fdat_c}
    \begin{minipage}[c]{0.32\linewidth}
    % \centering
    \includegraphics[width=1in]{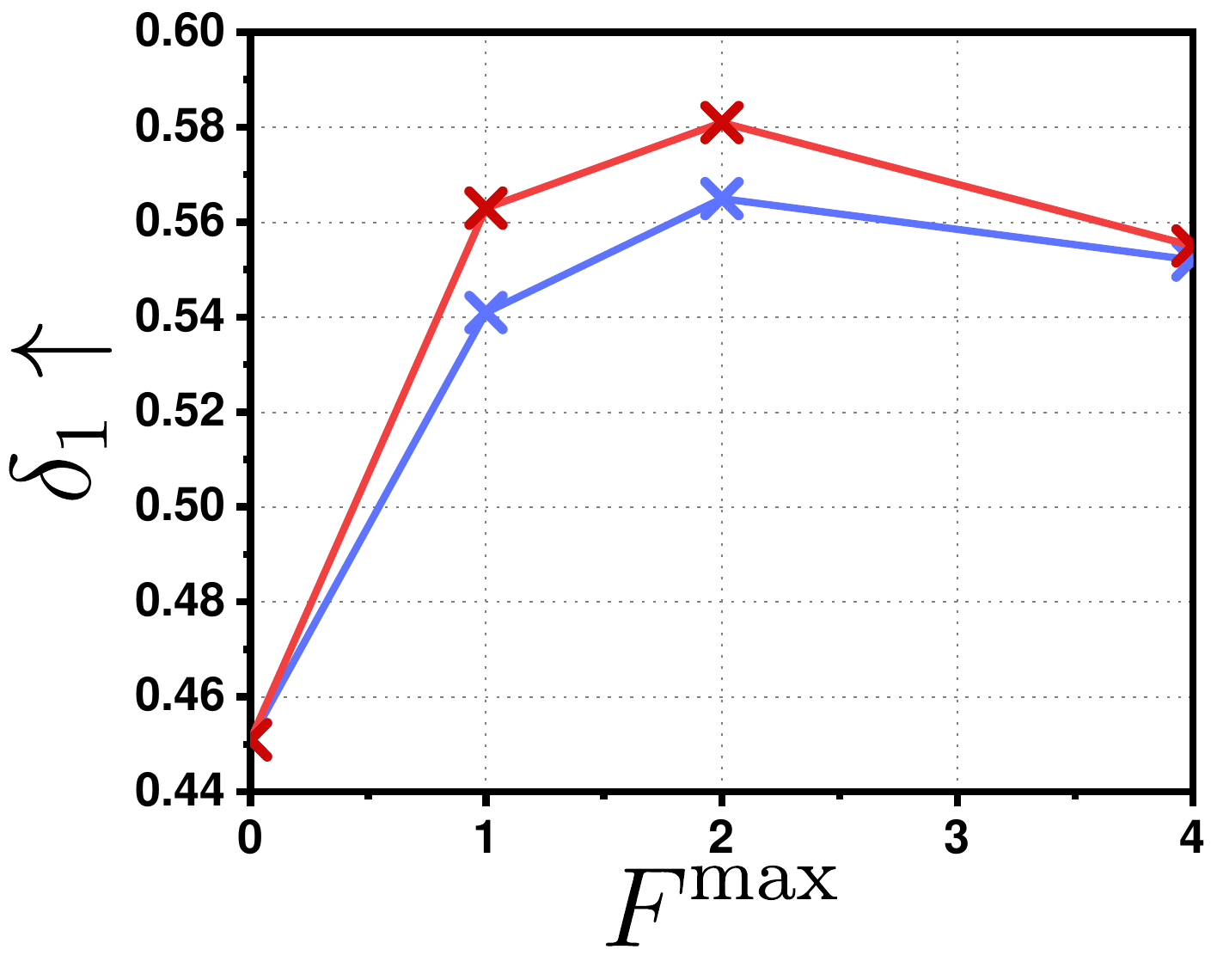}
    \end{minipage}
    }%
    \centering
    \vspace{-3mm}
    \caption{\textbf{Ablation on BPG.} $ F^{\max} = 0$ indicates the baseline. The result of BPG with or without Re-rendering is shown.  
   ~({Best viewed with zoom.})
    }
    \label{Fig:fdat}
\end{figure} 

\begin{figure}[t]
    \centering
    
    \subfigure[ABS rel]{
    \label{Fig:ndat_a}
    \begin{minipage}[t]{0.3\linewidth}
    \centering
    \includegraphics[width=1in]{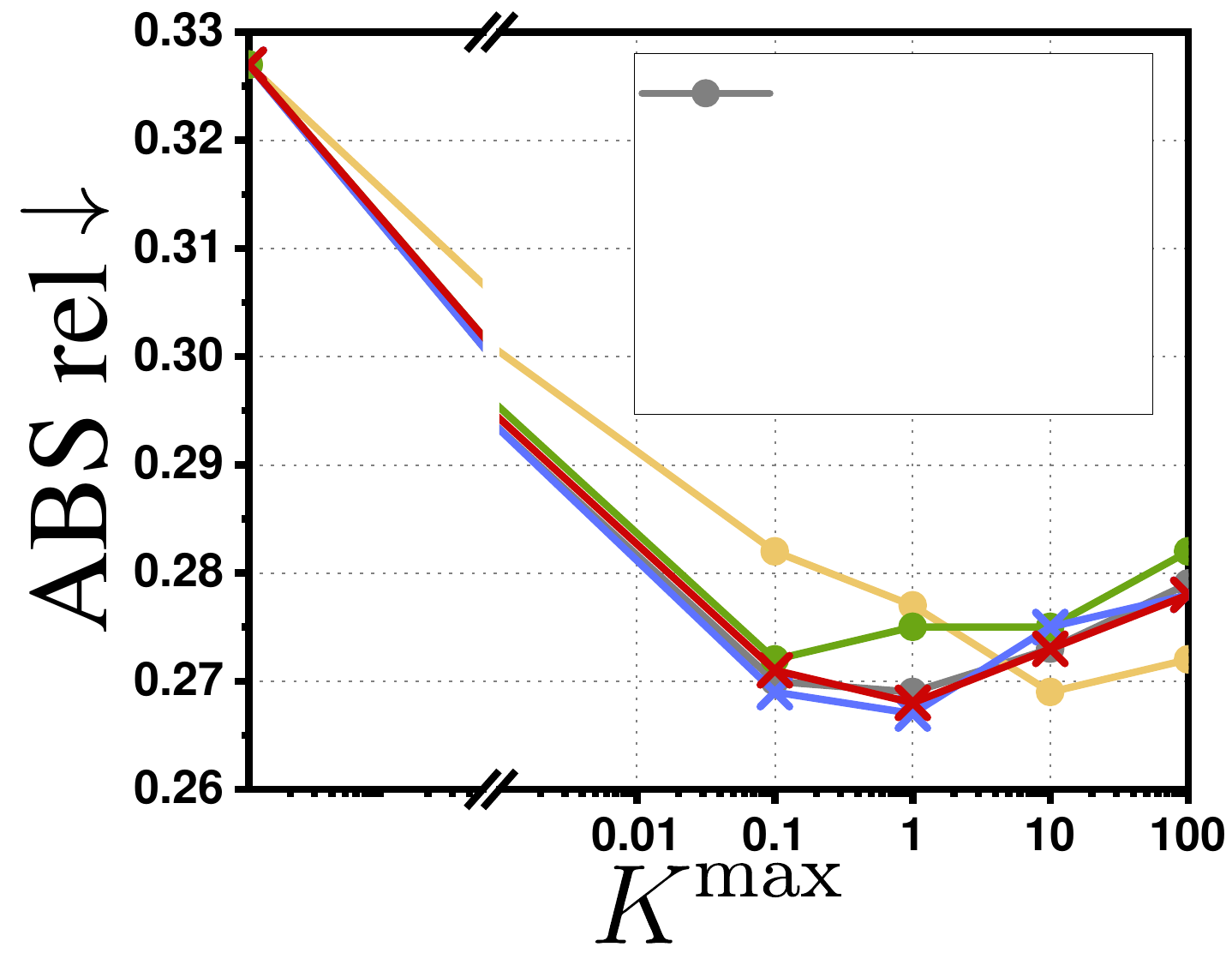}
    \end{minipage}%
    }%
    \hspace{0.015\linewidth}
    \subfigure[RMSE]{
    \label{Fig:ndat_b}
    \begin{minipage}[t]{0.3\linewidth}
    \centering
    \includegraphics[width=1in]{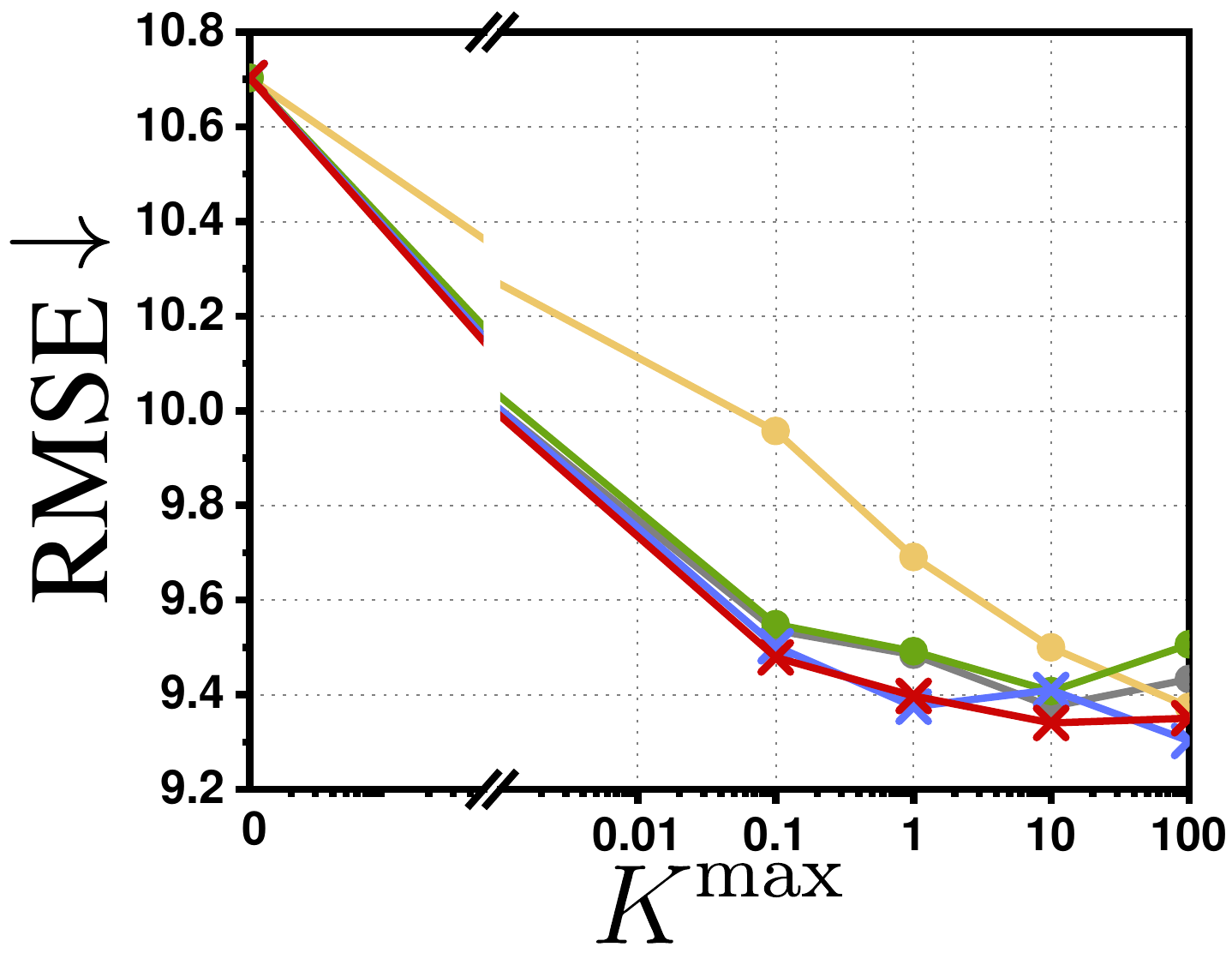}
    \end{minipage}%
    }%
    \hspace{0.015\linewidth}
    \subfigure[$ \delta_1 $]{
    \label{Fig:ndat_c}
    \begin{minipage}[t]{0.3\linewidth}
    \centering
    \includegraphics[width=1in]{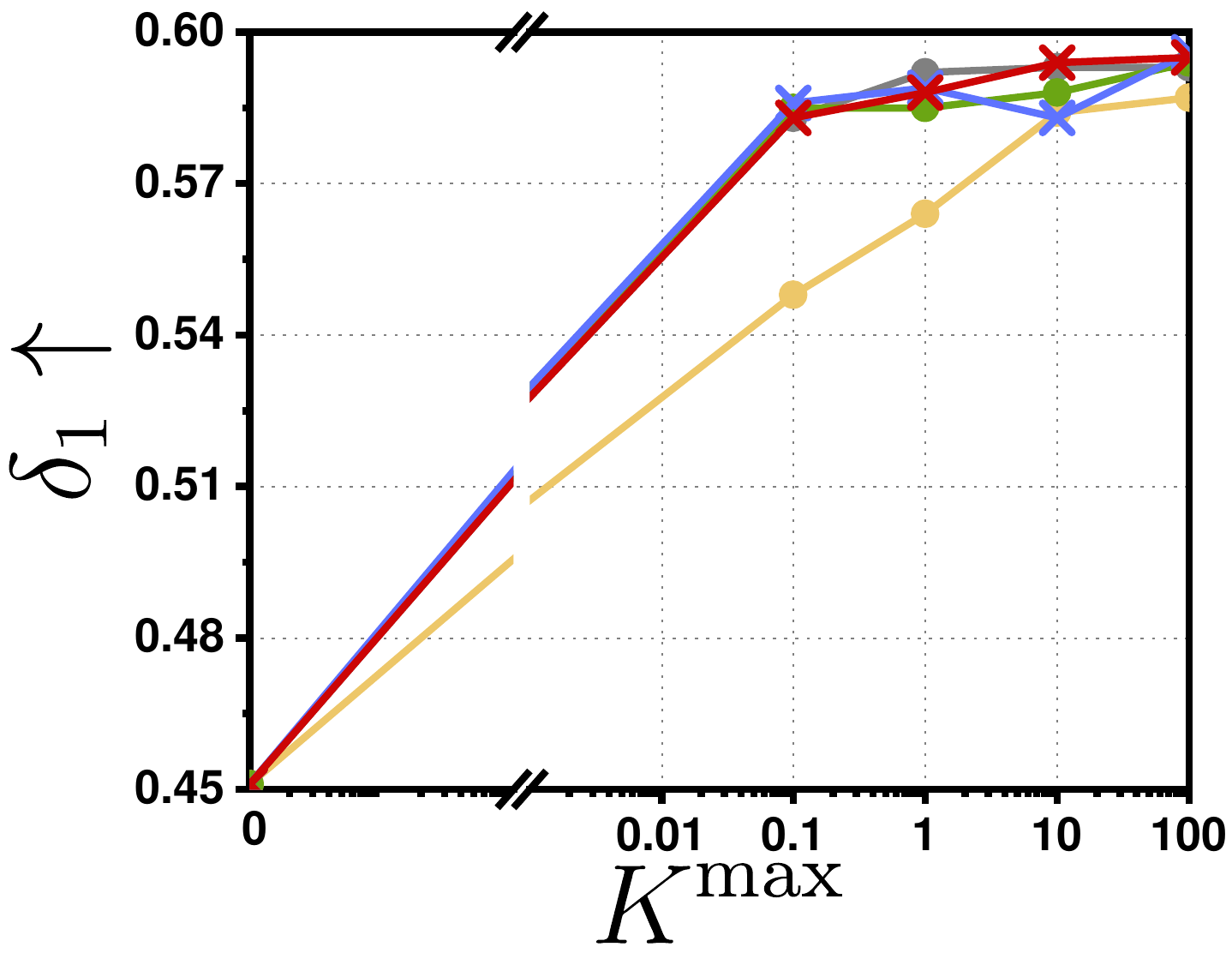}
    \end{minipage}
    }%
    \centering
    \vspace{-4mm}
    \caption{\textbf{Ablation on ING.} $ K^{\max}=0 $ represents the baseline.
    Result of $ \cal N $, $ \cal P $, $ TL $, $ \cal P + N $ and $ {\cal P} + TL$ 
    is shown.
    % (\textbf{Best viewed with zoom.})
    }
    \label{Fig:ndat}
    
\end{figure}

In our supplementary material, we provide additional tests on nuScenes-Day and Robotcar-Day, give a complete qualitative comparison on nuScenes-Night/nuScenes-Day and RobotCar-Night/RobotCar-Day, and offer additional visual results on other nighttime datasets to further prove the effectiveness of our method.

\subsection{Ablation Test on Pre-processing}
\label{Ablation Test}
\subsubsection{Contribution of Each Part}
Table~\ref{Talble:ablation} shows the contributions of each pre-processing part. The comparison of different self-supervised training approaches is left to the supplementary material to further prove the effectiveness of our training framework.
% The application of ICN improves performance as it makes the DepthNet learn better from triplets with uneven light.
% Since ICN is not the main contribution of our work, we view G: MonoViT~w.~ICN as our baseline. 

In nuScenes-Night, compared with the baseline, 
BPG improves ABS rel by 19.3\% and RMSE by 14.0\%, and ING improves ABS rel by 18.0\% and RMSE by 9.1\%.
Besides, the joint application of BPG and ING achieves the best scores with \textbf{20.8\%} improvement on ABS rel, \textbf{20.1\%} on RMSE and \textbf{35.3\%} on $ \delta_{1} $.
For RobotCar-Night, the joint application of BPG and ING still achieves a significant boost with \textbf{13.2\%} on ABS rel and \textbf{9.7\%} on RMSE.

{
In addition, Table~\ref{Talble:ablation} shows that the joint application of BPG and ING significantly improves RMSE~(7.2\% in nuScenec-Night and 8.4\% in RobotCar-Night compared to the second-best), suggesting that our data distribution compensation benefits on the prediction robustness.
}

\subsubsection{Ablating into BPG and ING}
Fig.~\ref{Fig:fdat} visualizes a further ablation study on our BPG.
The experiments show that our Re-rendering submodule plays an important role within BPG. 
Besides, BPG achieves the best result when $ F^{\max} $ is set to 2.  
Meanwhile, Fig.~\ref{Fig:ndat} presents a detailed ablation on our ING.
Apart from the standard groups~($ {\cal P} + TL$ and $ \cal P + N $ ),
we also test $ \cal P$, $ \cal N $ and $ TL $~(Eq.~\ref{equ:position}, Eq.~\ref{equ:logKlogsigma_N} and Eq.~\ref{equ:logKlogsigma_TL}) alone. 
Taken the three metrics shown in Fig.~\ref{Fig:ndat} together, 
$ {\cal P} + TL$ and $ \cal P + N $ still achieve better performance. 
According to Fig.~\ref{Fig:ndat}(b) and Fig.~\ref{Fig:ndat}(c), $ {\cal P} + TL $ maintains the most consistent performance as $ K^{\max} $ changes.
In addition, we set $ K^{\max} $ to 1 as ABS rel increases after $ K^{\max} $ is greater than 1, which suggests that inappropriately high intensity of ING will lead to overly conservative predictions.

\section{Conclusion} 
This paper proposes a self-supervised monocular depth training framework for nighttime, which requires no nighttime image during training but day-to-night data distribution compensation.
Focusing on day-night lighting differences, dissimilarities in photometric and noise distribution are located as two key components. 
We model the difference in corresponding distributions with the proposed BPG and ING.
The samples from the fused distribution are used for the training of the depth network.
Although no nighttime images were used during training, our model shows a more convincing performance than those nighttime frameworks, and our presented self-supervised method provides a new and feasible way for the nighttime monocular depth estimation task.

\section*{Acknowledgment}
This work was supported in part by the National Key Research and Development Program of China under Grant 2021YFB1714300, in part by the National Natural Science Foundation of China under Grant 62233005, in part by the National Natural Science Foundation of China under Grant 62203173, in part by the Program of Introducing Talents of Discipline to Universities through the 111 Project under Grant B17017 and Shanghai AI Lab. 
\bibliographystyle{named}
\bibliography{ijcai24}

\clearpage
\appendix
\section{Comparison of Different Self-supervised Training Methods}
To further discuss the effectiveness and efficiency of our training framework, in this section we compare several self-supervised training methods with our training, including full disturb, pose disturb, and self-teaching. %

\subsubsection{Analysis and Results}
\begin{figure}[htbp]
    \centering
    
    \subfigure[Full disturb.]{
    \begin{minipage}[t]{0.25\linewidth}
    \centering
    \includegraphics[width=1in]{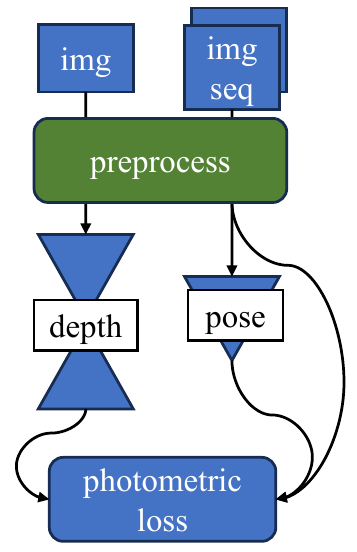}
    \end{minipage}%
    }%
    \hspace{0.2\linewidth}
    \subfigure[Pose disturb.]{
    \begin{minipage}[t]{0.25\linewidth}
    \centering
    \includegraphics[width=1in]{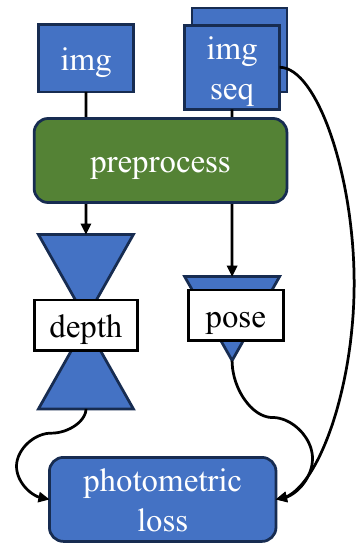}
    \end{minipage}%
    }
    \subfigure[Self-teaching.]{
    \begin{minipage}[t]{0.25\linewidth}
    \centering
    \includegraphics[width=1in]{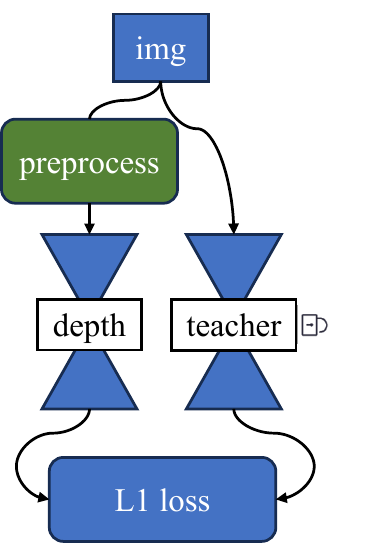}
    \end{minipage}
    }%
    \hspace{0.2\linewidth}
    \subfigure[Our training.]{
    \begin{minipage}[t]{0.25\linewidth}
    \centering
    \includegraphics[width=1in]{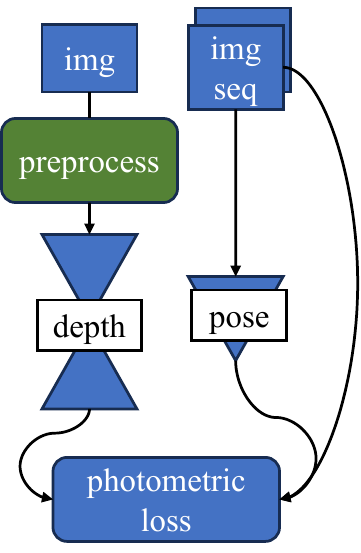}
    \end{minipage}
    }%
    \centering
    \vspace{-4mm}
    \caption{\textbf{Different self-supervised methods with preprocessing.} %
    The Illuminating Change Net (ICN) and the smooth loss are omitted from the figures. %
    }
    \label{Fig:diff_training_method}
    
\end{figure}

In Fig.\ref{Fig:diff_training_method}, we list several methods that use preprocessing (also called compensation in this script) in self-supervised training. %
Considering that the compensated key day-night differences will disturb the photometric loss and the pose network, the results of full disturb and pose disturb are not as good as self-teaching and our training. %
Both self-teaching and our training restrict the day-night differences to the depth network and achieve reasonable results. However, self-teaching is a typical two-step training method that requires about $\times 1.7$ total training time compared to our training, indicating that our training is a more efficient choice. %

\begin{table}[htbp]
    \centering
        \resizebox{0.48 \textwidth}{!}{
        \begin{tabular}{c | ccc | ccc | c}
            \hline
            Method & \cellcolor{cell1} ABS rel $\downarrow$ & \cellcolor{cell1} Sq rel $\downarrow$  & \cellcolor{cell1} RMSE $\downarrow$  & \cellcolor{cell2} $\delta_{1} \uparrow$ & \cellcolor{cell2} $\delta_{2} \uparrow$ & \cellcolor{cell2} $\delta_{3} \uparrow$ & Time\\
            \hline
            Full disturb   & 0.601     	        & 10.06     	   & 15.76    	      & 0.247     	    & 0.447     	    & 0.610            & $\times1.0$\\
            Pose disturb   & \underline{}{0.276}& 3.606     	   & 9.098     	      & \underline{0.625}& 0.840     	    & \underline{0.928}& $\times1.0$\\
            Self-teaching  & \textbf{0.259}     & \textbf{3.139}   & \underline{8.615}& \textbf{0.641}   & \underline{0.847}& \textbf{0.930}   & $\times1.7$\\
            Our training  & \textbf{0.259}     & \underline{3.147}& \textbf{8.547}   & \textbf{0.641}   & \textbf{0.850}   & \underline{0.928}& $\times1.0$\\
            \hline
        \end{tabular}
        }
        \caption{\textbf{Result of different self-supervised methods.} %
        K, N-D and N-N denote the KITTI, nuScenes-Day and nuScnes-Night training sets, respcetively. %
        Time means the total training time compared to our training. %
        All methods are trained on a single A60.
        \vspace{-1mm}
        }
    \label{Table:diff_training_method}

\end{table}

\subsubsection{Experiment Setting} 
In the training, all experiments in this part use our BPG and ING for preprocessing. %
Full disturb and pose disturb apply the same training settings as mentioned in our main paper~(see in Sec.~\ref{Sec:training_detail}) except from different pose network and photometric loss input. %

In self-teaching~(Fig.~\ref{Fig:diff_training_method}), the teacher network is trained on KITTI dataset as mentioned by MonoViT~\cite{monovit}. %
Besides, we apply l1 loss as the distilling loss follow HR-depth~\cite{hr-depth}. %

\section{Why Our Partial Compensation Work in Self-supervised Mono-depth Generalization}
\begin{figure}[htbp]
    \centering
    
    \subfigure[Attack on smooth area.]{
    \begin{minipage}[t]{\linewidth}
    \centering
    \includegraphics[width=\linewidth]{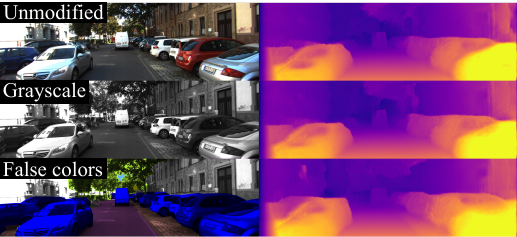}
    \end{minipage}%
    }

    \subfigure[Edge.]{
    \begin{minipage}[t]{\linewidth}
    \centering
    \includegraphics[width=\linewidth]{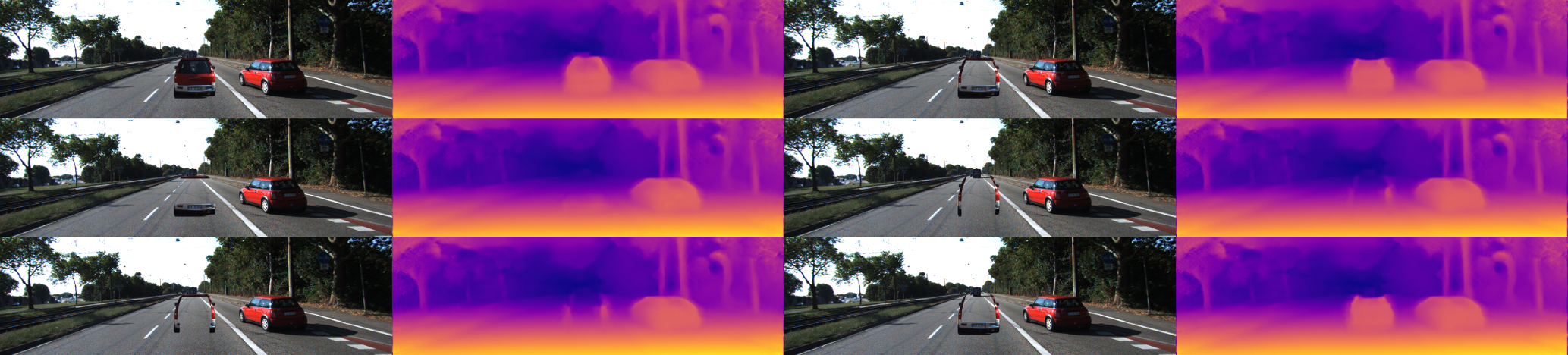}
    \end{minipage}
    }%
    \caption{\textbf{Generation ability of monocular depth network on smooth area and edges.} %
    These figures are taken directly from \protect\cite{what_monodepth_learn}. %
    All of the above results are generated by a single depth network trained on the normal KITTI dataset. %
    \vspace{-1mm}
    }
    \label{Fig:sm_and_edge}
    
\end{figure}
Though our partial compensation (BPG and ING) will \textbf{not} produce vivid night images, the depth network trained by our method shows good day-to-night generation ability. %  
In our main paper, we have explained this problem in the terms of day-night differences. %
In this section, we further discuss it in terms of the characteristics of monocular depth estimation. %
We also compare our partial compensation with other full day-night compensations, such as CUT~\cite{CUT} and Qs-Atten~\cite{QS_atten}, which are commonly used in image transfer after CycleGAN~\cite{cyclegan}.

\subsubsection{Analysis and Results}
As discussed in \cite{what_monodepth_learn}, unlike other semantic related tasks, such as segmentation and detectionm, monocular depth is not sensitive to color transformations in smooth regions of the image as shown in Fig.~\ref{Fig:sm_and_edge}(a). %
However, the integrity of the object edges has significant impact on monocular depth estimation, as shown in Fig.~\ref{Fig:sm_and_edge}(b). %
High-frequency artifacts, such as hightlights and noise in night scenes, can corrupt the depth-dependent high-frequency part of the image (i.e., object edges). %
Thus, the partial compensation of hightlights and noise (BPG and ING) allows the depth network to suppress these depth-uncorrelated disturbing terms and better generalize to the challenging nighttime scenes. %

Table~\ref{Table:partial_full} shows the comparison between our compensation and other full compensation (image transfer). %
Although CUT and Qs-Atten get better FID score on the nuScenes-Night testing set, indicating that their methods produce more night-like images, our methods still get better night depth estimation results. %
Compared to the second best, we improve ABS rel by 9.8\%  and RMSE by 9.8\%. %

\begin{table}[htbp]
    \centering
        \resizebox{0.48 \textwidth}{!}{
        \begin{tabular}{l | ccc | ccc | c}
            \hline
            Compensation method & \cellcolor{cell1} ABS rel $\downarrow$ & \cellcolor{cell1} Sq rel $\downarrow$  & \cellcolor{cell1} RMSE $\downarrow$  & \cellcolor{cell2} $\delta_{1} \uparrow$ & \cellcolor{cell2} $\delta_{2} \uparrow$ & \cellcolor{cell2} $\delta_{3} \uparrow$ & FID $\downarrow$\\
            \hline
            \multicolumn{8}{c}{\cellcolor{cell3} Train on KITTI - Test on nuScenes-Night} \\
            None                               & 0.327     	        & 3.740     	   & 10.70    	      & 0.451     	     & 0.733     	    & 0.855          & -\\
            Cut I2I: K to N-N                  & 0.300              & 3.771            & 9.596            & 0.565            & 0.793            & 0.895          & 46.26\\
            Qs-Atten I2I: K to N-N             & 0.291              & 3.659     	   & 9.554     	      & 0.577            & 0.801     	    & 0.903          & \textbf{45.05}\\
            \textbf{Ours compensation}         &\textbf{0.259}      & \textbf{3.147}   & \textbf{8.547}   & \textbf{0.641}   & \textbf{0.850}   & \textbf{0.928} & 60.01$\pm$1.81\\
            \hline
            \multicolumn{8}{c}{\cellcolor{cell3} Train on nuScenes - Test on nuScenes-Night} \\
            Cut I2I: N-D to N-N                & 0.290     	        & 3.653     	   & 9.614     	      & 0.580     	     & 0.807     	    & 0.911          & 54.40\\
            Qs-Atten I2I: N-D to N-N           & 0.287     	        & 3.582     	   & 9.473     	      & 0.580     	     & 0.813     	    & 0.913          & 46.97\\
            \hline
        \end{tabular}
        }
        \caption{\textbf{Ours result compared to the full compensation methods}.%
         K, N-D and N-N denote the KITTI, nuScenes-Day and nuScnes-Night training sets, respectively. %
         We append the poplar FID metric~\protect\cite{FID_source} to measure the day-to-night quality of the day-to-night full compensation. %
        \vspace{-1mm}
        }
    \label{Table:partial_full}
    \vspace{-4mm}
\end{table}

\begin{figure}[htbp]
    \centering
    
    \subfigure[Full compensation results.]{
    \begin{minipage}[t]{\linewidth}
    \centering
    \includegraphics[width=\linewidth]{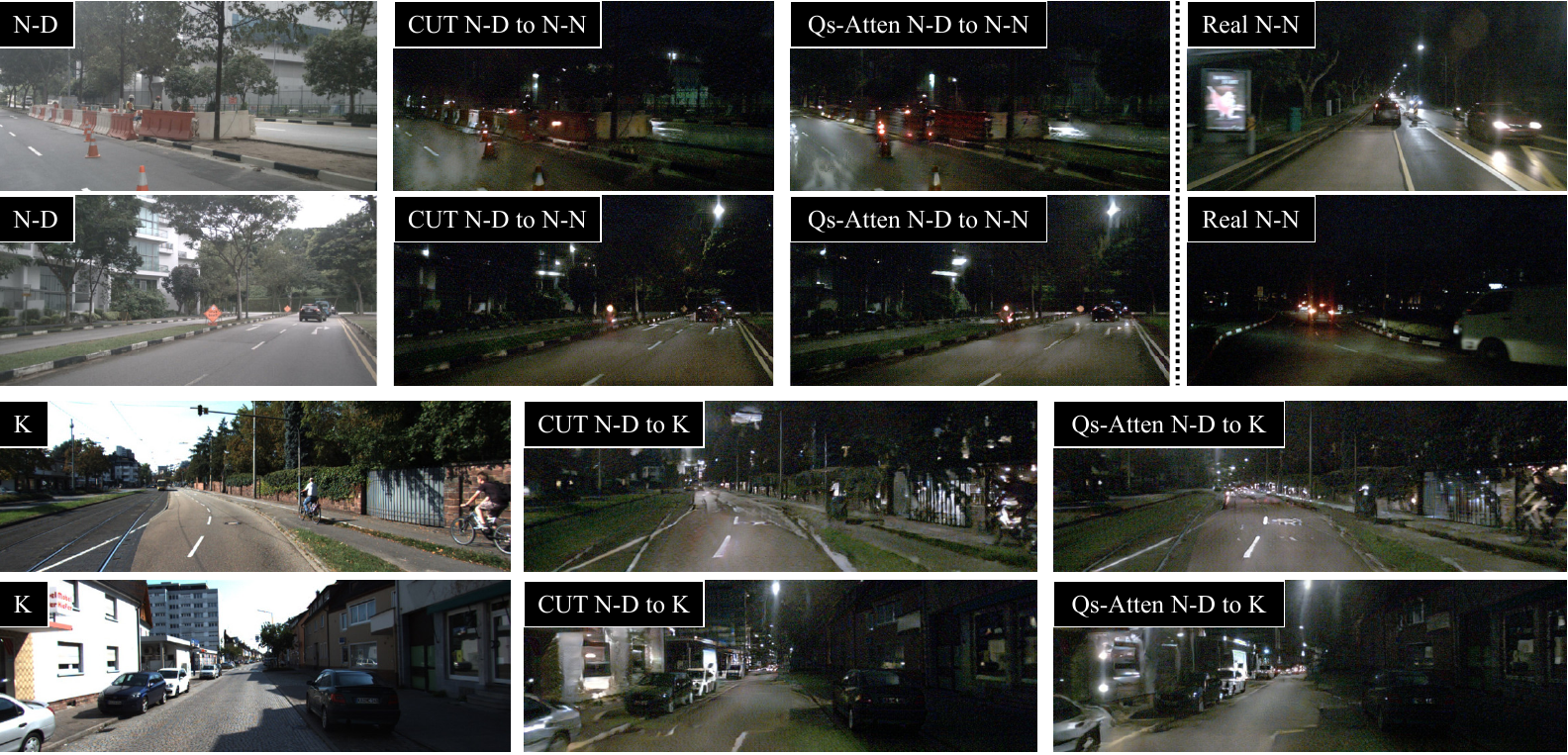}
    \end{minipage}%
    }

    \subfigure[Our partial compensation results.]{
    \begin{minipage}[t]{\linewidth}
    \centering
    \includegraphics[width=\linewidth]{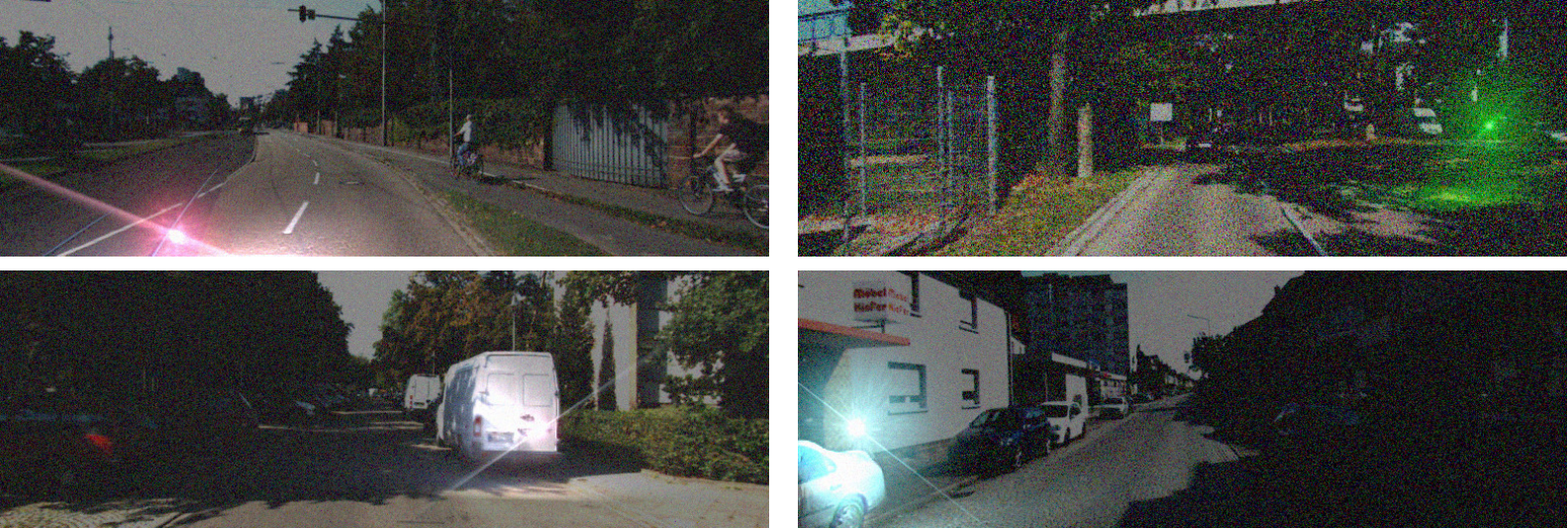}
    \end{minipage}
    }%
    \caption{\textbf{Visualization of full compensation and our partial compensation results.} %
    \vspace{-1mm}
    }
    \label{Fig:sm_and_edge}
    
\end{figure}

\subsubsection{Day-to-Night Image Transfer Experiment Setting} 
To train the day-to-night image transfer network, we randomly select 4,000 images from the KITTI~\cite{kitti_eigen}, nuScene-Day~\cite{rnw} and nuScenes-Night~\cite{rnw} monocular depth training sets, respectively, as the I2I~(image to image transfer) training set. % 
Note that the KITTI dataset contains daytime images only. % 
The I2I testing set is the corresponding monocular depth testing set. %
As in our main paper, all images in nuScenes use the pre-crop mentioned in RNW~\cite{rnw}. %

In the K to N-N image transfer training, the images from KITTI dataset are first resized to 288$\times$864 and then randomly cropped to 256$\times$768. %
The images from nuScenes-Night dataset are center cropped to 512$\times$1536, resized to 288$\times$864 and then randomly cropped to 256$\times$768 to match the resolution of KITTI images.

In the N-D to N-N image transfer training, both the images from nuScenes-Day and nuScenes-Night are resized to 384$\times$768 and then randomly cropped to 320$\times$640.

The total training epoch is set to 200. % 
And it takes about 5 days on a single A60 to train Qs-Atten. % 
Other training settings are the same as for CUT and Qs-Atten. %
Considering the randomization of our partial compensation (BPG and ING), we test FID 100 times and take the average. %

\subsubsection{Monocular Depth Experiment Setting}
All depth networks in Table~\ref{Table:partial_full} are trained with the self-supervised method mentioned in Fig.~\ref{Fig:diff_training_method}(d), that is our training. %
We use 50\% as the preprocessing rate for CUT I2I and Qs-Atten I2I, which gives a slightly better result than setting it to 100\%. %
Other settings are the same as for the main paper~(see in Sec.\ref{Sec:training_detail}).

\section{Limitation}
\subsubsection{Analysis and Results}
Although the proposed compensation method (BPG+ING) works well for day-to-night self-supervised monocular depth generation, the corresponding supervised generation remains unsolved. % 

In the KITTI dataset, as in all other self-driving outdoor datasets, the depth ground truth is sparse. %
And at the far end of the image (e.g., the sky) there is always no ground truth. %
This sparsity of ground truth could break the representational ability of the depth network encoder, resulting in worse day-to-night generation. %

As shown in Table \ref{Table:limitation}, similar to two full compensation methods (image transfer), our compensation gets better results in self-supervised training than in supervised training. %
Note that although all methods get unsatisfactory results, our method still remains the best. %
\begin{table}[htbp]
    \centering
        \resizebox{0.48 \textwidth}{!}{
        \begin{tabular}{l | ccc | ccc}
            \hline
            Compensation method & \cellcolor{cell1} ABS rel $\downarrow$ & \cellcolor{cell1} Sq rel $\downarrow$  & \cellcolor{cell1} RMSE $\downarrow$  & \cellcolor{cell2} $\delta_{1} \uparrow$ & \cellcolor{cell2} $\delta_{2} \uparrow$ & \cellcolor{cell2} $\delta_{3} \uparrow$\\
            \hline
            \multicolumn{7}{c}{\cellcolor{cell3} \textbf{Self-supervised} train w. compensation method on KITTI - Test on nuScenes-Night} \\
            None                               & 0.327     	        & 3.740     	   & 10.70    	      & 0.451     	     & 0.733     	    & 0.855\\
            CUT I2I: K to N-N                  & 0.300              & 3.771            & 9.596            & 0.565            & 0.793            & 0.895\\
            Qs-Atten I2I: K to N-N             & 0.291              & 3.659     	   & 9.554     	      & 0.577            & 0.801     	    & 0.903\\
            Ours compensation                  &\textbf{0.259}      & \textbf{3.147}   & \textbf{8.547}   & \textbf{0.641}   & \textbf{0.850}   & \textbf{0.928}\\
            \hline
            \multicolumn{7}{c}{\cellcolor{cell3} \textbf{Supervised} train w. compensation method on KITTI - Test on nuScenes-Night} \\
            None                               & 0.381     	        & 4.893     	   & 11.47    	      & 0.407     	     & 0.703     	    & 0.843\\
            CUT I2I: K to N-N                  & 0.367              & 4.594            & 11.42            & 0.419            & 0.702            & 0.834\\
            Qs-Atten I2I: K to N-N             & 0.353              & 4.298     	   & 11.11     	      & 0.428            & 0.721     	    & 0.847\\
            Ours compensation                  & \textbf{0.333}     & \textbf{3.860}   & \textbf{10.74}   & \textbf{0.452}   & \textbf{0.733}   & \textbf{0.858}\\
            \hline
        \end{tabular}
        }
        \caption{\textbf{Result of supervised and self-supervised methods with different compensation on nuScenes-Night.} %
        None means training directly on KITTI without compensation method.
        \vspace{-1mm}
        }
    \label{Table:limitation}
    \vspace{-4mm}
\end{table}

\subsubsection{Supervised Experiment Setting}
In the supervised training, we apply the popularly used Scale-Invariant loss~\cite{kitti_eigen,adabins,NeWCRFs,AAE}~(SILog) to supervise the monocular depth network. %
The training step is set to 80,000 step. %
Other training settings are the same as to the main paper (Sec.~\ref{Sec:training_detail}). %
The median scale is not used in their testing as it will make the result even worse. %

We do not provide supervised results using nuScenes as the training set because it provides a lower ground truth density than the KITTI dataset, typically less than 1\%. %
(Unlike the KITTI dataset, nuScenes uses the raw 3D LiDAR output as the depth ground truth, which is be more accurate but less dense.)

\section{Other Implement Details}

\subsection{Details of Self-supervised Loss}
In self-supervised monocular depth estimation, we cast the training problem as an image reconstruction 
task. By approaching the estimated depth $ D_{t} $ and the related pose $ T_{t,t+n} $, we recover a target 
frame $ I_{t} $ from a source frame $ I_{t+n} $ with differentiable bilinear sampling~\cite{bisampler} 
operator $ \mathtt{bs}(.,.) $. 
Following \cite{alldaydemono}, the potential illuminating change between $ I_{t} $ and $ I_{t+n} $
are also considered during the training. We use a sequential feature decoder~(ICN) to estimate the linear brightness
change $ \{C_{t,t+n}\in{\mathbb{R}^{H\times{W\times{1}}}},B_{t,t+n}\in{\mathbb{R}^{H\times{W\times{1}}}}\} $. 
The reconstruction of $ I_{t} $ can be described as
\begin{equation}
    \begin{split}
        & \dot{V}_{t,t+n}(u) = K_I T_{t,t+n}D_{t}(u)K_I^{-1}\dot{u}, \\
        & {\tilde{I}}_{t+n} = \mathtt{bs}(I_{t+n}\odot{C_{t,t+n}}+B_{t,t+n},V_{t,t+n}),
    \end{split}        
\end{equation}
with $ u $ being a pixel in frame $ I_{t} $, $ \dot{u} $ being the corresponding homogeneous form,
and $ \tilde{I}_{t+n} $ being the reconstructed version of target frame $ I_{t} $.  
We assume the camera intrinsic $ K_I\in\mathbb{R}^{3\times3} $ is already known during the training.

Following~\cite{alldaydemono,monodepth2,rnw}, we choose the combination of Structured Similarity 
Index Measure (SSIM)~\cite{ssim} and the L1 difference as the measure of photometric error, i.e.
\begin{equation}
    \label{equ:pe}
    \mathtt{pe}(\tilde{I}, I)=\frac{\alpha}{2}(1-\mathtt{SSIM}(\tilde{I},I)) + (1-\alpha)|\tilde{I}-I|, \\
\end{equation}
where we set $ \alpha $ to 0.85 in all our experiments. 

To handle potential occlusions within image sequences, we use the popular
minimum re-projection and Auto-mask strategies~\cite{monodepth2} in the per-pixel photometric loss:
\begin{equation}
    L_{ss} = \min_{\tilde{I}}\mathtt{pe}(\tilde{I},I_{t}),    
\end{equation}
with $ \tilde{I} \in \{ \tilde{I}_{t+n}, \tilde{I}_{t-n}, I_{t+n}, I_{t-n} \} $.

In addition, to maintain the spatial smoothness of the estimation, we apply
the edge-aware gradient smoothing loss $ L_{g} $~\cite{monodepth2}:
\begin{equation}
    \label{equ:sm}
    L_{g}=|\partial_{x}{d^{*}}|e^{-|\partial_{x}{I_{t}}|}+
    |\partial_{y}{d^{*}}|e^{-|\partial_{y}{I_{t}}|},
\end{equation}
with $ d^{*} = d/\hat{d} $ being the inverse normalized depth map.   

Finally, the output at each scale level 
$ l\in\{1,\frac{1}{2},\frac{1}{4},\frac{1}{8}\} $ will be used to compute $ L_{ss} $ and $ L_{g} $. 
We use the averaged summation of them as the final total loss, i.e.
\begin{equation} \label{equ:9}
    L_{total}=\frac{1}{4}\sum\limits_{l}(L_{ss}+\lambda\cdot{L_{g}}),
\end{equation}   
where $ \lambda $ is set to $ 1e^{-3} $.

\subsection{Details of Re-rendering}
In this part, we detail the calculation of reflection image $ I_i^{R} $, including surface normal, coarse material of objects and application of Phong model~\cite{phong}.  

\subsubsection{Surface Normal.}
\begin{figure}[htbp]
    \vspace{-3mm}
    \centering
    
    \subfigure[w. $ \mathbf{N}(u) $.]{
    \label{Fig:SN_a}
    \begin{minipage}[t]{0.33\linewidth}
    \centering
    \includegraphics[width=1in]{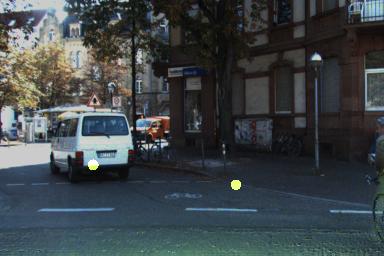}
    
    \end{minipage}%
    }%
    \subfigure[w. $ \mathbf{N}(u) $.]{
    \label{Fig:SN_b}
    \begin{minipage}[t]{0.33\linewidth}
    \centering
    \includegraphics[width=1in]{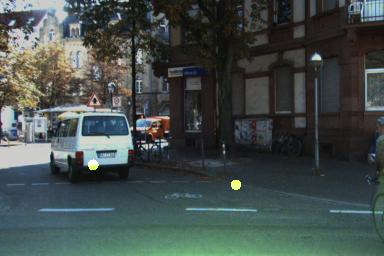}
    \end{minipage}%
    }%
    \subfigure[Verters .]{
    \label{Fig:SN_c}
    \begin{minipage}[t]{0.33\linewidth}
    \centering
    \includegraphics[width=1in]{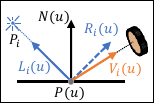}
    \end{minipage}
    }%
    \centering
    \caption{Re-rendering comparison of different surface normal methods and the visualized
    definition of vectors used in Re-rendering submodule. Compare to (a), (b) shows
    a more smooth result. The viewpoint in (c) is set to the original point.}
    \label{Fig:SN}
    \vspace{-1.5mm}
\end{figure}

Most rendering models require surface normal, following~\cite{depth2nv}, we directly apply the depth map $ D'_t $ to calculate the surface normal, i.e.
\begin{equation} \label{equ:surface normal}
    \mathbf{N}(u)=\mathtt{norm}\left(\left[-s \partial_x D'_t(u),
    -s \partial_y D'_t(u),
    1
    \right]^T\right),
\end{equation}
with $ \mathtt{norm}(.) $ being vector L2 normalization. 
We find that Eq.~\ref{equ:surface normal} achieves better results of re-rendering than that applying vector fork product on a neighborhood~\cite{depth_scale_recovery}. 
Fig.~\ref{Fig:SN}(a) and Fig.~\ref{Fig:SN}(b) show the comparison.

Then the corresponding distance $ r_i(u) $, inverse incidence vector $ \mathbf{L}_i(u) $, 
reflection vector $ \mathbf{R}_i(u) $, which is shown in Fig.~\ref{Fig:SN}(c), refer to:
\begin{equation}
    \begin{split}
        & r_i(u) =  \mathtt{dist}(P_i, P(u)),\\
        & \mathbf{L}_i(u)=\mathtt{norm}(P_i-P(u)),\\
        & \mathbf{R}_i(u)= \mathtt{norm}(2\left(\mathbf{L}_i(u)*\mathbf{N}(u)\right) \cdot \mathbf{N}(u)-\mathbf{L}_i(u)).
    \end{split}        
\end{equation}
where $ \mathtt{dist}(.,.) $ represents the Euclidean distance between two points.

\subsubsection{Coarse Material.}In Re-rendering, the material of objects is another essential. 
We assume that the original daytime image $ I_t $ is well-lit, and the 
RGB values can represent the material information to a certain extent. 
Thus, we calculate diffusion factor $ K_d $ and 
reflection factor $ K_s $ with
\begin{equation}
    \label{equ:texture}
    \begin{split}
        & I_{t}^{p} = \text{avgpool}(I_t),\\
        & K_d(u) = \frac{k_d I_{t}^p(u)}{\max\limits_{i\in\{r,g,b\}} I_{t,i}^p(u)+e_{ps}},\\
        & K_s(u) = \frac{k_s}{3}\sum_{i\in\{r,g,b\}}I_{t,i}^p(u),    
    \end{split}    
\end{equation}
where kernel size of avg pooling is set to 3, and $ e_{ps} $  is set to $ 1e^{-8} $. $ k_d $  and $ k_s $ are two additional scale factors. 
We set $ k_d=2,k_s=5 $ in all our experiments. 

\subsubsection{Application of Phong Model.}
Finally, with the 3D structure, material information, and the color of the light source $ I_F $, the reflection image $ I_i^{R} $  will be 
\begin{equation} 
    \begin{split}
        & C_1(u)=\max{\left(0,\mathbf{N}\left(u\right)*{\mathbf{L}_i\left(u\right)}\right)}/r_i\left(u\right)^2, \\
        & C_2(u)=\max{\left(0,\mathbf{R}_i\left(u\right)*{\mathbf{V}_i\left(u\right)}\right)}^{g_r}/r_i\left(u\right)^2, \\
        & I_{i}^{R}= s_F I_F (K_d C_1 + K_s C_2).
    \end{split}
\end{equation}
Following~\cite{phong}, we set the exponential $ g_r $ to 8. Here, $ s_F $ is the scale factor of light source image $ L_S $. 

\subsection{Training detail}
\label{Sec:training_detail}
Following \cite{monodepth2,monovit}, we use AdamW~\cite{AdamW}
as the optimizer, and the batch size is set to 12. 
We set $ lr $  of the Transformer-based encoder to $ 5e^{-5} $ and double it for the rest trainable part. 
An exponential decay learning rate scheduler is also used with 0.9 as exponential. 
The total training step is set to 40,000. 
We use the same augmentation for the image splits as applied in~\cite{adds,rnw,monodepth2,monovit}.

For BPG, we set $ F^{\min}=s_F^{\min}=0.5$
% (Eq.~\ref{equ:flare_intensity})
, empirically. 
Besides, we use $ s_F^{\max}= 2, F^{\max}=2$ as the default setting. 
As to ING, we set $ K^{\min}=0.1 $~
% (Eq.~\ref{noise_intensity})
following ELD~\cite{denoise_ELD}. We use $ K^{\max} = 1$ with $ {\cal P} + TL $ as the default setting. 
We further discuss the effect of $ F^{\max}, K^{\max} $ as well as the composition of ING in our main paper. 
% Fig.~\ref{Fig:fdat} and Fig.~\ref{Fig:ndat}. 

Since the processing of Re-rendering requires the corresponding depth
map as input, decay usage of BPG and ING is applied, and the starting step
is set to 20,000. To maintain performance on daytime images, BPG and ING
processing rates are set to 0.5, and these two probabilities are independent
of each other. 
\subsection{Augmentation on light source images}
For the sampled light source image, a random rotation $ {\cal{U}} (0,2\pi)$ and a random flip
is applied. Then, a random weight follows $ {\cal{U}} (1,3)$ is used to scale the brightness.
$ {\cal{U}} (0.8,1.2)$ is used for the scale of contrast and saturation.
A Gaussian blur is also used with a sigma limit range of $ [0.1,3] $ 
to enhance the low-passing glow effect.
\subsection{Pre-crop of nuScenes and Oxford RobotCar}
Pre-cropping is used in RNW~\cite{rnw} and ADDS~\cite{adds} to balance the image aspect ratio and cut out the imaging of car hoods.
We follow the same approach as proposed in these papers.
For nuScenes, the images are first center-cropped from 1600$ \times $900 to 1536$ \times $768 before testing or
the training of DA or DT types methods. 
As to the Oxford RobotCar dataset, they are center-cropped from  1280$ \times $960 to 1280$ \times $640.  
The camera intrinsic $ K_{I} $ is adjusted accordingly (only when the training set is used).
% \subsection{Nighttime Validation Set}
% Neither nuScenes-Night~\cite{NuScenes,rnw} nor RobotCar-Night~\cite{robotcar,adds} provides a corresponding validation set.
% Thus, we build our own validation set.
% Since the generated depth maps of nuScenes do not suffer from significant cumulative errors~\cite{rnw}, we sample our validation set from nuScenes dataset instead of Oxford RobotCar dataset.
% The original nuScenes dataset contains 99 different nighttime scenes.
% While the nuScenes Night training set~\cite{rnw} occupies 44 of them, nuScenes-Night test set takes 16 of them.
% Using the remaining 39 splits, 800 images are randomly sampled to make up our nighttime validation set.
% Note that we apply this nighttime validation set in all our experiments.
\subsection{Evaluation Metrics}
We use seven standard metrics for evaluation~\cite{kitti_eigen,monocular_review_2020}, including ABS rel, Sq rel, RMSE, RMSE log, $ \delta_1 $, $ \delta_2 $ and $ \delta_3 $:
\begin{equation}
    \begin{split}
        & \text{Abs rel} = \frac{1}{|N_d|} \sum_{i \in N_d} 
        \frac{\rvert d_i-d_i^{*} \rvert}{d_i^{*}} , \\
        & \text{Sq rel} =  \frac{1}{|N_d|} \sum_{i \in N_d}
        \frac{\rVert d_i-d_i^{*}\rVert^2}{d_i^{*}}  , \\
        & \text{RMSE} = \sqrt{ \frac{1}{|N_d|} \sum_{i \in N_d}\rVert d_i-d_i^{*} \rVert^2 }, \\
        & \text{RMSE log} = \sqrt{ \frac{1}{ \rvert N_d \rvert} \sum_{i \in N_d}\rVert \log{d_i}-\log{d_i^{*}} \rVert^2 }, \\
        & \delta_j: \, \% \; \text{of} \; d_i \; s.t. \ \max\left(\frac{d_i}{d_i^{*}}, \frac{d_i^{*}}{d_i}\right) < 1.25^{j}, \\  
    \end{split}
    \label{equ:metrics}
\end{equation}
where $ d_i $  is the predicted depth value of pixel $ i $, $ d_i^{*} $ refers to the ground truth of depth, and $ N_d $  denotes the total number of pixels with real-depth values.
% According to Eq.~\ref{equ:metrics}, Sq rel and RMSE will be more sensitive to outliers compared to ABS rel.
The definitions of Sq rel and RMSE suggest that they are more sensitive to prediction outliers compared to ABS rel.

\section{Additional results}
\subsection{Evaluation on nuScenes-Day and RobotCar-Day}
Both ADDS~\cite{adds} and WSGD~\cite{alldaydemono} are proposed to be all-day monocular depth estimation frameworks (tested on day and night).
To complete our effective testing part, in Table~\ref{Table:A1}, we present the qualitative comparison on nuScenes-Day~\cite{rnw,NuScenes} and RobotCar-Day~\cite{adds,robotcar}.

In the domain adaptation-based or photometric loss repair-based methods, which suffer from poor self-supervision on trainable networks, the trained depth network does not take full advantage of the additional image information.
Furthermore, in WSGD~\cite{alldaydemono}, due to the strong coupling of the DepthNet, PoseNet, Residual Flow Net and Illuminating Change Net during training, their approach does not fully benefit from the global and local attention of MonoViT~\cite{monovit}. 
In contrast, making use of the good generalization ability of the improved DepthNet backbone~(compared to the pure CNN backbone), our method generalizes well to \emph{unseen} day scenes and outperforms 
% or presents a competitive performance compared with 
two all-day approaches (DA:~ADDS, DT:~WSGD).

Furthermore, when tested on nuScenes-Day and RobotCar-Day, our method outperforms G: MonoViT~\cite{monovit} in Sq rel and RMSE which again suggests that the DepthNet trained by our framework prefers to make fewer prediction outliers rather than take the risk of making potentially more accurate but radical predictions.

\begin{table*}[htbp]
	\begin{center}
	\scriptsize
        \resizebox{0.9\linewidth}{!}{
        \begin{tabular}{c| l | c | c | c | c c c c | ccc}
            \hline
            Type & Method & Train on & Train Res. & Max depth & \cellcolor{cell1} ABS rel $ \downarrow $  & \cellcolor{cell1} Sq rel $ \downarrow $    & \cellcolor{cell1} RMSE $ \downarrow $  & \cellcolor{cell1} RMSE log $ \downarrow $   & \cellcolor{cell2} $\delta_{1}$ $ \uparrow $  & \cellcolor{cell2} $\delta_{2}$ $ \uparrow $  & \cellcolor{cell2} $\delta_{3}$ $ \uparrow $ \\
            \hline
            \multicolumn{12}{c}{\cellcolor{cell3} Test on nuScenes-Day} \\

            \multirow{2}{*}{DT}
            & MonoViT\cite{monovit}           & N $ d \& n $ &  640 $\times$ 320  & 60     & 0.201     	    & 4.346     	& 7.316     	& \underline{0.261}& \textbf{0.792}& \textbf{0.931} & 0.966 \\
            & WSGD\cite{alldaydemono}& N $ d \& n $ &  640 $\times$ 320  & 60     & 0.405     	    & 4.500     	& 11.281    	& 0.521     	& 0.355     	& 0.618     	& 0.793  \\
            \hline
            \multirow{4}{*}{DA}
            & ITDFA\cite{ITDFA}               & N $ d \& n $ &  640 $\times$ 320  & 60       & 0.352         & 7.941     	& 12.706    	& 0.395     	& 0.583     	& 0.812     	& 0.906  \\        
            & RNW\cite{rnw}                   & N $ d \& n $ &  640 $\times$ 320  & 60       & 0.268         & 3.722     	& 8.897     	& 0.333     	& 0.621     	& 0.855     	& 0.938     \\
            & ADDS\cite{adds}                 & N $ d \& n $ &  640 $\times$ 320  & 60       & \underline{0.183}         & 2.525     	& 7.728     	& \textbf{0.259}& 0.783     	& 0.926     	& \textbf{0.967}    \\
            \hline
            \multirow{5}{*}{G}
            & ITDFA\cite{ITDFA}               & R $ d \& n $ &  640 $\times$ 320  & 60       & 0.279     	    & 2.502     	& 7.938     	& 0.346     	& 0.526     	& 0.819     	& 0.934  \\
            & RNW\cite{rnw}                   & R $ d \& n $ &  640 $\times$ 320  & 60       & 0.286     	    & 2.991     	& 8.819     	& 0.383     	& 0.544     	& 0.799     	& 0.910  \\
            & ADDS\cite{adds}                 & R $ d \& n $ &  640 $\times$ 320  & 60       & 0.250     	    & 3.863     	& 9.257     	& 0.319     	& 0.678     	& 0.870     	& 0.943 \\
            & MonoFormer\cite{monoformer}     & K            &  768 $\times$ 256  & 60       & 0.204     	    & 2.219     	& 7.332     	& 0.286     	& 0.697     	& 0.901     	& 0.963 \\
            & MonoViT\cite{monovit}           & K            &  768 $\times$ 256  & 60       & \textbf{0.181}   & \underline{1.918}     	& \underline{7.070}     	& 0.274     	& \underline{0.743}& \underline{0.911}& \underline{0.964}  \\
            & \textbf{Ours}                   & K            &  768 $\times$ 256  & 60       & \underline{0.183}& \textbf{1.888}& \textbf{7.019}& 0.276     	& 0.736     	& 0.908     	& \underline{0.964}  \\
            \hline
            \multicolumn{12}{c}{\cellcolor{cell3} Test on RobotCar-Day} \\
            \multirow{2}{*}{DT}
            & MonoViT\cite{monovit}           & R $ d \& n $ &  640 $\times$ 320 & 40       & 0.140     	& 1.010     	& 4.713     	& 0.180     	& 0.808     	& 0.966     	& 0.989 \\
            & WSGD\cite{alldaydemono}& R $ d \& n $ &  640 $\times$ 320 & 40       & 0.123     	& 0.694     	& 3.957     	& 0.163     	& 0.835     	& \textbf{0.978}& \textbf{0.995} \\
            \hline
            \multirow{4}{*}{DA}
            & ITDFA\cite{ITDFA}               & R $ d \& n $ &  640 $\times$ 320 & 40       & 0.156     	& 0.786     	& 3.971     	& 0.192     	& 0.778     	& 0.965     	& 0.992 \\
            & ADDS\cite{adds}\dag             & R $ d \& n $ &  640 $\times$ 320 & 40       & \underline{0.109}& 0.584         & 3.578         & 0.153         & 0.631         & 0.908         & 0.962 \\
            & ADDS\cite{adds}                 & R $ d \& n $ &  640 $\times$ 320 & 40       & 0.120     	& 0.778     	& 4.113     	& 0.172     	& 0.856     	& 0.967     	& 0.987  \\
            & RNW\cite{rnw}                   & R $ d \& n $ &  640 $\times$ 320 & 40       & 0.156     	& 0.701     	& 3.838     	& 0.202     	& 0.768     	& 0.958     	& 0.991 \\
            \hline     
            \multirow{5}{*}{G}     
            & ITDFA\cite{ITDFA}               & N $ d \& n $ &  640 $\times$ 320 & 40       & 0.230     	& 2.018     	& 6.336     	& 0.281     	& 0.653     	& 0.875     	& 0.956 \\
            & ADDS\cite{adds}                 & N $ d \& n $ &  640 $\times$ 320 & 40       & 0.180     	& 1.441     	& 5.680     	& 0.232     	& 0.727     	& 0.930     	& 0.981 \\
            & RNW\cite{rnw}                   & N $ d \& n $ &  640 $\times$ 320 & 40       & 0.271     	& 3.939     	& 9.204     	& 0.314     	& 0.661     	& 0.830     	& 0.927 \\
            & MonoFormer\cite{monoformer}     & K            &  768 $\times$ 256 & 40       & 0.138     	& 0.665     	& 3.730     	& 0.181     	& 0.824     	& 0.971     	& \underline{0.993} \\
            & MonoViT\cite{monovit}           & K            &  768 $\times$ 256 & 40       & \textbf{0.102}& \underline{0.500}& \underline{3.364}& \textbf{0.147}& \textbf{0.905}& \underline{0.977}     	& 0.992  \\
            & \textbf{Ours}                   & K            &  768 $\times$ 256 & 40       & \underline{0.109}& \textbf{0.477}& \textbf{3.261}& \underline{0.151}& \underline{0.891}& \textbf{0.978}& \underline{0.993}  \\
            \hline
        \end{tabular}
        }
    \end{center}
    \caption{
    \textbf{Effective test on nuScenes-Day~\protect\cite{rnw,NuScenes} and Robotcar-Day~\protect\cite{adds,robotcar}}. 
    All methods use the \emph{same} DepthNet backbone unless marked.
    K, N and R represent the KITTI~\protect\cite{kitti_eigen,KITTI}, nuScenes~\protect\cite{NuScenes} and Oxford RobotCar~\protect\cite{robotcar} datasets accordingly.
    $ d $ and $ n $ refer to the day and night training splits proposed by ADDS~\protect\cite{adds} or RNW~\protect\cite{rnw}.  
    \dag~means the original reported result in the paper with pure CNN backbone and Max depth as the clipping up range for the predicted depth~(i.e., a more relaxed approach for evaluation) instead of 80m.}
    \label{Table:A1}
\end{table*}

\subsection{Additional Visual Results on nuScenes and RobotCar}
\begin{figure*}[htbp]
    \centering
    \newcommand{\turnheightnew}{0.38\columnwidth}
    \begin{tabular}{@{\hskip 0mm}c@{\hskip 1mm}c@{\hskip 1mm}c@{\hskip 1mm}c@{\hskip 1mm}||@{\hskip 1mm} c@{\hskip 1mm}c@{}}

    {\rotatebox{90}{\hspace{5mm}\scriptsize{Input}}} &
    \includegraphics[width=\turnheightnew]{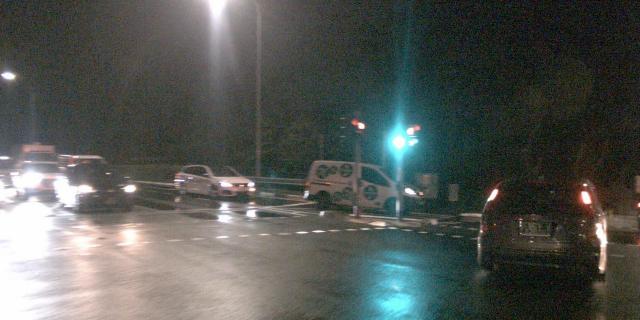} &
    \includegraphics[width=\turnheightnew]{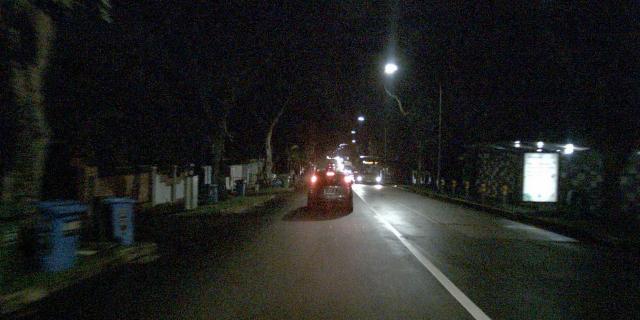} &
    \includegraphics[width=\turnheightnew]{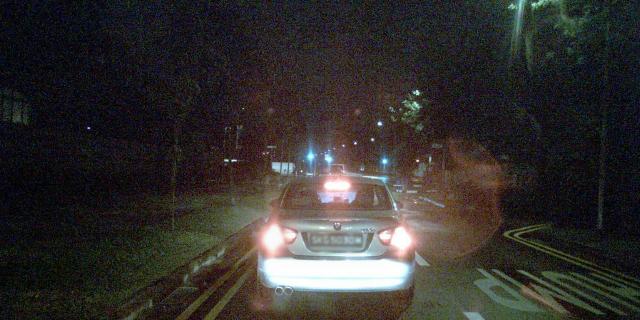} &
    \includegraphics[width=\turnheightnew]{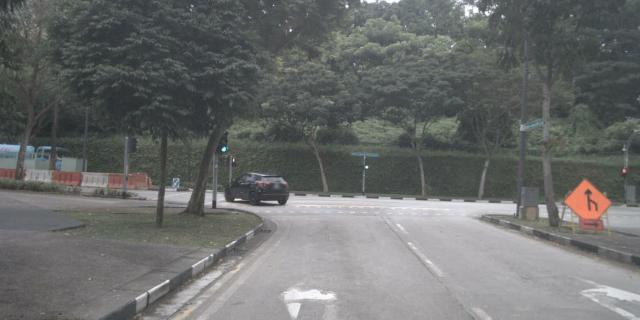} &
    \includegraphics[width=\turnheightnew]{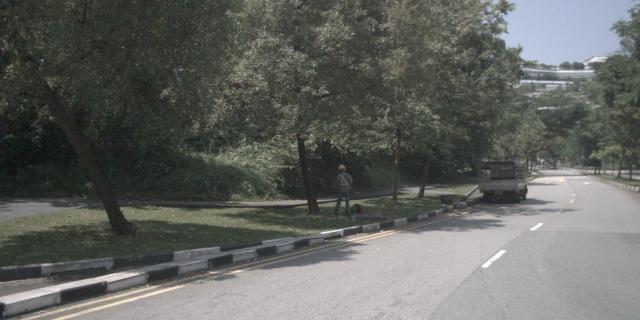} \\
    % \vspace{1mm}

    \rotatebox{90}{\hspace{2mm}\tiny{DT:~MonoViT}}&
    \includegraphics[width=\turnheightnew]{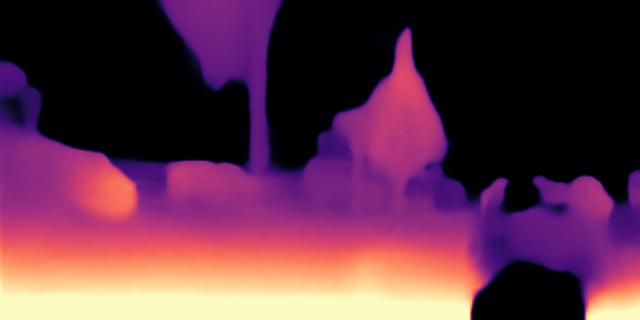} &
    \includegraphics[width=\turnheightnew]{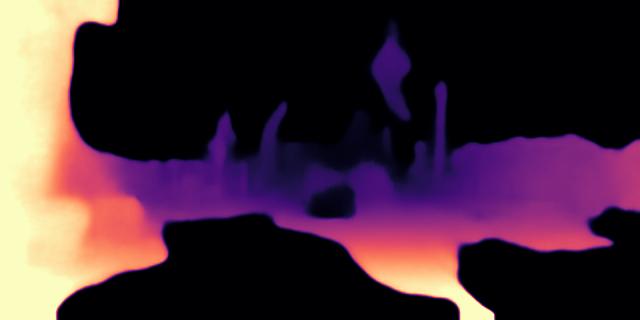} &
    \includegraphics[width=\turnheightnew]{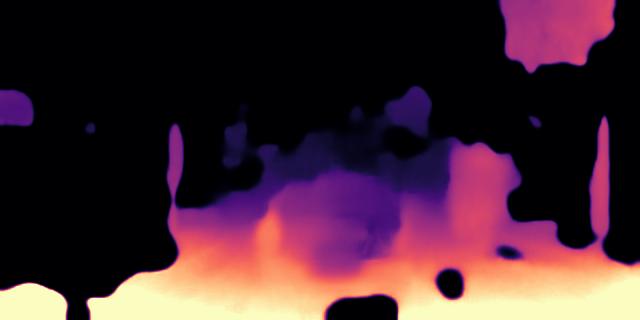} &
    \includegraphics[width=\turnheightnew]{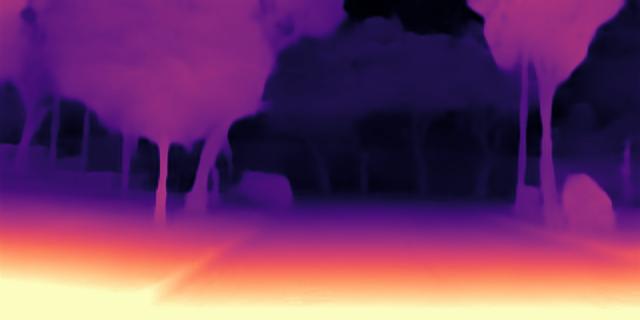} &
    \includegraphics[width=\turnheightnew]{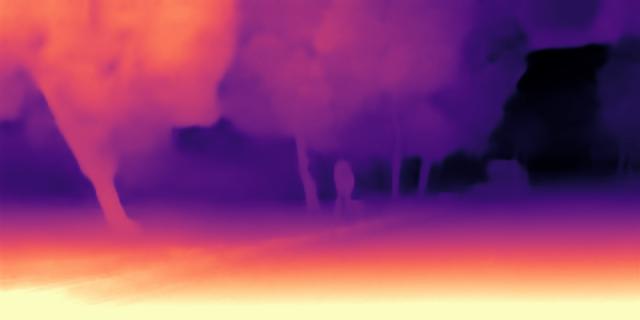} \\
    
    {\rotatebox{90}{\hspace{0mm}\tiny{DT:~WSGD}}} &
    \includegraphics[width=\turnheightnew]{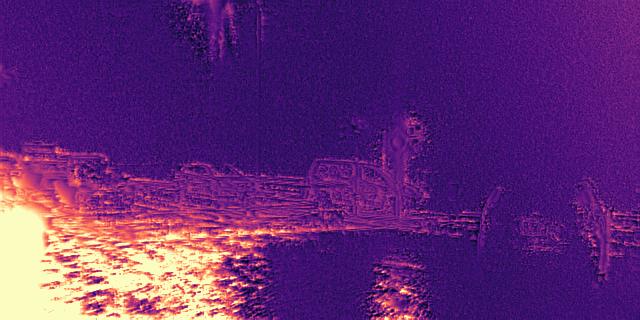} &
    \includegraphics[width=\turnheightnew]{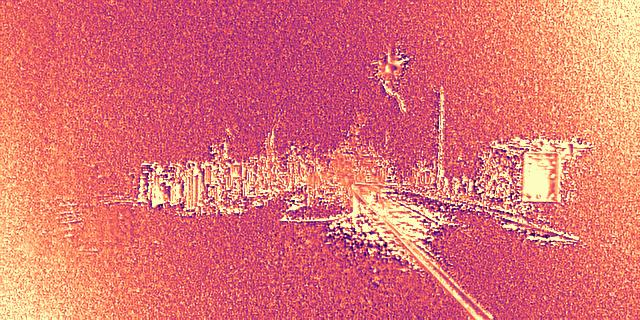} &
    \includegraphics[width=\turnheightnew]{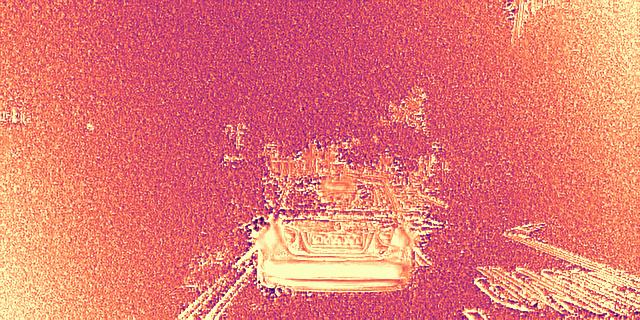} &
    \includegraphics[width=\turnheightnew]{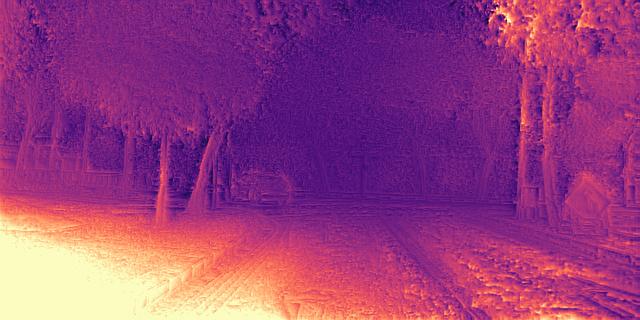} &
    \includegraphics[width=\turnheightnew]{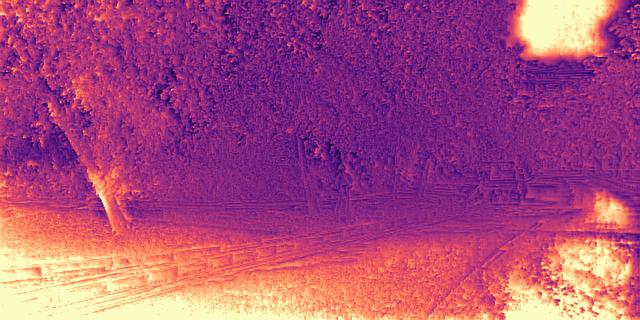} \\

    \rotatebox{90}{\hspace{0mm}\scriptsize{DA:~ITDFA}}&
    \includegraphics[width=\turnheightnew]{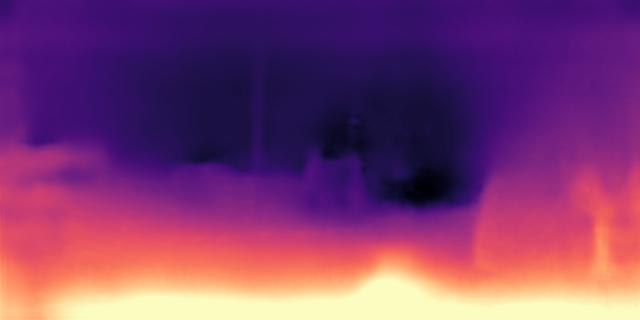} &
    \includegraphics[width=\turnheightnew]{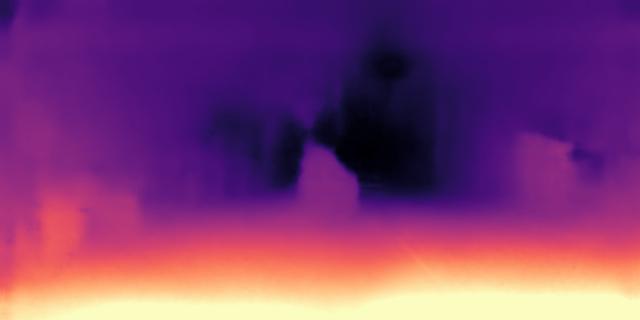} &
    \includegraphics[width=\turnheightnew]{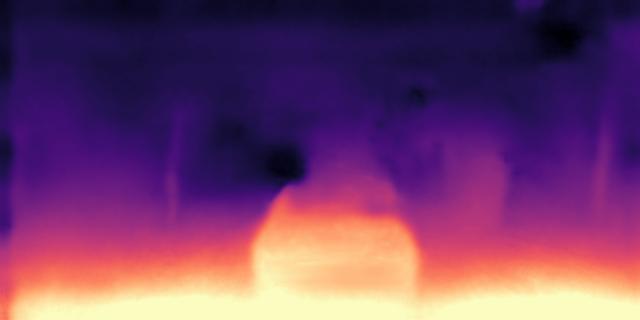} &
    \includegraphics[width=\turnheightnew]{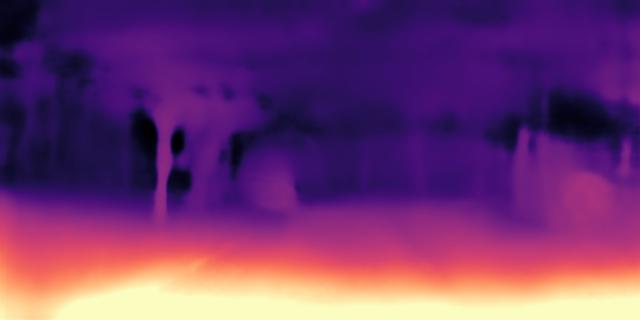} &
    \includegraphics[width=\turnheightnew]{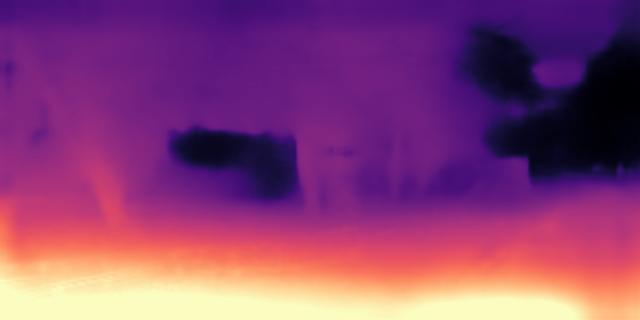} \\

    \rotatebox{90}{\hspace{1mm}\scriptsize{DA:~RNW}}&
    \includegraphics[width=\turnheightnew]{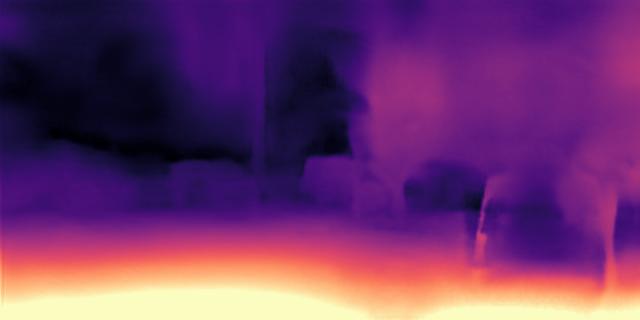} &
    \includegraphics[width=\turnheightnew]{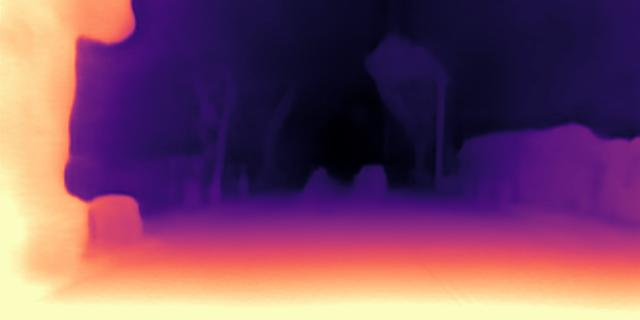} &
    \includegraphics[width=\turnheightnew]{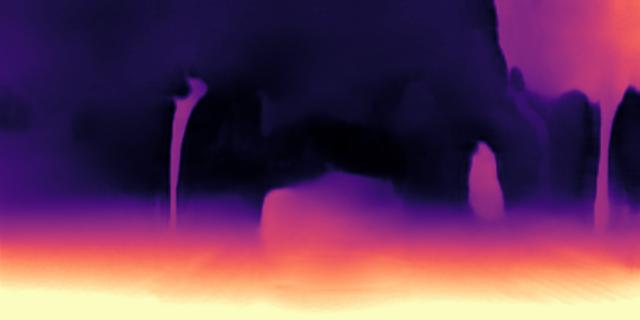} &
    \includegraphics[width=\turnheightnew]{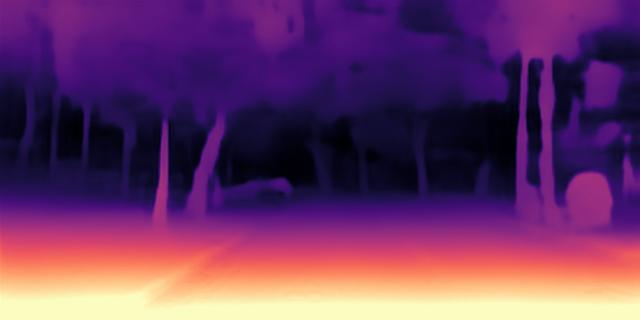} &
    \includegraphics[width=\turnheightnew]{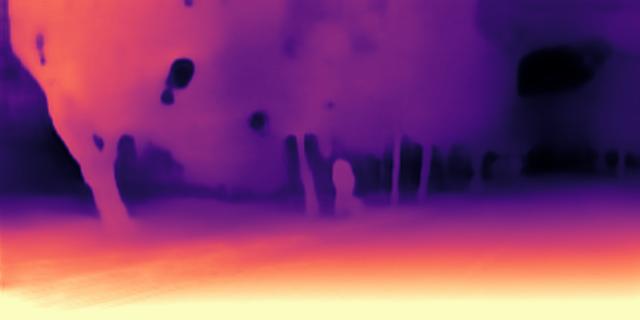} \\

    \rotatebox{90}{\hspace{0mm}\scriptsize{DA:~ADDS}}&
    \includegraphics[width=\turnheightnew]{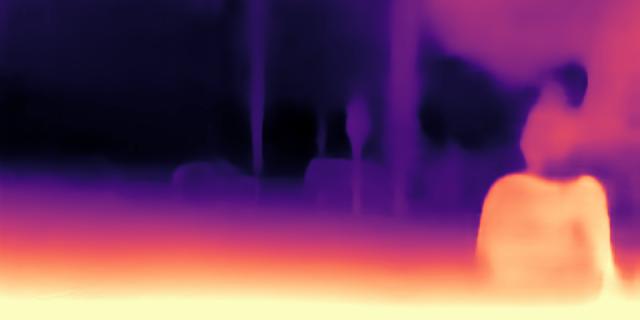} &
    \includegraphics[width=\turnheightnew]{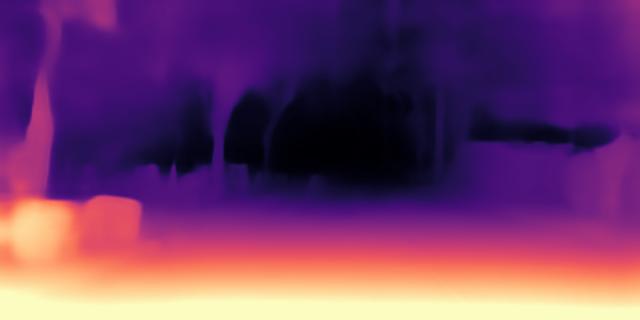} &
    \includegraphics[width=\turnheightnew]{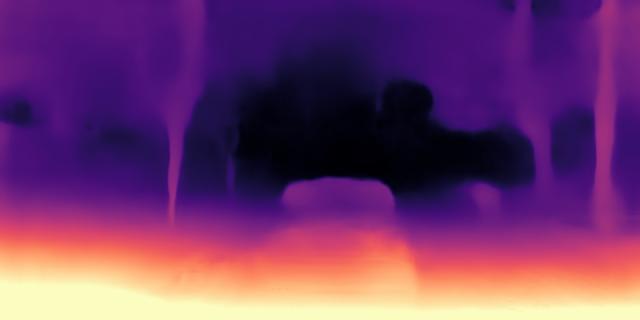} &
    \includegraphics[width=\turnheightnew]{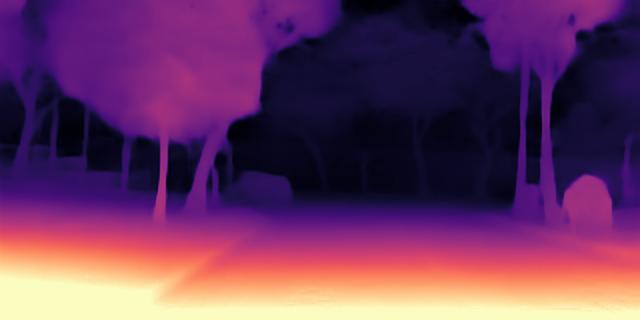} &
    \includegraphics[width=\turnheightnew]{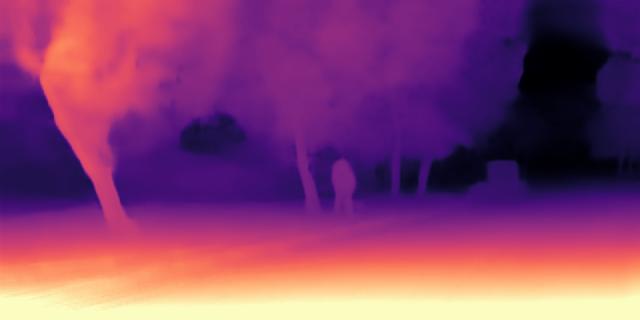} \\

    \rotatebox{90}{\hspace{0mm}\scriptsize{G:~ITDFA}}&
    \includegraphics[width=\turnheightnew]{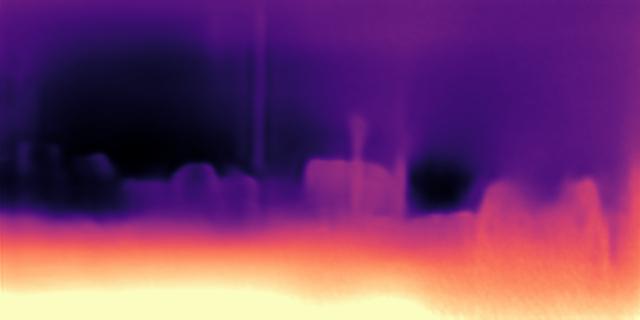} &
    \includegraphics[width=\turnheightnew]{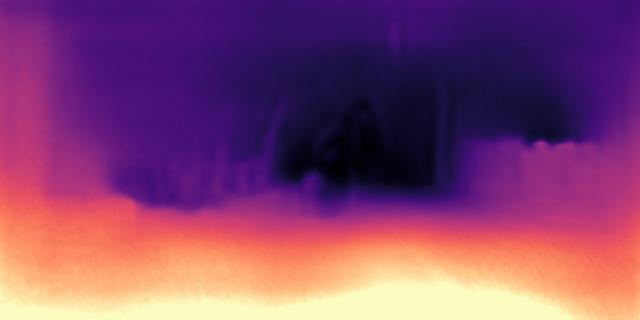} &
    \includegraphics[width=\turnheightnew]{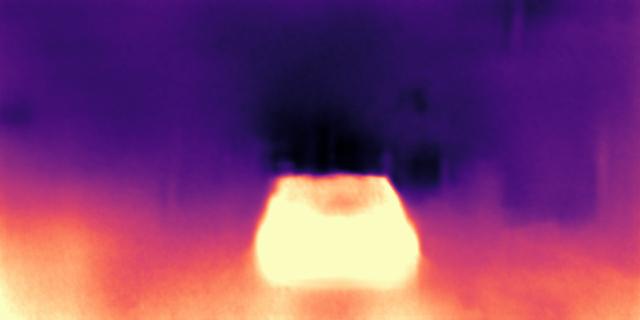} &
    \includegraphics[width=\turnheightnew]{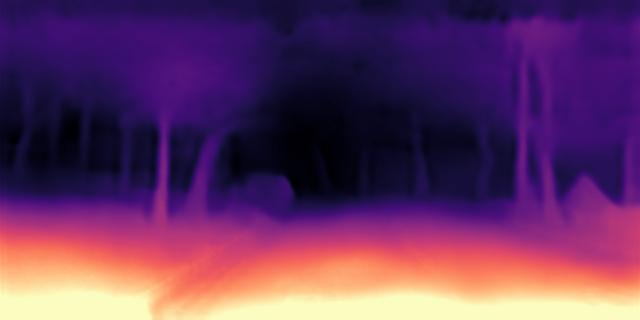} &
    \includegraphics[width=\turnheightnew]{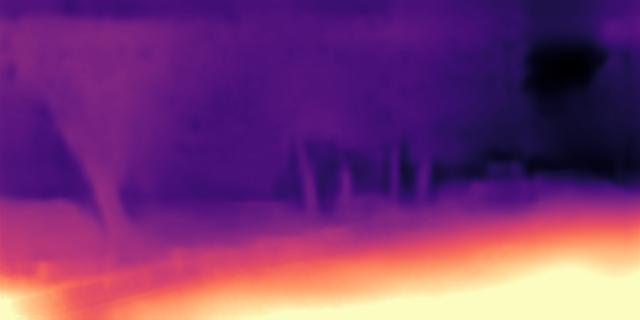} \\

    \rotatebox{90}{\hspace{2mm}\scriptsize{G:~RNW}}&
    \includegraphics[width=\turnheightnew]{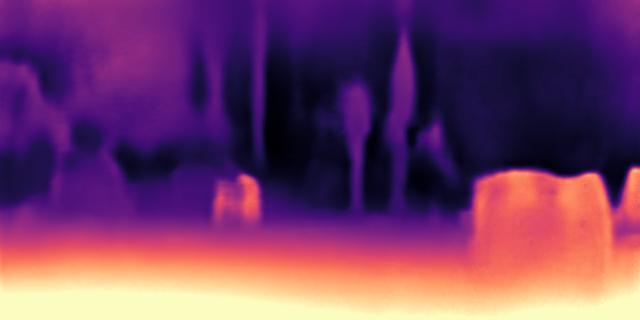} &
    \includegraphics[width=\turnheightnew]{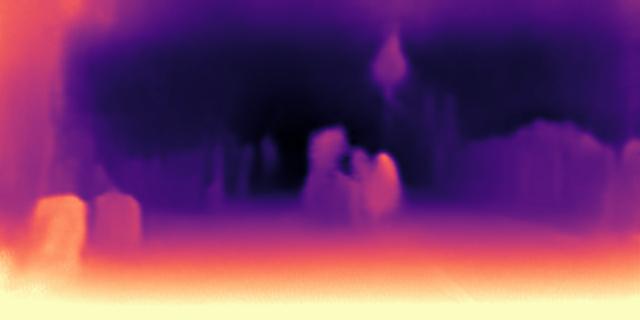} &
    \includegraphics[width=\turnheightnew]{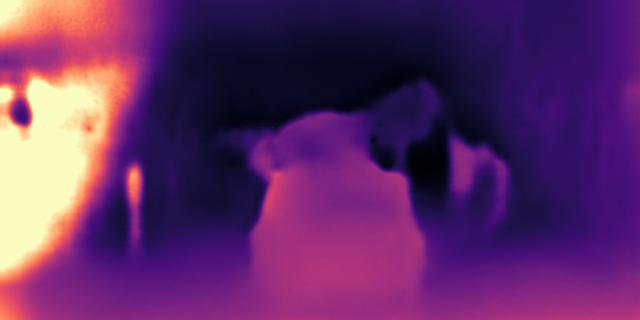} &
    \includegraphics[width=\turnheightnew]{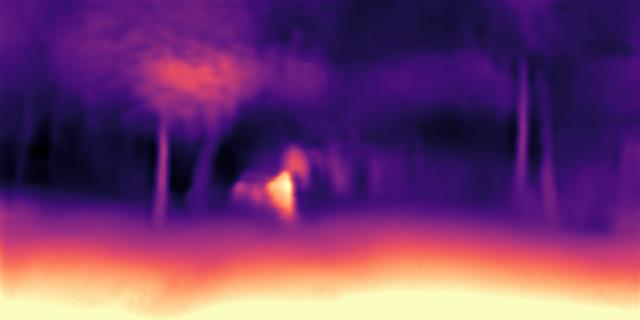} &
    \includegraphics[width=\turnheightnew]{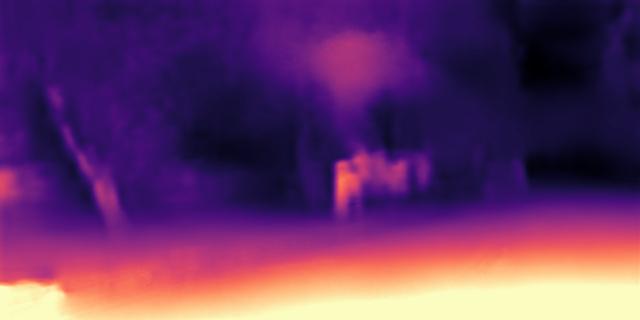} \\

    \rotatebox{90}{\hspace{1mm}\scriptsize{G:~ADDS}}&
    \includegraphics[width=\turnheightnew]{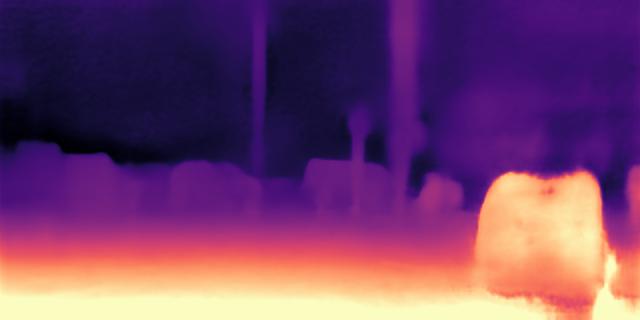} &
    \includegraphics[width=\turnheightnew]{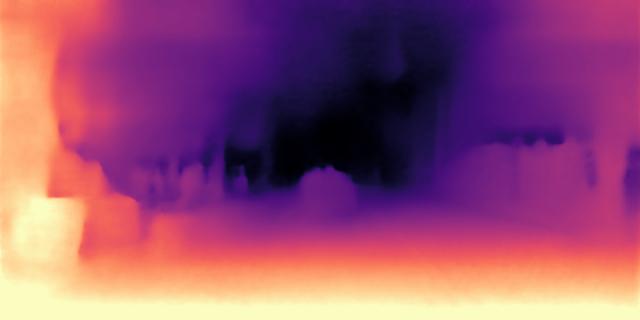} &
    \includegraphics[width=\turnheightnew]{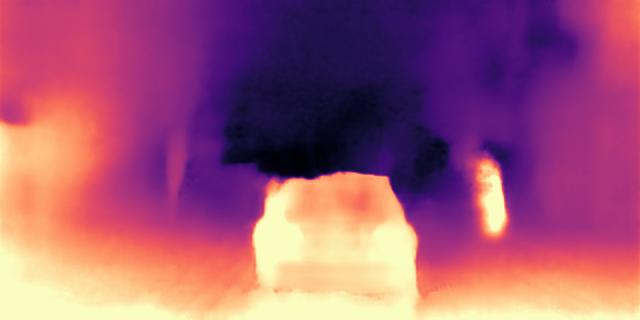} &
    \includegraphics[width=\turnheightnew]{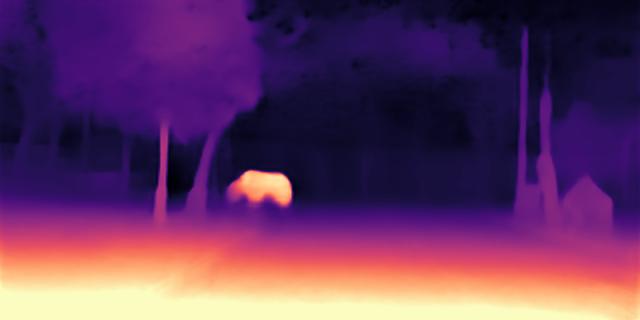} &
    \includegraphics[width=\turnheightnew]{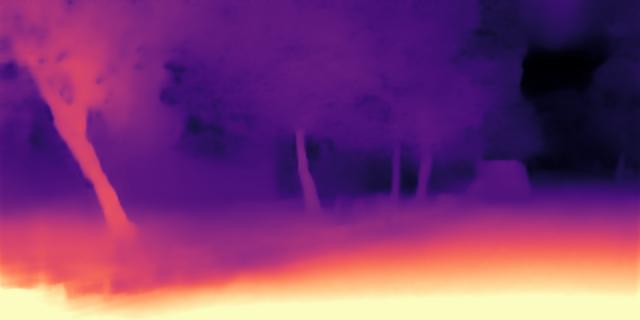} \\
    
    \rotatebox{90}{\hspace{-1.5mm}\scriptsize{G:~MonoFormer}}&
    \includegraphics[width=\turnheightnew]{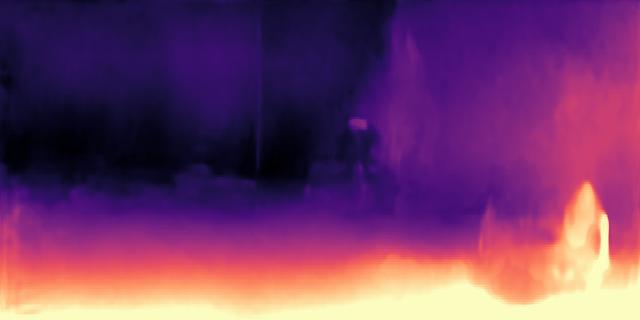} &
    \includegraphics[width=\turnheightnew]{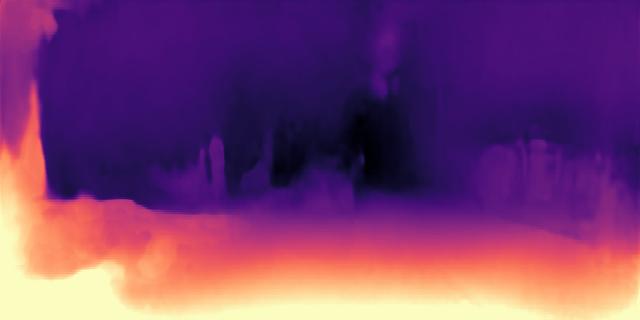} &
    \includegraphics[width=\turnheightnew]{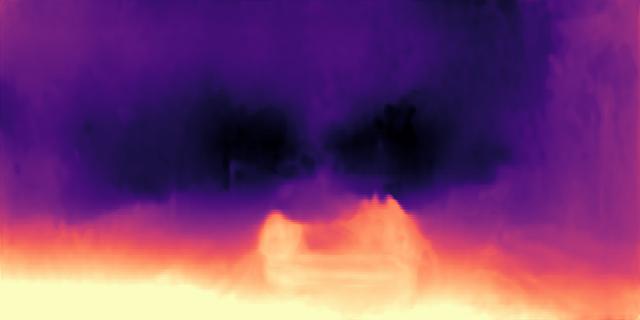} &
    \includegraphics[width=\turnheightnew]{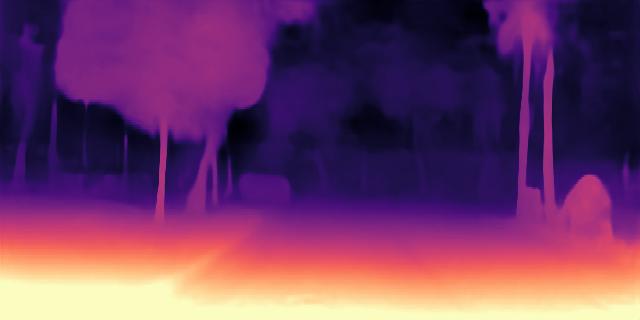} &
    \includegraphics[width=\turnheightnew]{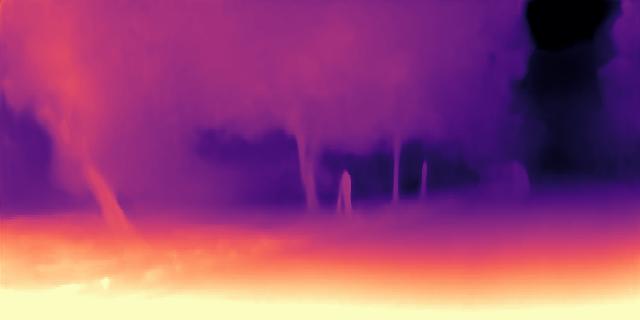} \\

    \rotatebox{90}{\hspace{-1.5mm}\scriptsize{G:~MonoViT}}&
    \includegraphics[width=\turnheightnew]{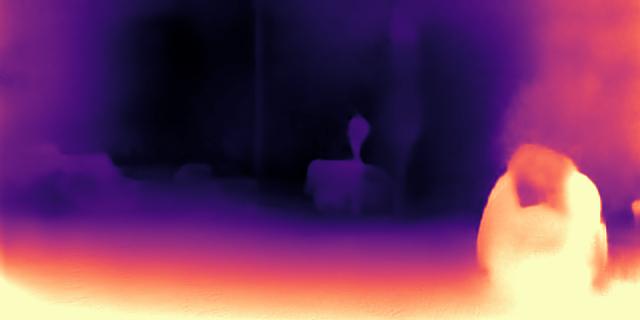} &
    \includegraphics[width=\turnheightnew]{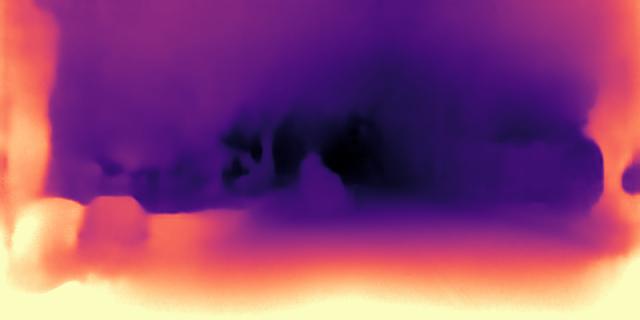} &
    \includegraphics[width=\turnheightnew]{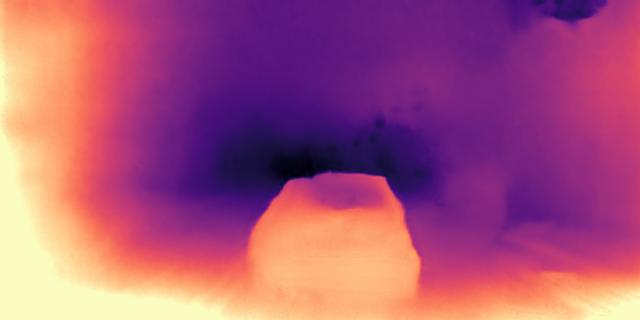} &
    \includegraphics[width=\turnheightnew]{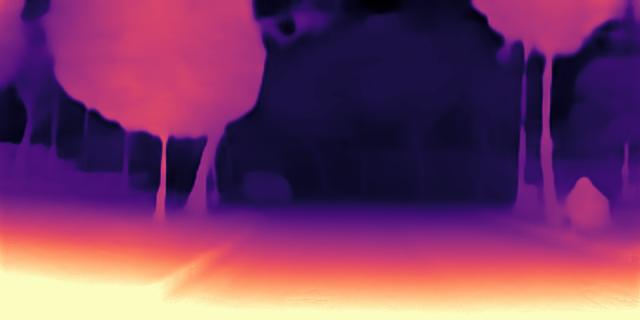} &
    \includegraphics[width=\turnheightnew]{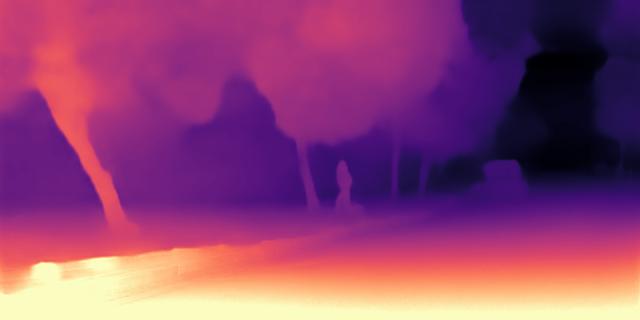} \\

    \rotatebox{90}{\hspace{4mm}\scriptsize{G:~Ours}}&
    \includegraphics[width=\turnheightnew]{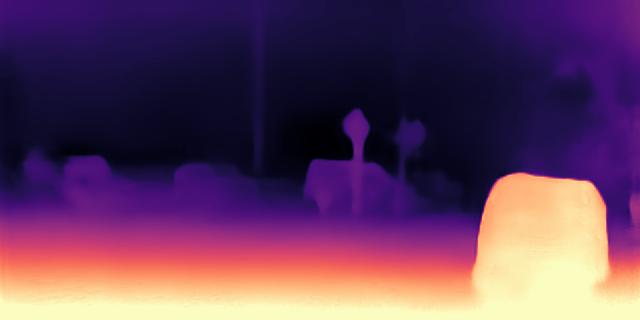} &
    \includegraphics[width=\turnheightnew]{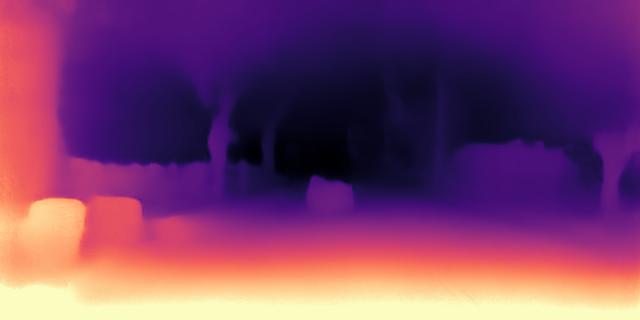} &
    \includegraphics[width=\turnheightnew]{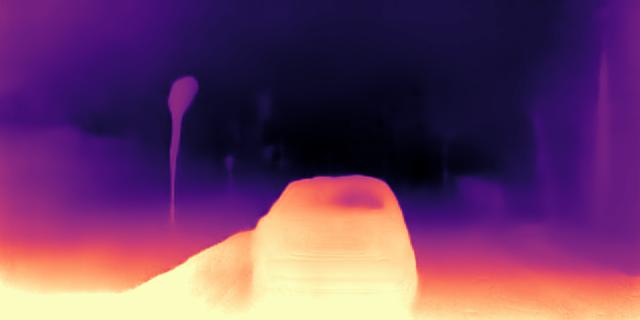} &
    \includegraphics[width=\turnheightnew]{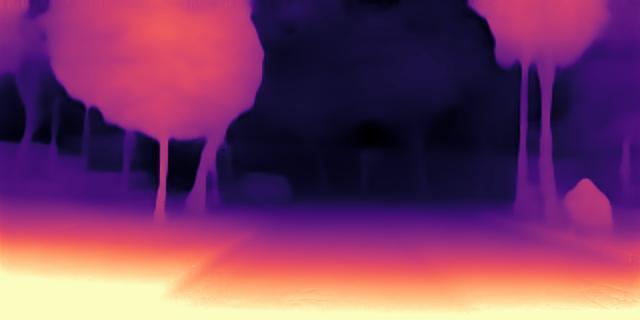} &
    \includegraphics[width=\turnheightnew]{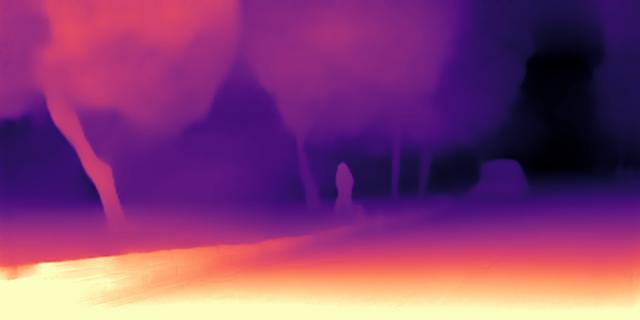} \\

    \end{tabular}
    \caption{\textbf{Qualitative results on nuScenes-Night and nuScenes-Day~\protect\cite{rnw,NuScenes}.}} 
    \label{fig:nnd}
\end{figure*}

\begin{figure*}[htbp]
    \centering
    \newcommand{\turnheightnew}{0.38\columnwidth}
    \begin{tabular}{@{\hskip 0mm}c@{\hskip 1mm}c@{\hskip 1mm}c@{\hskip 1mm}c@{\hskip 1mm}||@{\hskip 1mm} c@{\hskip 1mm}c@{}}

    {\rotatebox{90}{\hspace{5mm}\scriptsize{Input}}} &
    \includegraphics[width=\turnheightnew]{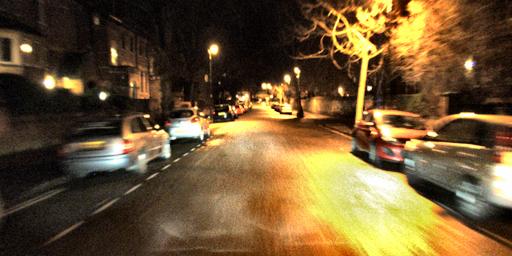} &
    \includegraphics[width=\turnheightnew]{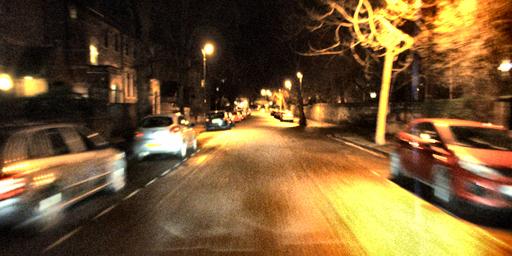} &
    \includegraphics[width=\turnheightnew]{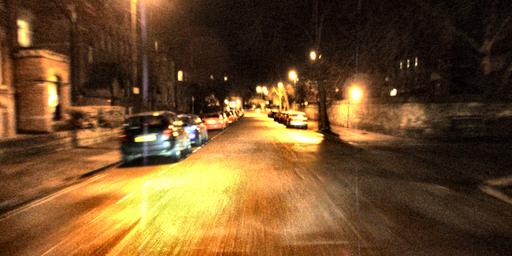} &
    \includegraphics[width=\turnheightnew]{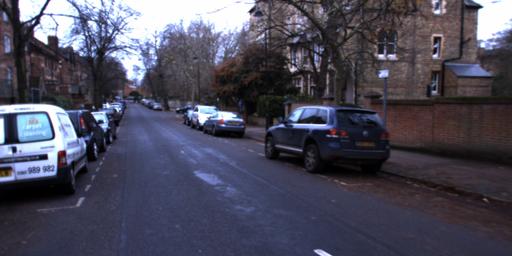} &
    \includegraphics[width=\turnheightnew]{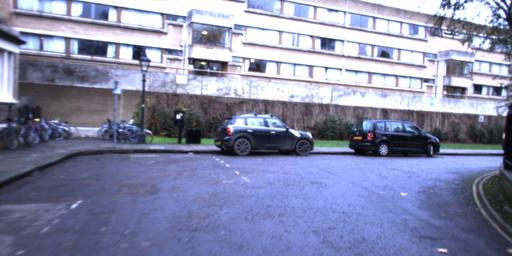} \\
    % \vspace{1mm}

    \rotatebox{90}{\hspace{2mm}\tiny{DT:~MonoViT}}&
    \includegraphics[width=\turnheightnew]{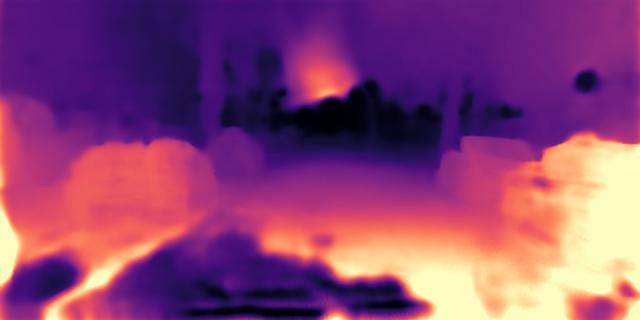} &
    \includegraphics[width=\turnheightnew]{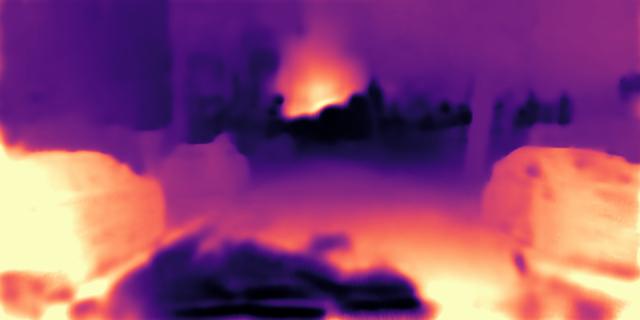} &
    \includegraphics[width=\turnheightnew]{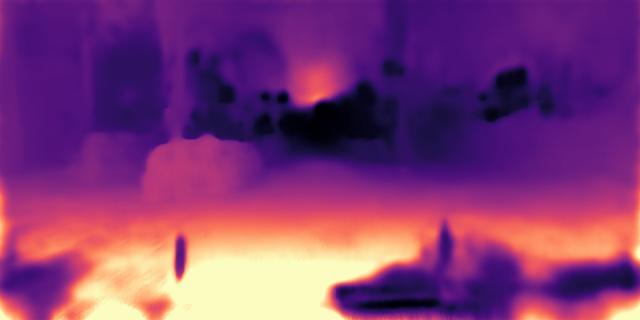} &
    \includegraphics[width=\turnheightnew]{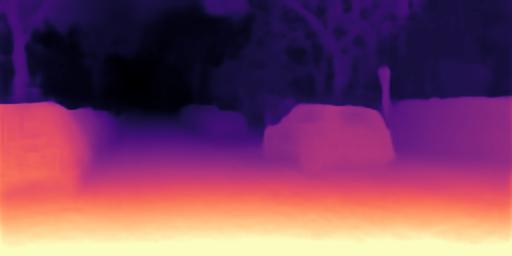} &
    \includegraphics[width=\turnheightnew]{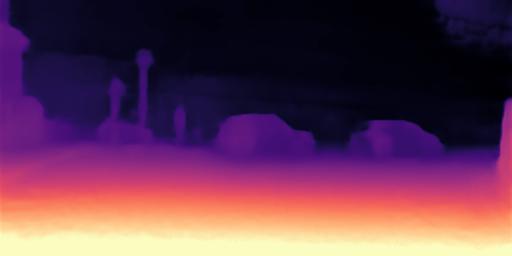} \\
    
    {\rotatebox{90}{\hspace{0mm}\tiny{DT:~WSGD}}} &
    \includegraphics[width=\turnheightnew]{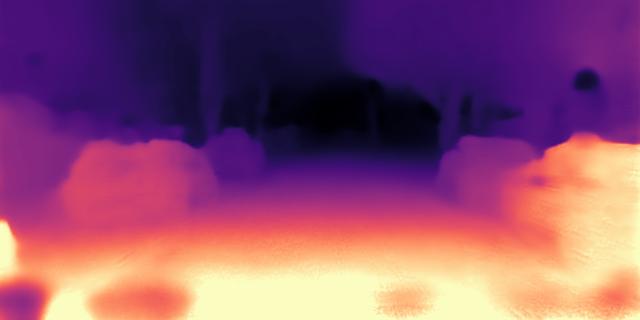} &
    \includegraphics[width=\turnheightnew]{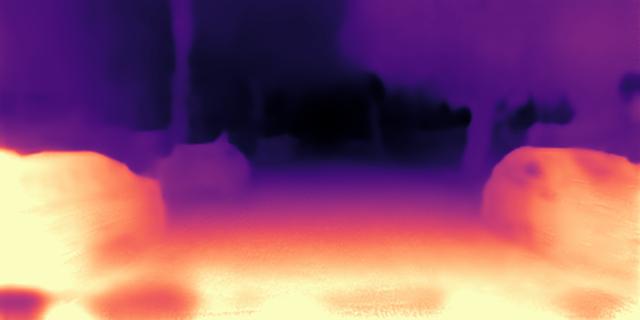} &
    \includegraphics[width=\turnheightnew]{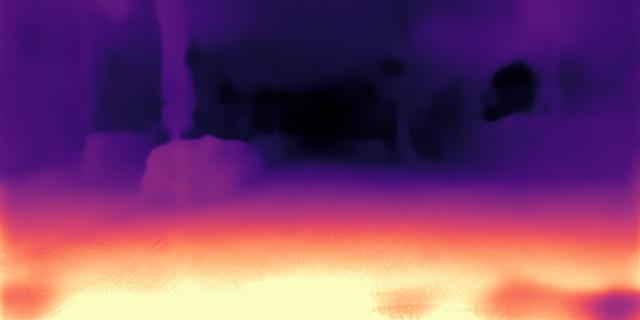} &
    \includegraphics[width=\turnheightnew]{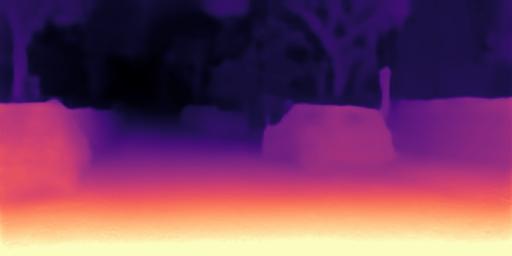} &
    \includegraphics[width=\turnheightnew]{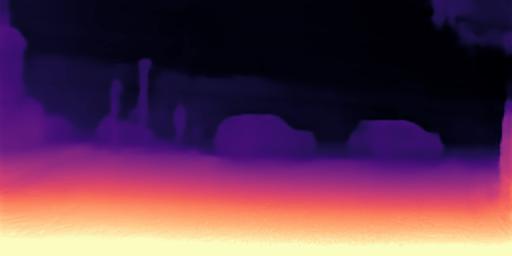} \\

    \rotatebox{90}{\hspace{0mm}\scriptsize{DA:~ITDFA}}&
    \includegraphics[width=\turnheightnew]{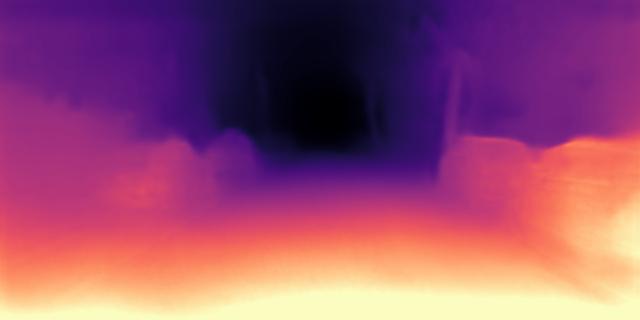} &
    \includegraphics[width=\turnheightnew]{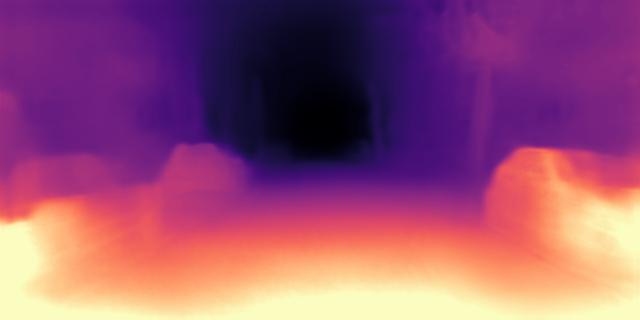} &
    \includegraphics[width=\turnheightnew]{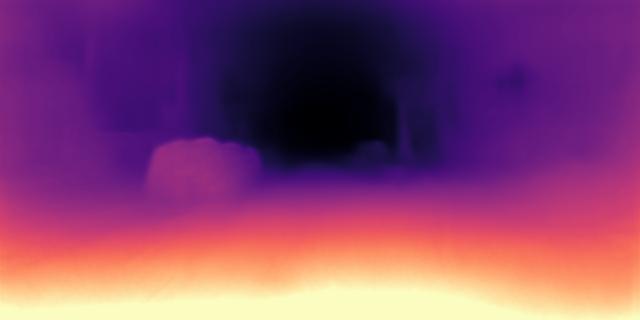} &
    \includegraphics[width=\turnheightnew]{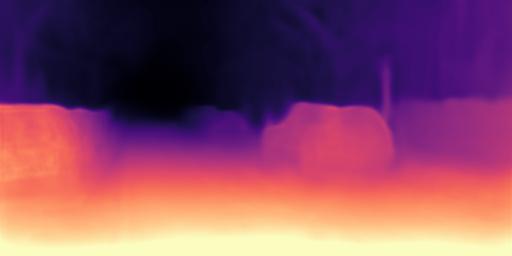} &
    \includegraphics[width=\turnheightnew]{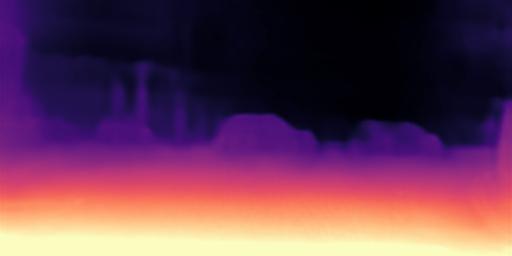} \\

    \rotatebox{90}{\hspace{1mm}\scriptsize{DA:~RNW}}&
    \includegraphics[width=\turnheightnew]{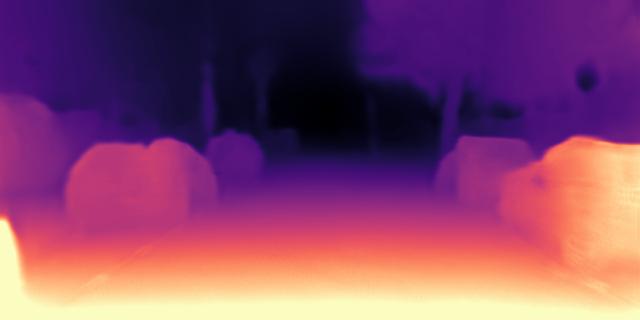} &
    \includegraphics[width=\turnheightnew]{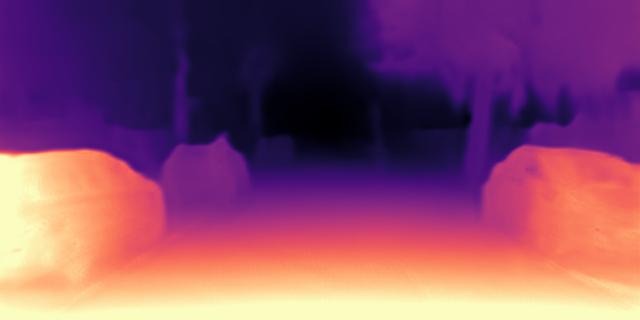} &
    \includegraphics[width=\turnheightnew]{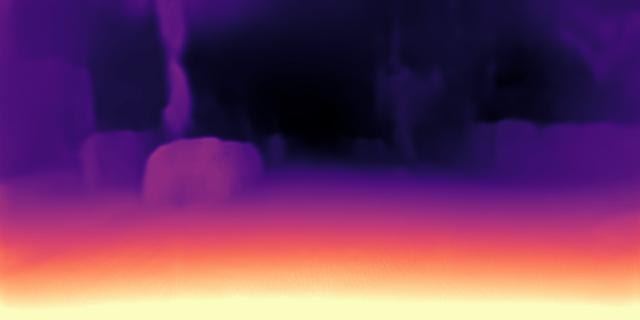} &
    \includegraphics[width=\turnheightnew]{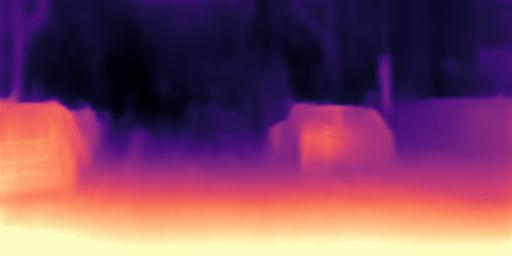} &
    \includegraphics[width=\turnheightnew]{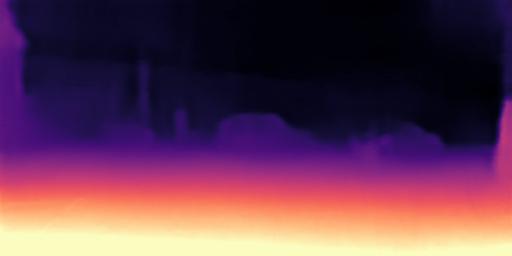} \\

    \rotatebox{90}{\hspace{0mm}\scriptsize{DA:~ADDS}}&
    \includegraphics[width=\turnheightnew]{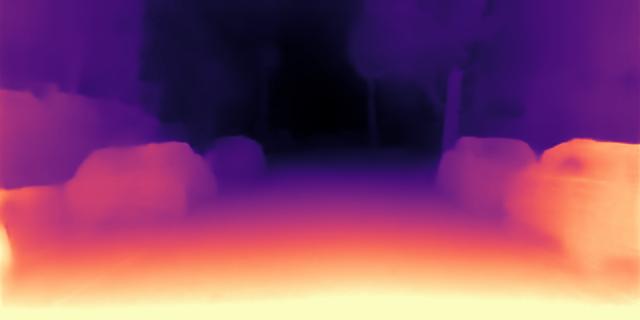} &
    \includegraphics[width=\turnheightnew]{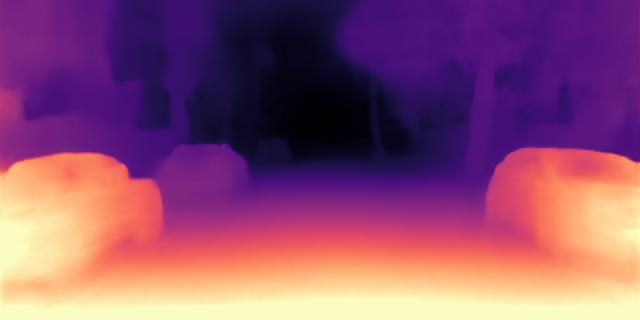} &
    \includegraphics[width=\turnheightnew]{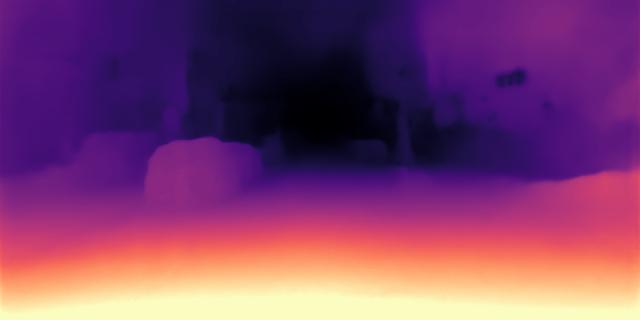} &
    \includegraphics[width=\turnheightnew]{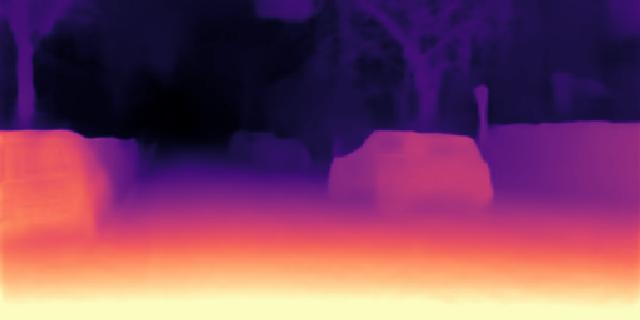} &
    \includegraphics[width=\turnheightnew]{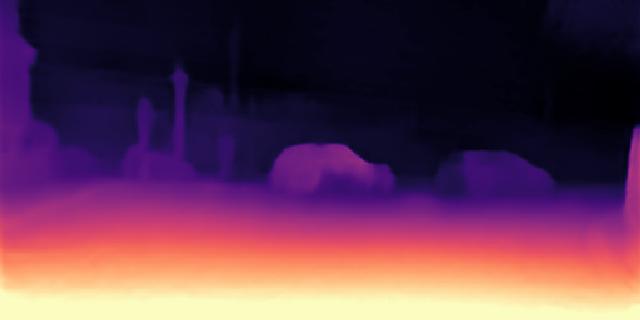} \\

    \rotatebox{90}{\hspace{0mm}\scriptsize{G:~ITDFA}}&
    \includegraphics[width=\turnheightnew]{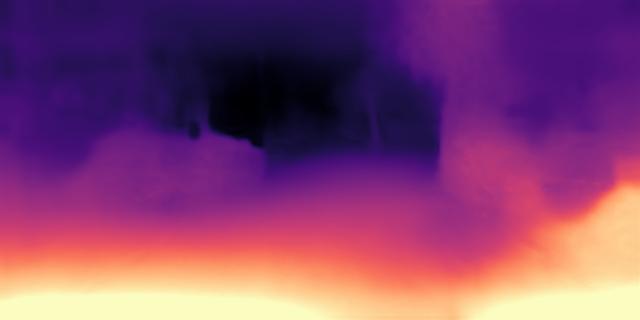} &
    \includegraphics[width=\turnheightnew]{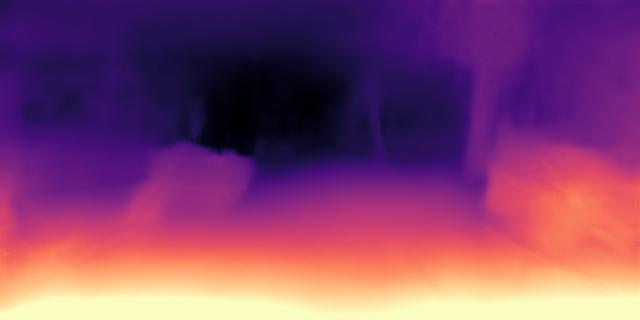} &
    \includegraphics[width=\turnheightnew]{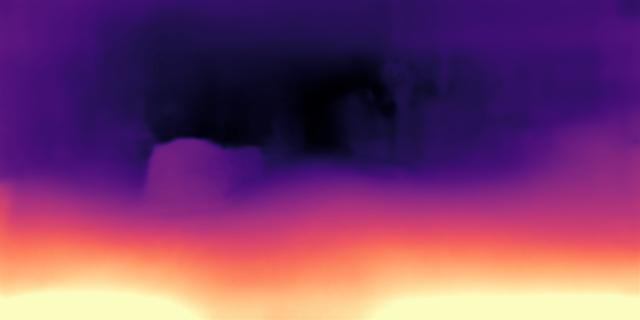} &
    \includegraphics[width=\turnheightnew]{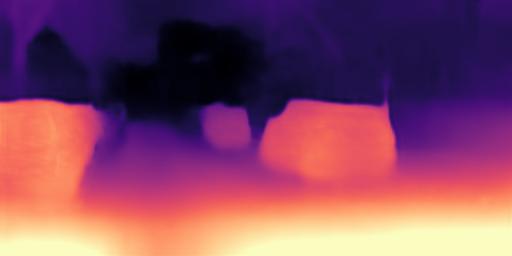} &
    \includegraphics[width=\turnheightnew]{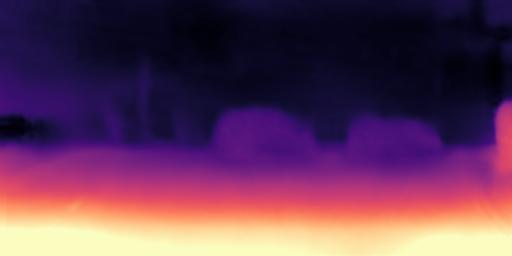} \\

    \rotatebox{90}{\hspace{2mm}\scriptsize{G:~RNW}}&
    \includegraphics[width=\turnheightnew]{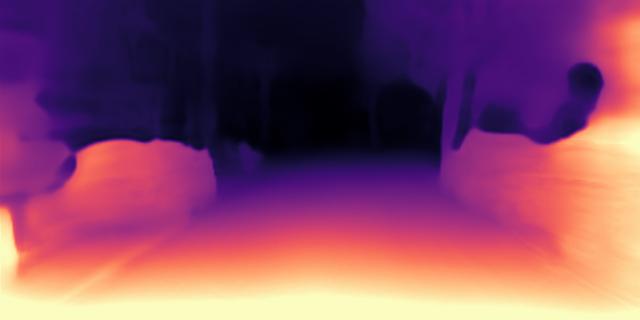} &
    \includegraphics[width=\turnheightnew]{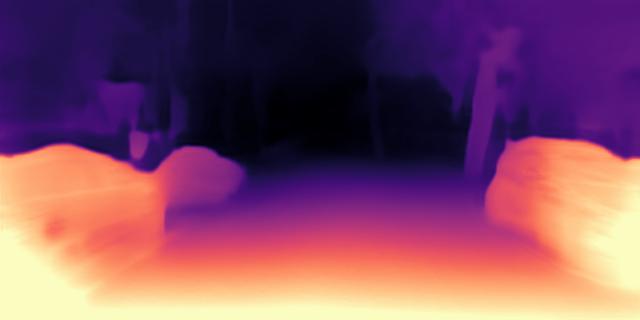} &
    \includegraphics[width=\turnheightnew]{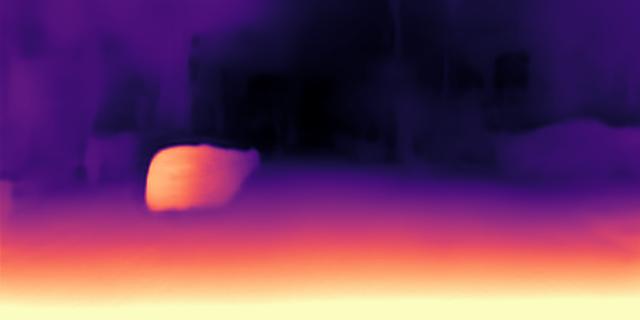} &
    \includegraphics[width=\turnheightnew]{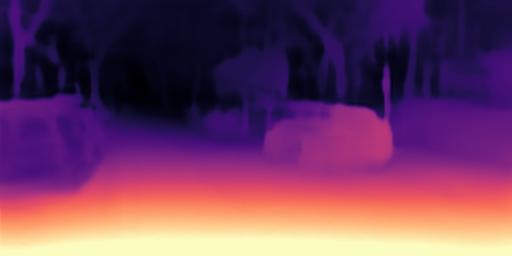} &
    \includegraphics[width=\turnheightnew]{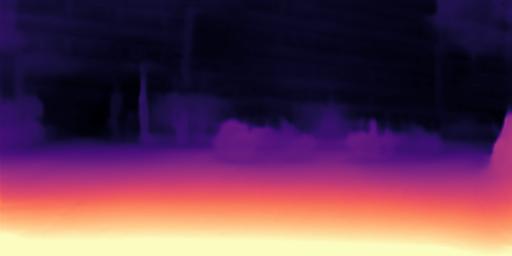} \\

    \rotatebox{90}{\hspace{1mm}\scriptsize{G:~ADDS}}&
    \includegraphics[width=\turnheightnew]{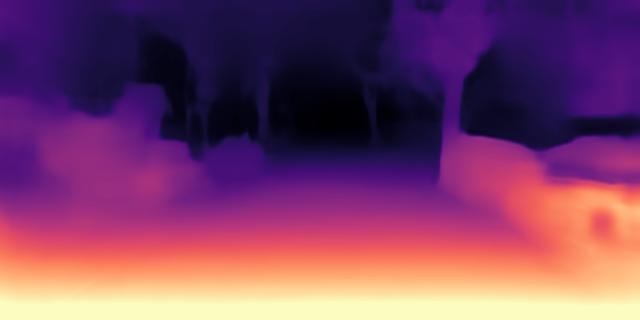} &
    \includegraphics[width=\turnheightnew]{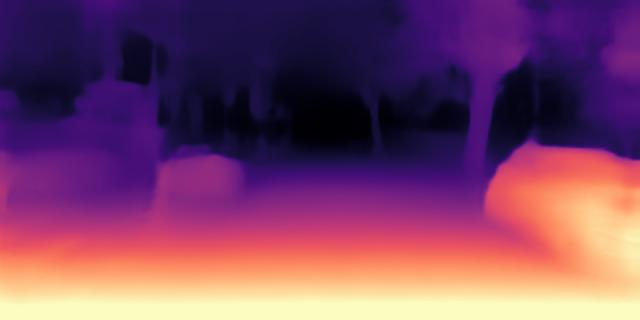} &
    \includegraphics[width=\turnheightnew]{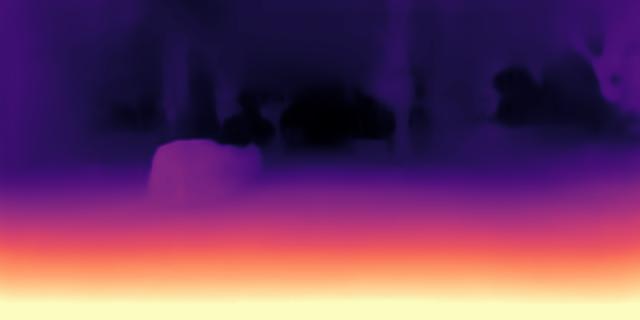} &
    \includegraphics[width=\turnheightnew]{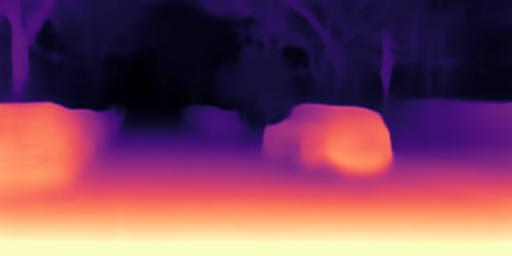} &
    \includegraphics[width=\turnheightnew]{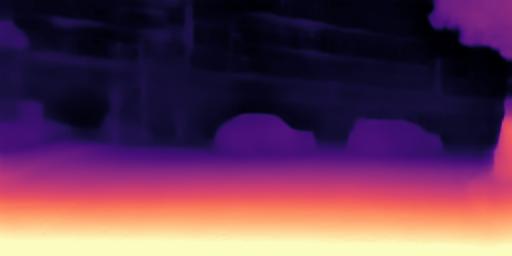} \\

    \rotatebox{90}{\hspace{-1.5mm}\scriptsize{G:~MonoFormer}}&
    \includegraphics[width=\turnheightnew]{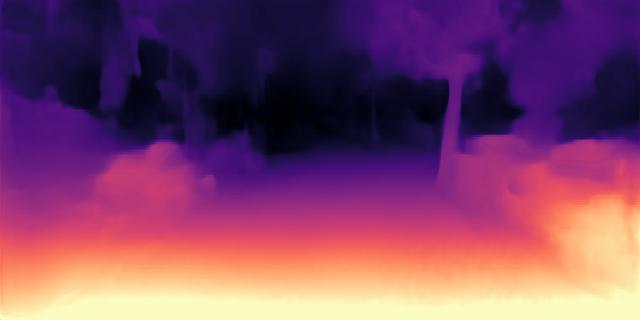} &
    \includegraphics[width=\turnheightnew]{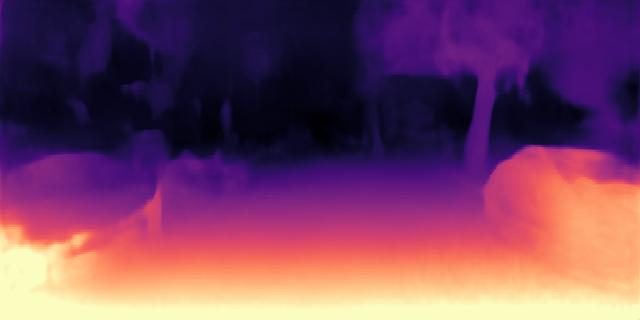} &
    \includegraphics[width=\turnheightnew]{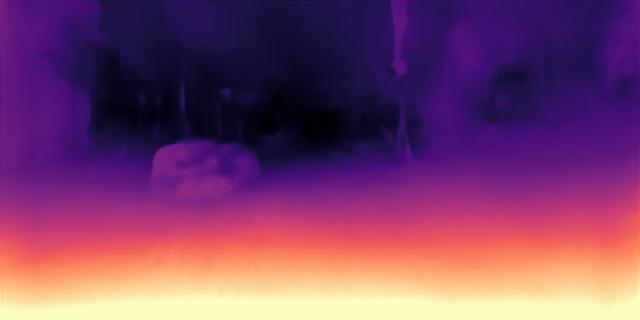} &
    \includegraphics[width=\turnheightnew]{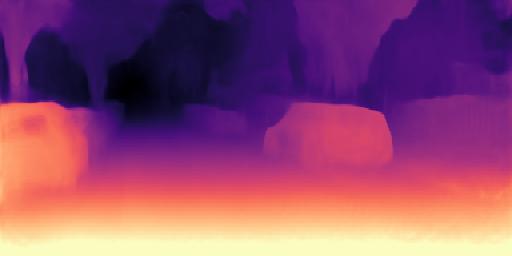} &
    \includegraphics[width=\turnheightnew]{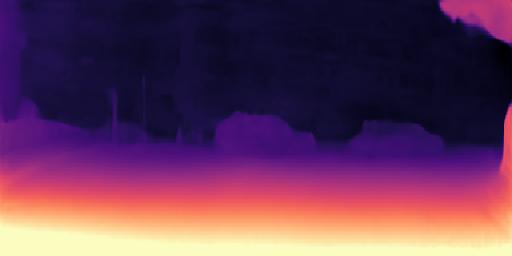} \\

    \rotatebox{90}{\hspace{-1.5mm}\scriptsize{G:~MonoViT}}&
    \includegraphics[width=\turnheightnew]{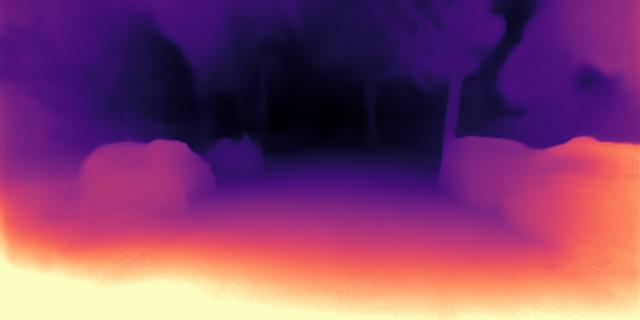} &
    \includegraphics[width=\turnheightnew]{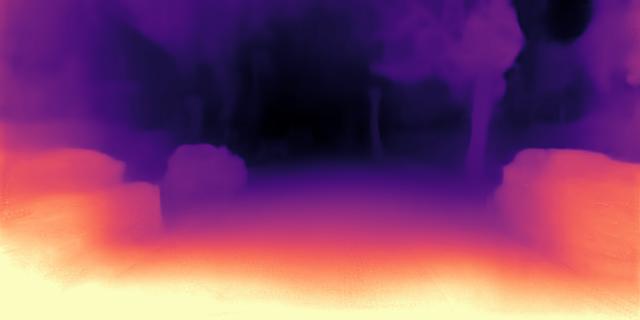} &
    \includegraphics[width=\turnheightnew]{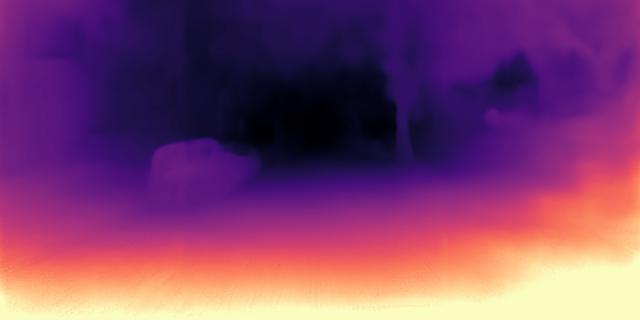} &
    \includegraphics[width=\turnheightnew]{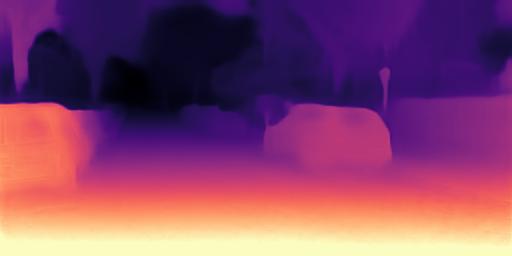} &
    \includegraphics[width=\turnheightnew]{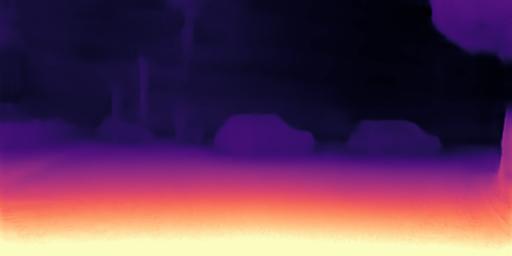} \\

    \rotatebox{90}{\hspace{4mm}\scriptsize{G:~Ours}}&
    \includegraphics[width=\turnheightnew]{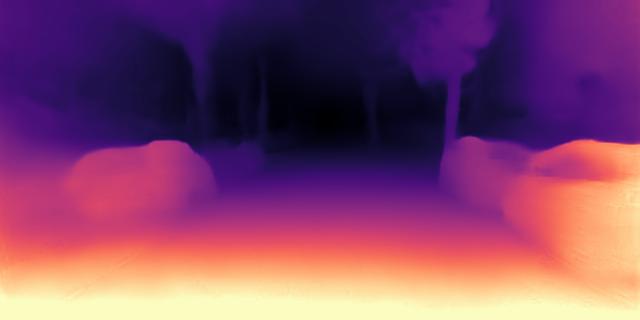} &
    \includegraphics[width=\turnheightnew]{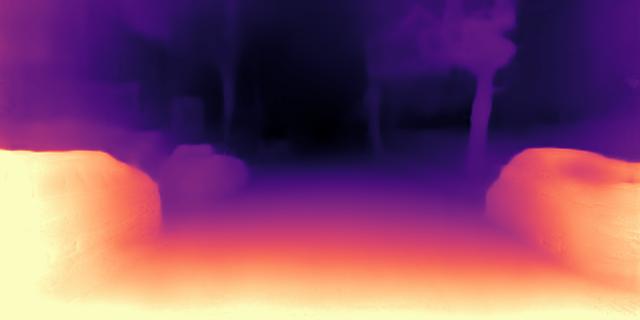} &
    \includegraphics[width=\turnheightnew]{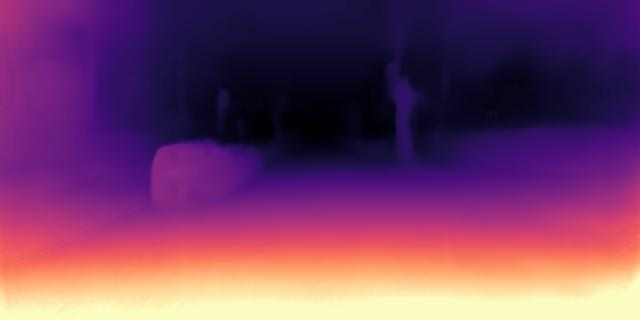} &
    \includegraphics[width=\turnheightnew]{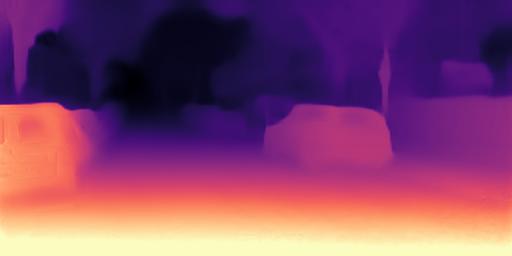} &
    \includegraphics[width=\turnheightnew]{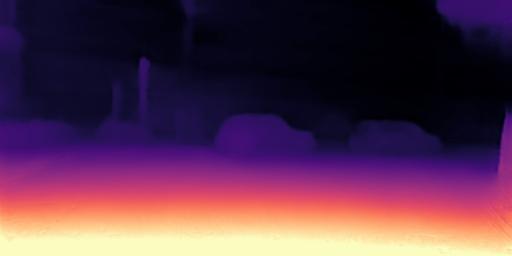} \\

    \end{tabular}
    \caption{\textbf{Qualitative results on RobotCar-Night and RobotCar-Day~\protect\cite{adds,robotcar}.}} 
    \label{fig:rnd}
\end{figure*}

We provide a complete quantitative comparison in this part. 
The visual results tested on nuScenes-Night and nuScenes-Day are shown in Fig.~\ref{fig:nnd} and those on RobotCar-Night and RobotCar-Day are shown in Fig.~\ref{fig:rnd}. 
\subsection{More Visual Results on Other Nighttime Datasets}
\begin{figure*}[htbp]
    \centering
    \newcommand{\turnheightnew}{0.35\columnwidth}
    \begin{tabular}{@{\hskip 0mm}c@{\hskip 1mm}c@{\hskip 1mm}c@{\hskip 1mm}c@{\hskip 1mm}c@{\hskip 1mm}c@{}}
    \vspace{-0.5mm}
    {\rotatebox{90}{\hspace{2mm}\scriptsize{darkZurich}}} &
    \includegraphics[width=\turnheightnew]{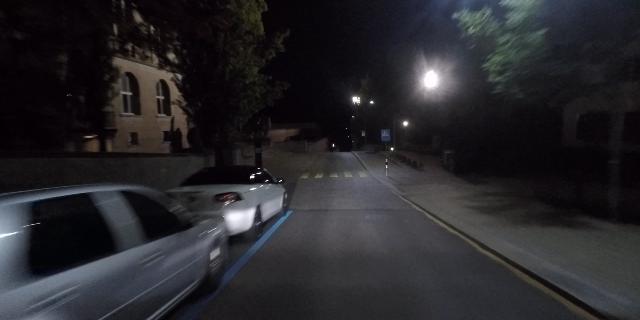} &
    \includegraphics[width=\turnheightnew]{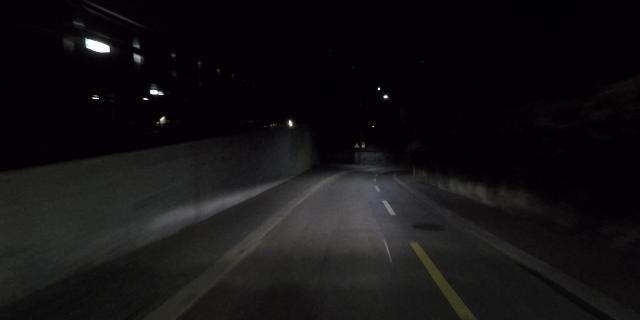} &
    \includegraphics[width=\turnheightnew]{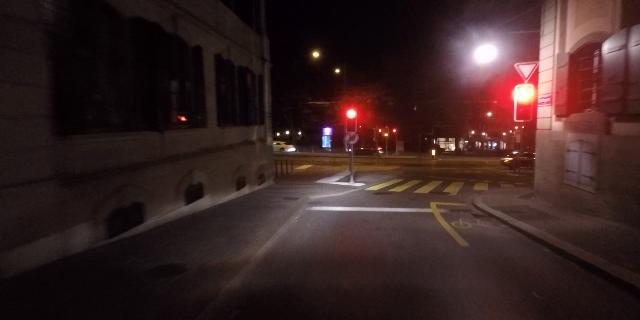}&
    \includegraphics[width=\turnheightnew]{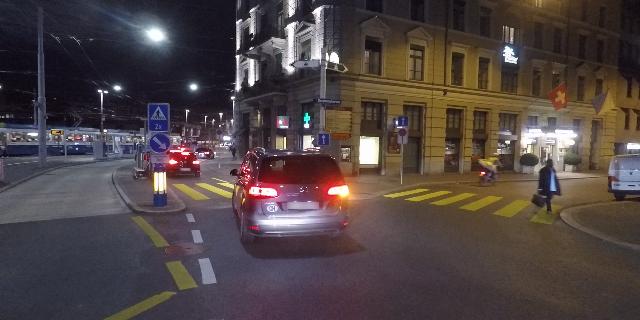}&
    \includegraphics[width=\turnheightnew]{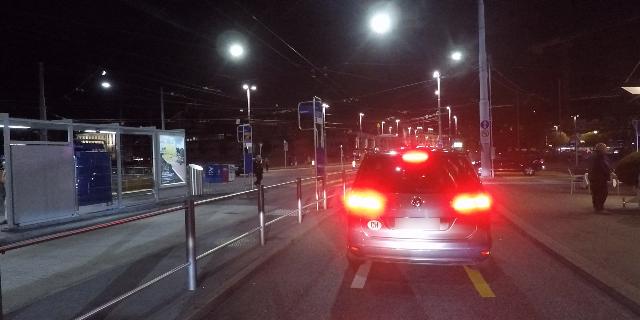}\\
    
    \vspace{0mm}
    \rotatebox{90}{\hspace{0mm}\scriptsize{G: MonoFormer}}&
    \includegraphics[width=\turnheightnew]{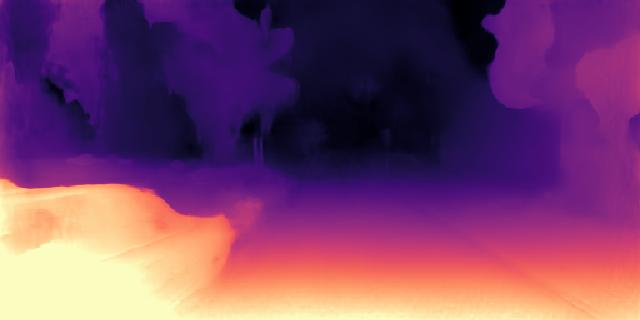} &
    \includegraphics[width=\turnheightnew]{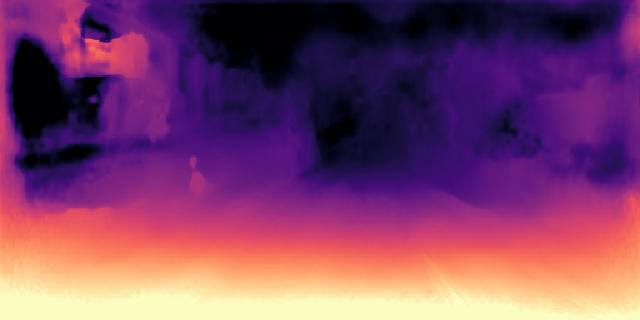} &
    \includegraphics[width=\turnheightnew]{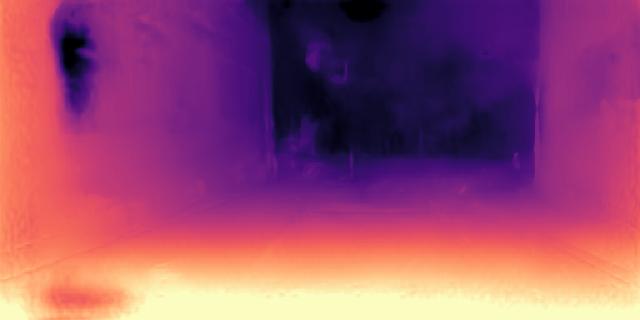} &
    \includegraphics[width=\turnheightnew]{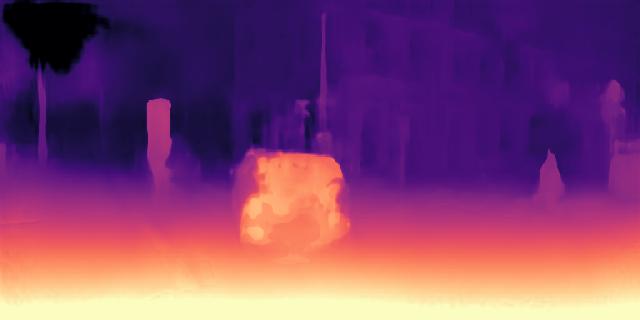} &
    \includegraphics[width=\turnheightnew]{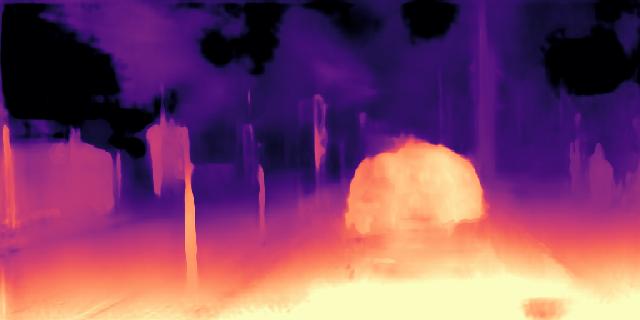} \\
    
    \vspace{0mm}
    \rotatebox{90}{\hspace{4mm}\scriptsize{G: Ours}}&
    \includegraphics[width=\turnheightnew]{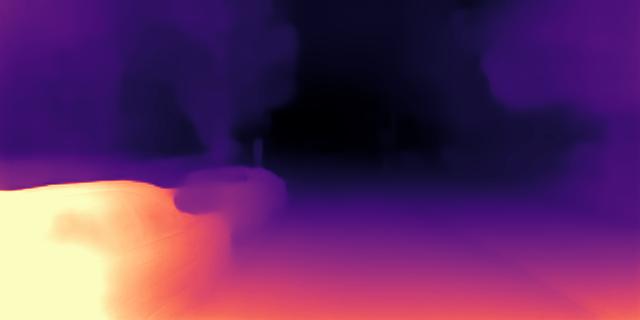} &
    \includegraphics[width=\turnheightnew]{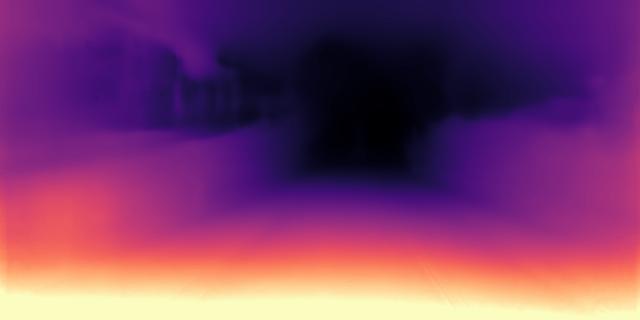} &
    \includegraphics[width=\turnheightnew]{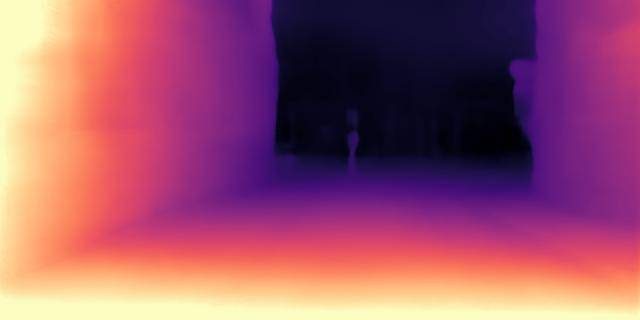} &
    \includegraphics[width=\turnheightnew]{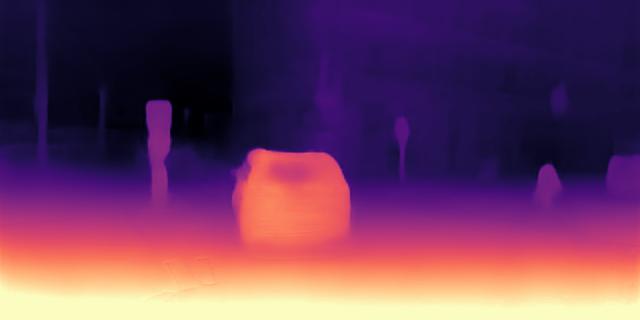} &
    \includegraphics[width=\turnheightnew]{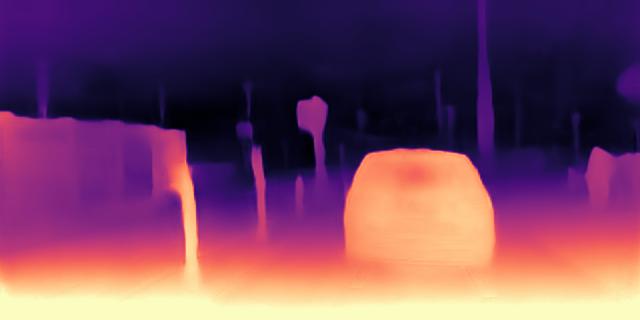} \\

    \vspace{-0.5mm}
    {\rotatebox{90}{\hspace{2mm}\scriptsize{darkZurich}}} &
    \includegraphics[width=\turnheightnew]{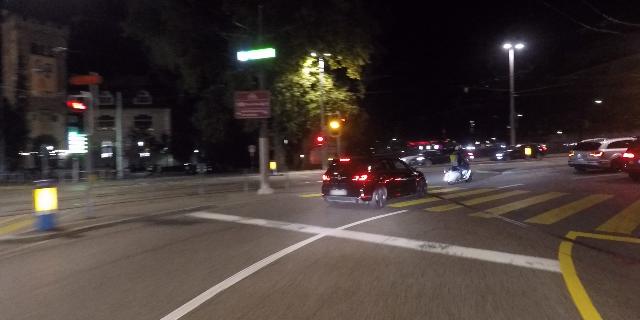} &
    \includegraphics[width=\turnheightnew]{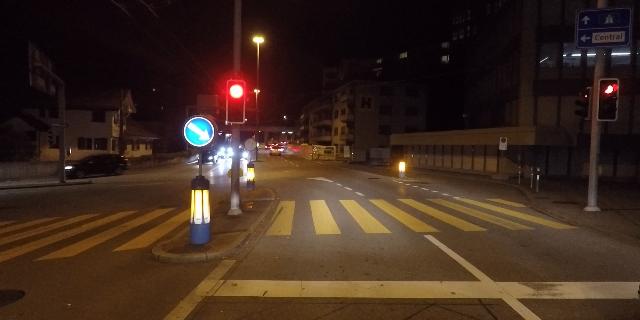} &
    \includegraphics[width=\turnheightnew]{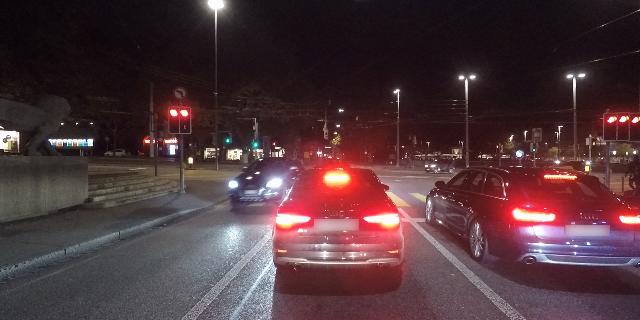} &
    \includegraphics[width=\turnheightnew]{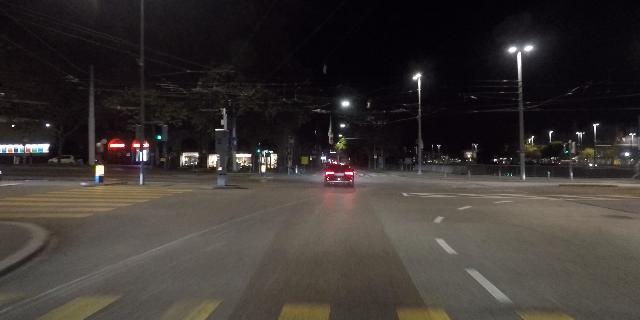} &
    \includegraphics[width=\turnheightnew]{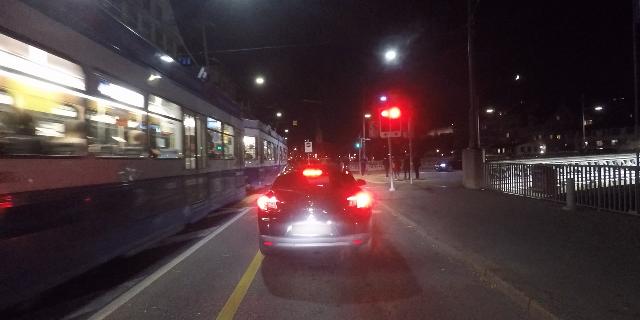} \\
    
    \vspace{0mm}
    \rotatebox{90}{\hspace{0mm}\scriptsize{G: MonoFormer}}&
    \includegraphics[width=\turnheightnew]{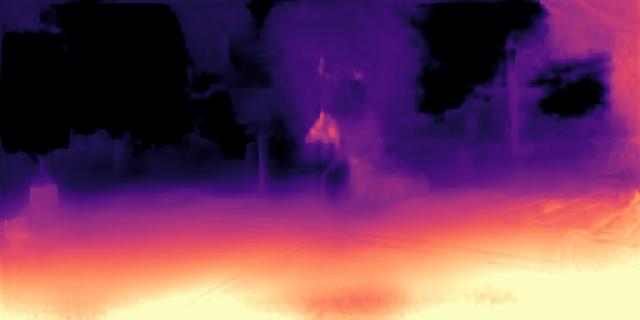} &
    \includegraphics[width=\turnheightnew]{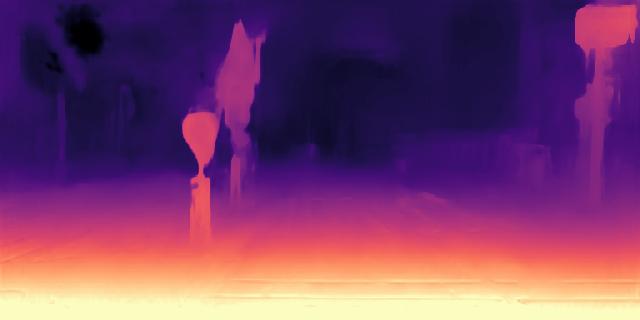} &
    \includegraphics[width=\turnheightnew]{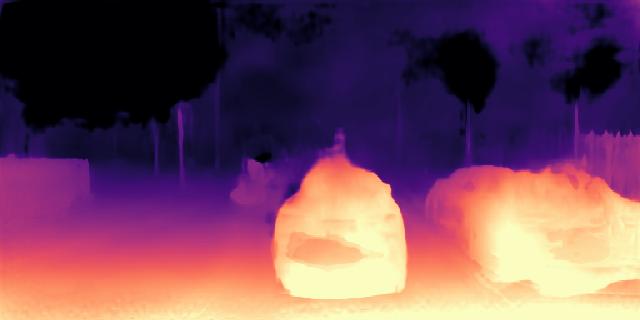} &
    \includegraphics[width=\turnheightnew]{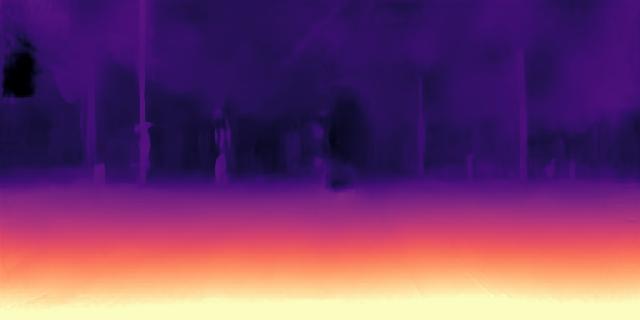} &
    \includegraphics[width=\turnheightnew]{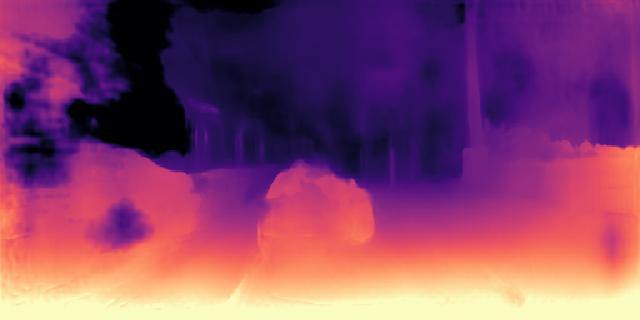} \\
    
    \vspace{0mm}
    \rotatebox{90}{\hspace{4mm}\scriptsize{G: Ours}}&
    \includegraphics[width=\turnheightnew]{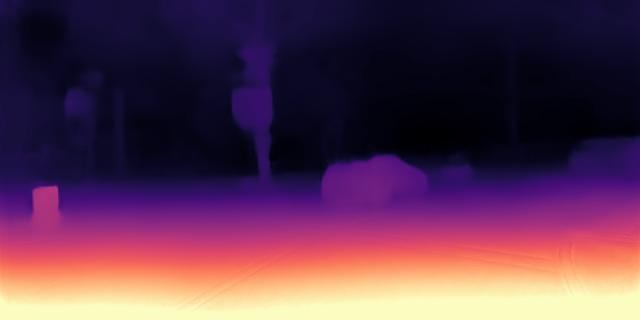} &
    \includegraphics[width=\turnheightnew]{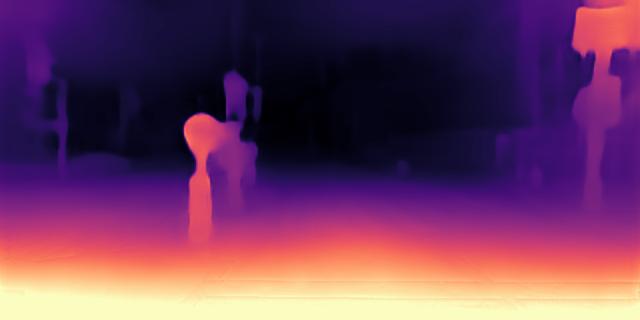} &
    \includegraphics[width=\turnheightnew]{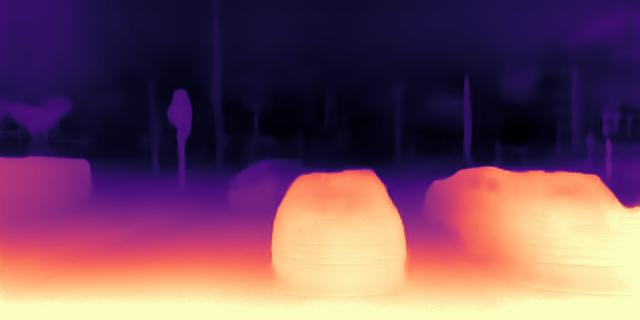} &
    \includegraphics[width=\turnheightnew]{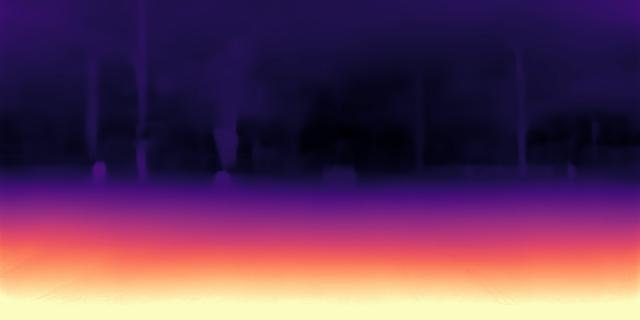} &
    \includegraphics[width=\turnheightnew]{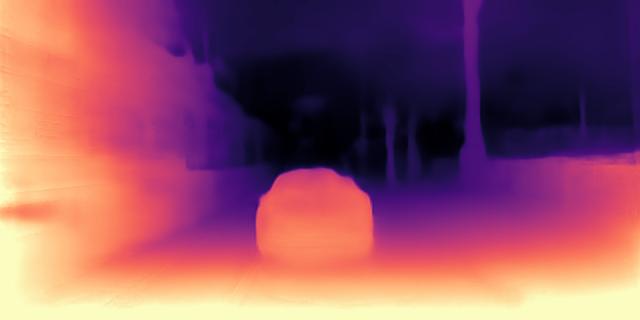} \\
    \end{tabular}
    \caption{\textbf{Qualitative results on darkZurich~\protect\cite{darkZurich}.} 
    Note that G:~MonoFormer applies a larger Transformer-CNN hybrid backbone for the depth network~(about $ \times $12 of parameters).
    } 
    \label{fig:dz}
\end{figure*}

\begin{figure*}[htbp]
    \centering
    \newcommand{\turnheightnew}{0.35\columnwidth}
    \begin{tabular}{@{\hskip 0mm}c@{\hskip 1mm}c@{\hskip 1mm}c@{\hskip 1mm}c@{\hskip 1mm}c@{\hskip 1mm}c@{}}
    \vspace{-0.5mm}
    {\rotatebox{90}{\hspace{2mm}\scriptsize{NightCity}}} &
    \includegraphics[width=\turnheightnew]{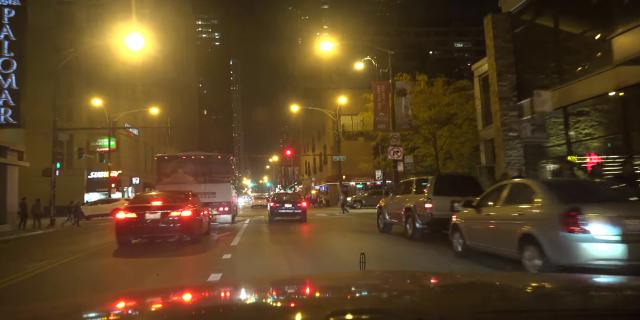} &
    \includegraphics[width=\turnheightnew]{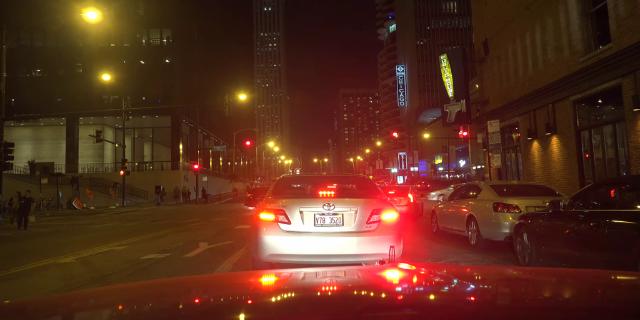} &
    \includegraphics[width=\turnheightnew]{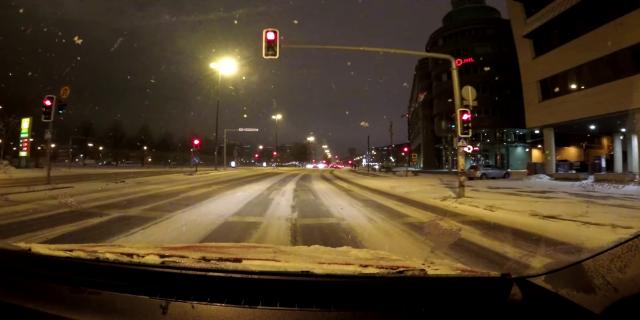} &
    \includegraphics[width=\turnheightnew]{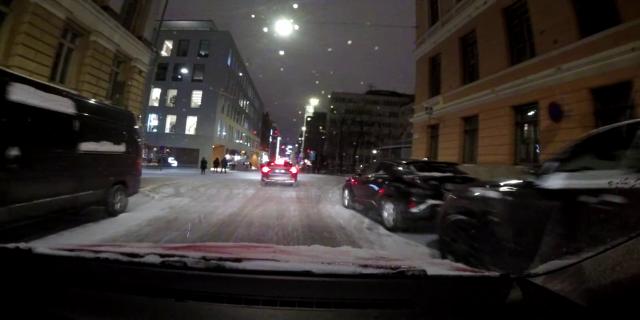} &
    \includegraphics[width=\turnheightnew]{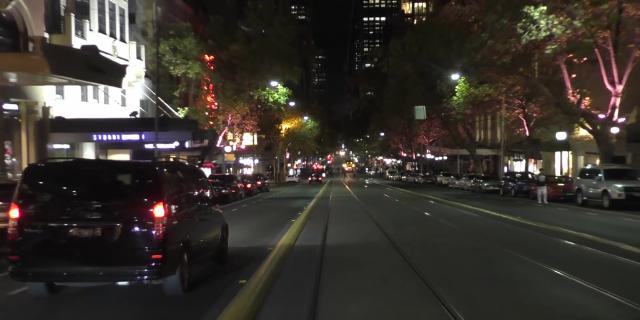} \\
    
    \vspace{0.5mm}
    \rotatebox{90}{\hspace{0mm}\scriptsize{G: MonoFormer}}&
    \includegraphics[width=\turnheightnew]{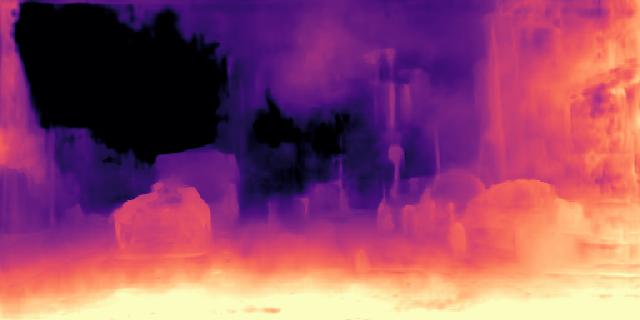} &
    \includegraphics[width=\turnheightnew]{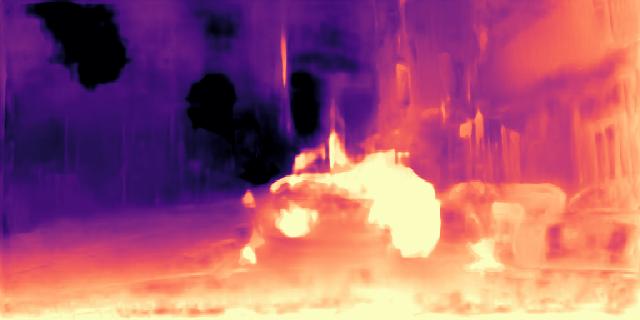} &
    \includegraphics[width=\turnheightnew]{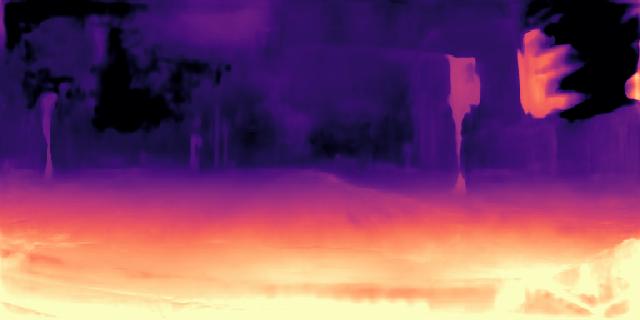} &
    \includegraphics[width=\turnheightnew]{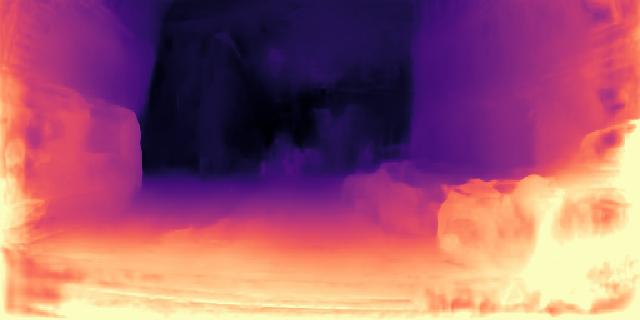} &
    \includegraphics[width=\turnheightnew]{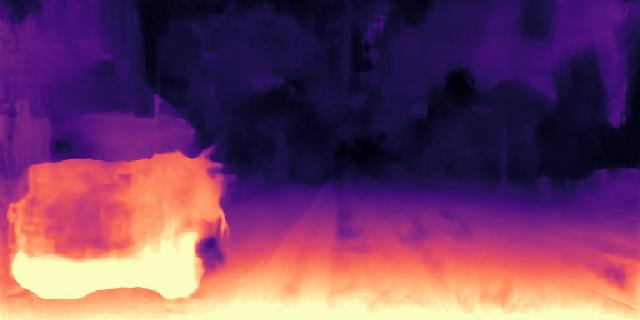} \\
    
    \vspace{0.5mm}
    \rotatebox{90}{\hspace{4mm}\scriptsize{G: Ours}}&
    \includegraphics[width=\turnheightnew]{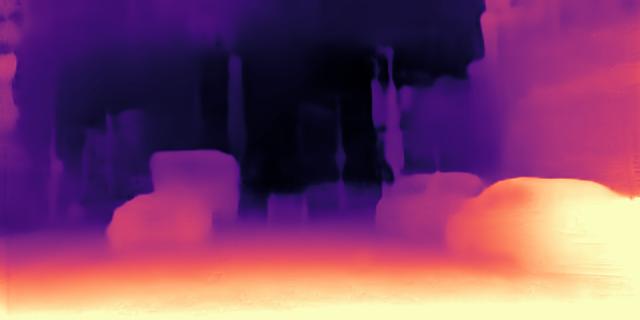} &
    \includegraphics[width=\turnheightnew]{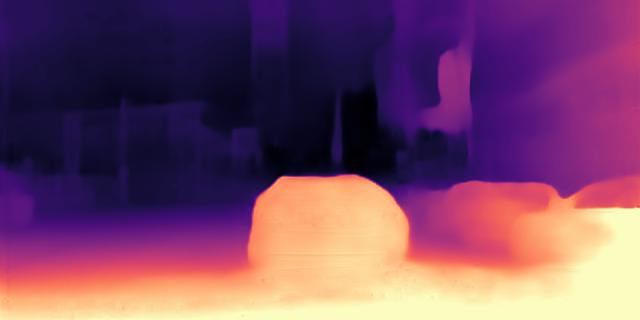} &
    \includegraphics[width=\turnheightnew]{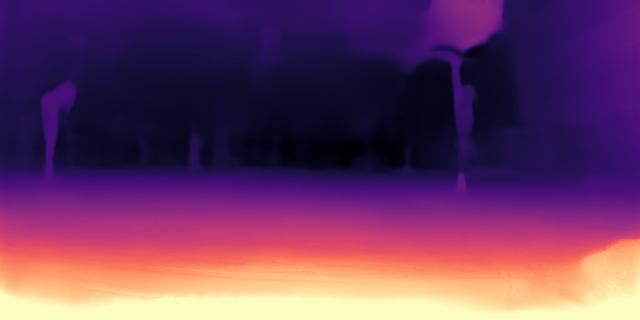} &
    \includegraphics[width=\turnheightnew]{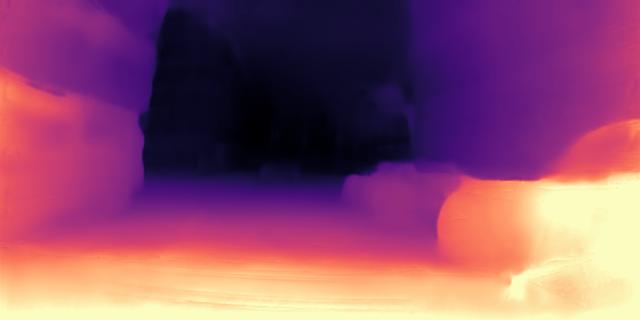} &
    \includegraphics[width=\turnheightnew]{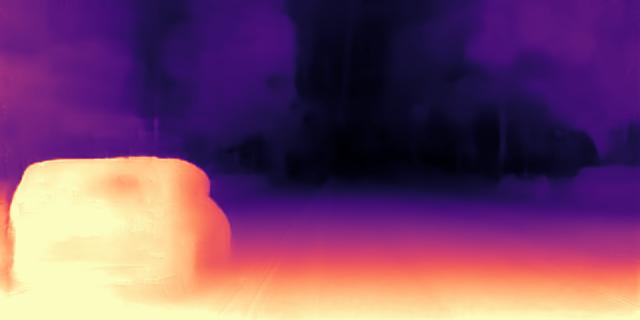} \\

    \vspace{-0.5mm}
    {\rotatebox{90}{\hspace{2mm}\scriptsize{NightCity}}} &
    \includegraphics[width=\turnheightnew]{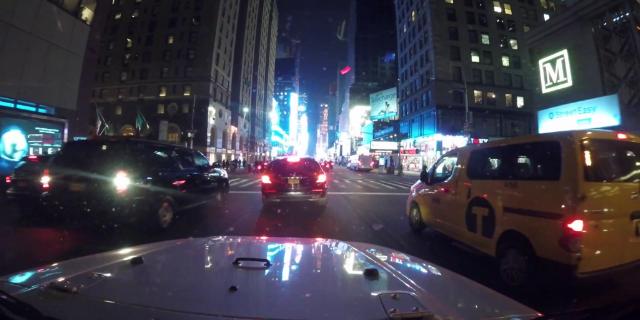} &
    \includegraphics[width=\turnheightnew]{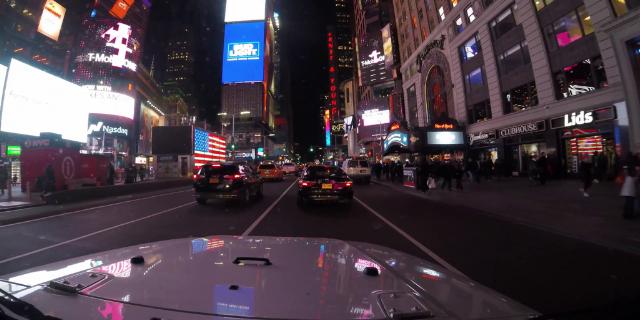} &
    \includegraphics[width=\turnheightnew]{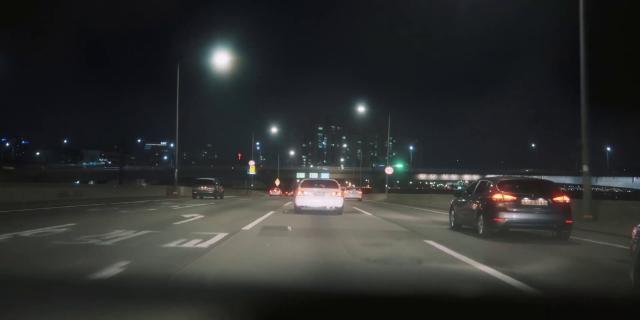} &
    \includegraphics[width=\turnheightnew]{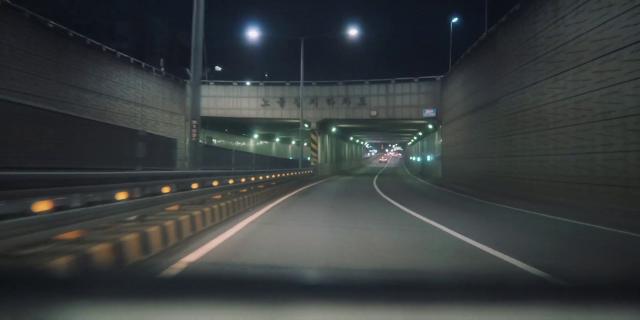} &
    \includegraphics[width=\turnheightnew]{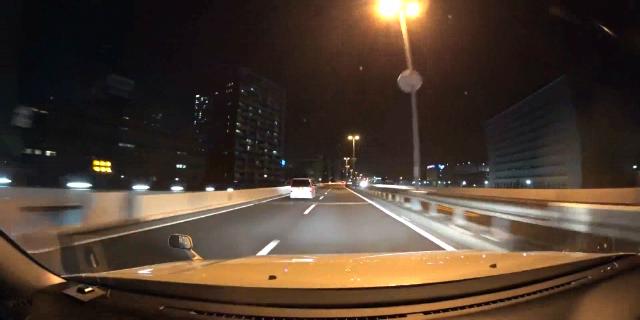} \\
    
    \vspace{0.5mm}
    \rotatebox{90}{\hspace{0mm}\scriptsize{G: MonoFormer}}&
    \includegraphics[width=\turnheightnew]{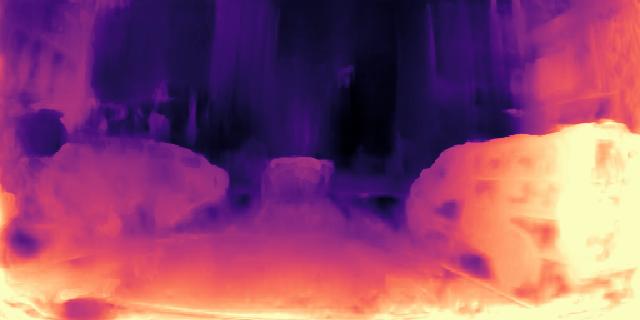} &
    \includegraphics[width=\turnheightnew]{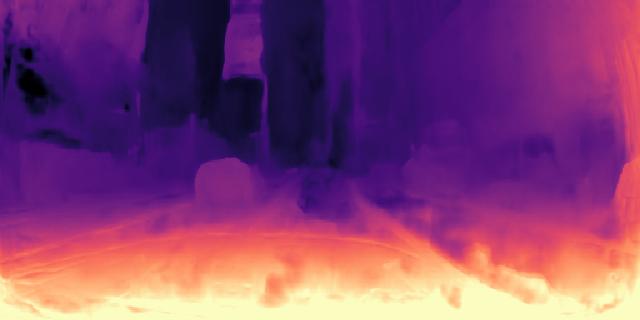} &
    \includegraphics[width=\turnheightnew]{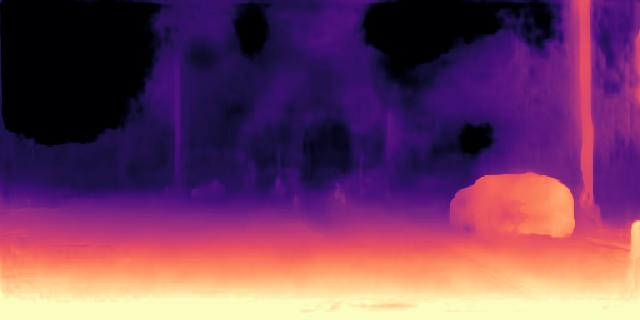} &
    \includegraphics[width=\turnheightnew]{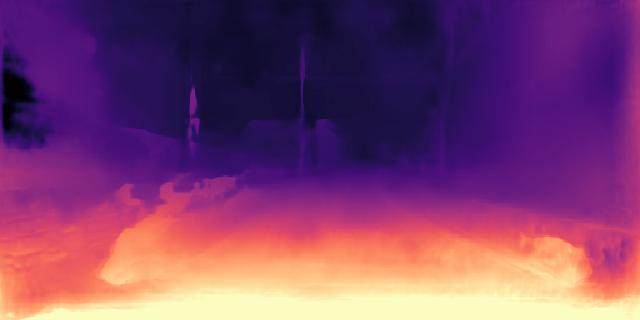} &
    \includegraphics[width=\turnheightnew]{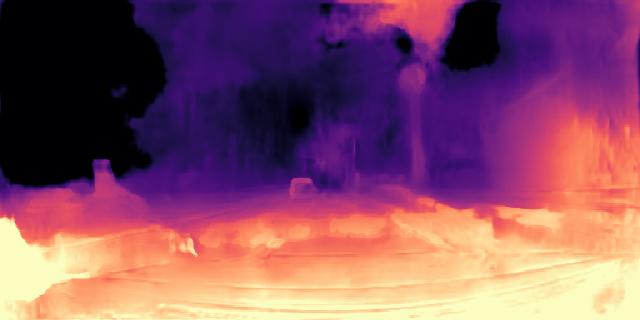} \\
    
    \vspace{0.5mm}
    \rotatebox{90}{\hspace{4mm}\scriptsize{G: Ours}}&
    \includegraphics[width=\turnheightnew]{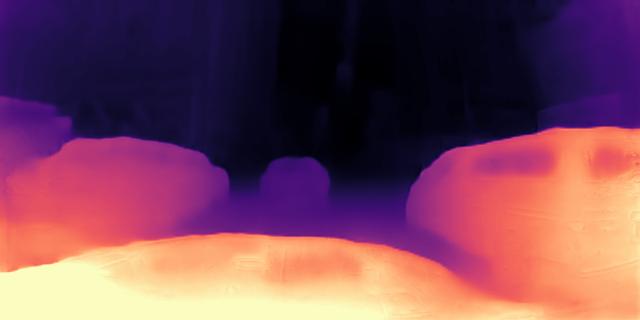} &
    \includegraphics[width=\turnheightnew]{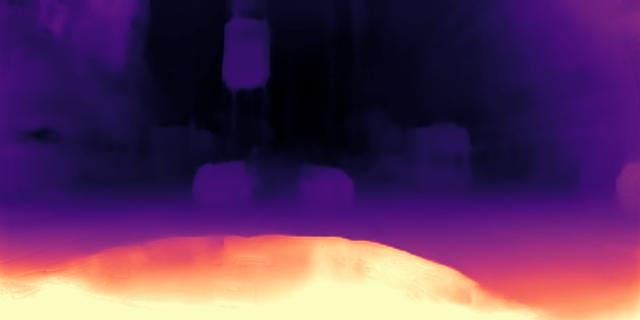} &
    \includegraphics[width=\turnheightnew]{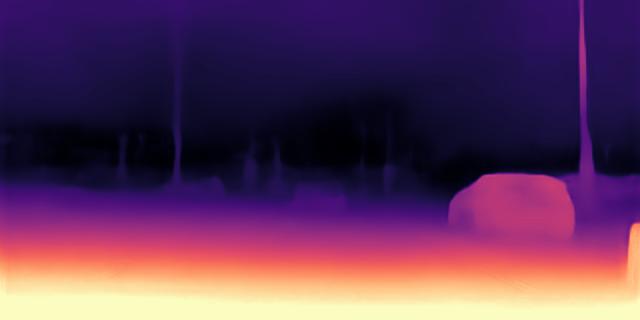} &
    \includegraphics[width=\turnheightnew]{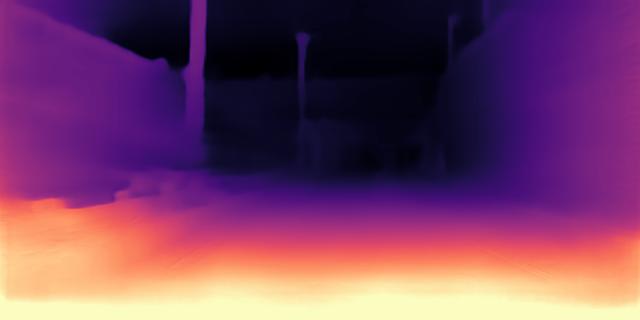} &
    \includegraphics[width=\turnheightnew]{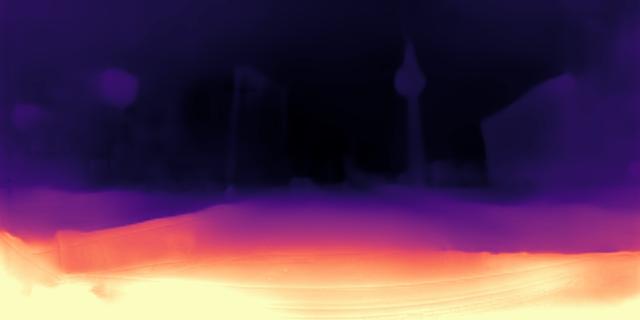} \\
    \end{tabular}
    \caption{\textbf{Qualitative results on NightCity~\protect\cite{NightCity}.}
    Compared to darkZurich, the NightCity dataset is more challenging because it consists of images taken by camares with different hardware settings~(e.g., intrinsic, distortion factor, extrinsic relative to the vehicle , etc).
    } 
    \label{fig:nc}
\end{figure*}

To further present the effectiveness of our method on unseen nighttime scenes, we show the visual results on darkZurich (Fig.~\ref{fig:dz}) and NightCity (Fig.~\ref{fig:nc}), correspondingly.
Note that darkZurich and NightCity are two popular used nighttime segmentation datasets. 
As shown in Fig.~\ref{fig:nc}, the input images have different camera hardware settings which makes it a more challenging setting for testing.
While our presented self-supervised method still maintains reasonable performance and outperforms G: MonoFormer~\cite{monoformer} though it utilizes a much larger DepthNet backbone.

\subsection{More visualization for ablation analyses}
\begin{figure}[htbp]
    \vspace{-2mm}
    \centering
    \newcommand{\turnheightnew}{0.21\columnwidth}
    \begin{tabular}{@{\hskip 0mm}c@{\hskip 1mm}c@{\hskip 1mm}c@{\hskip 1mm}|@{\hskip 1mm} c@{\hskip 1mm}c@{}}
    \vspace{-1mm}
    {\rotatebox{90}{\hspace{2mm}\scriptsize{Input}}} &
    \includegraphics[width=\turnheightnew]{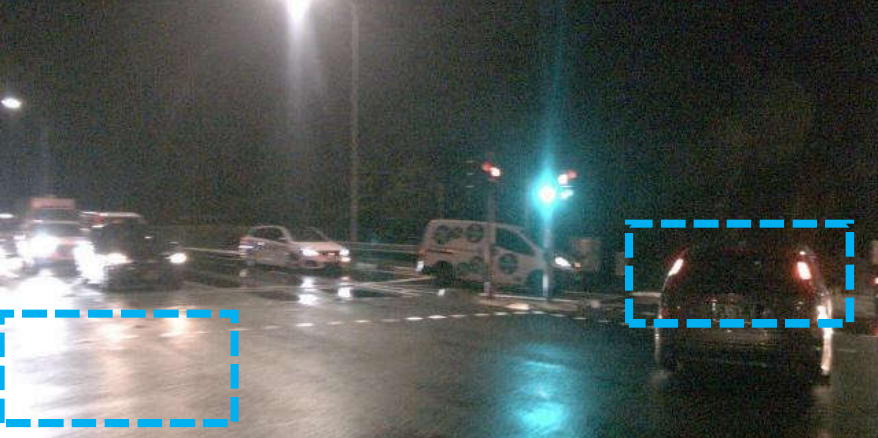} &
    \includegraphics[width=\turnheightnew]{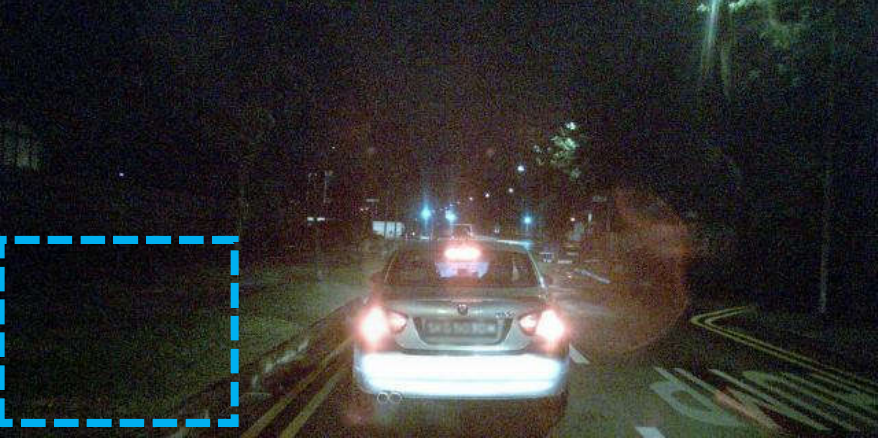} &
    \includegraphics[width=\turnheightnew]{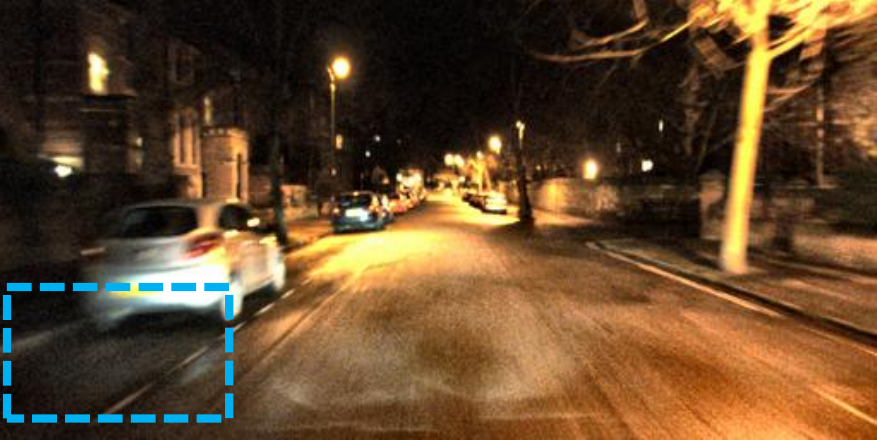} &
    \includegraphics[width=\turnheightnew]{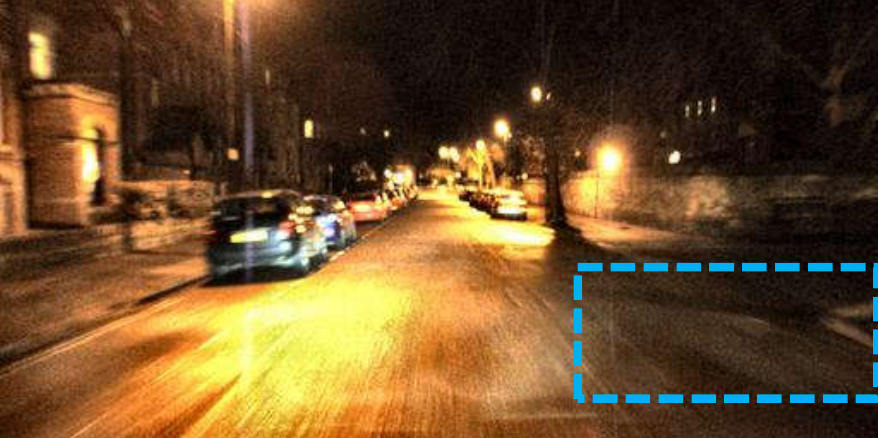} \\
    % \vspace{1mm}

    \vspace{-0.5mm}
    
    \rotatebox{90}{\hspace{0mm}\scriptsize{Baseline}}&
    \includegraphics[width=\turnheightnew]{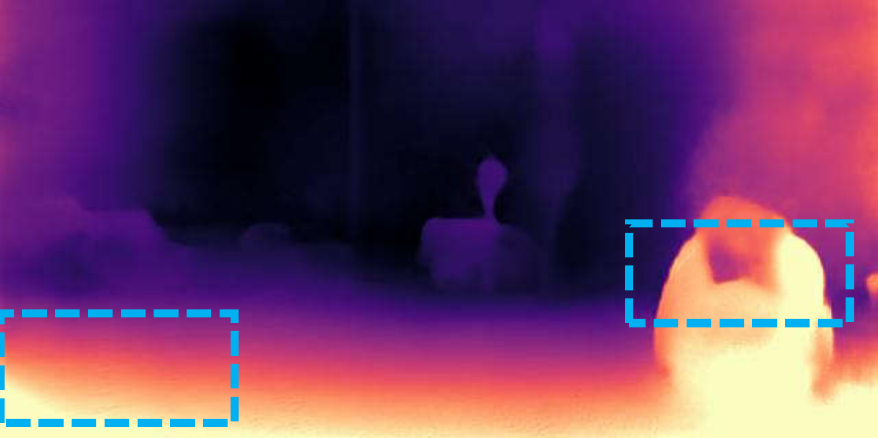} &
    \includegraphics[width=\turnheightnew]{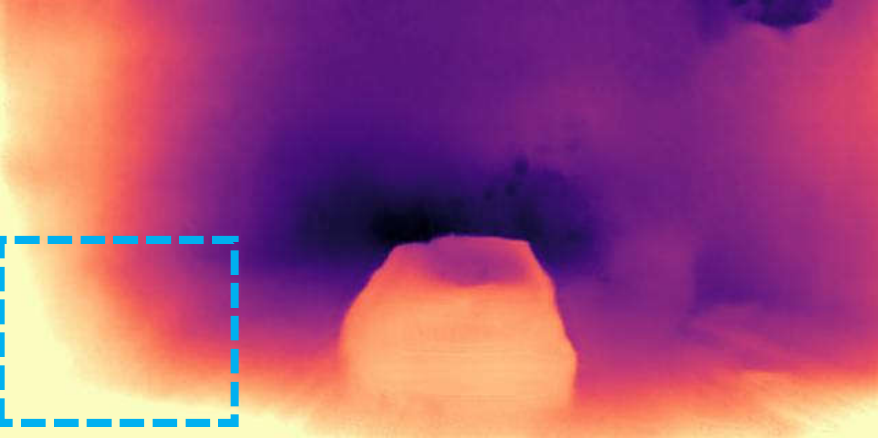} &
    \includegraphics[width=\turnheightnew]{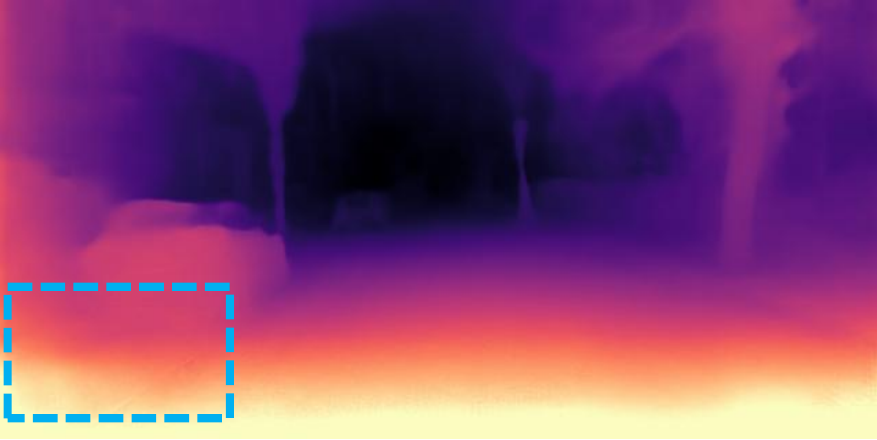} &
    \includegraphics[width=\turnheightnew]{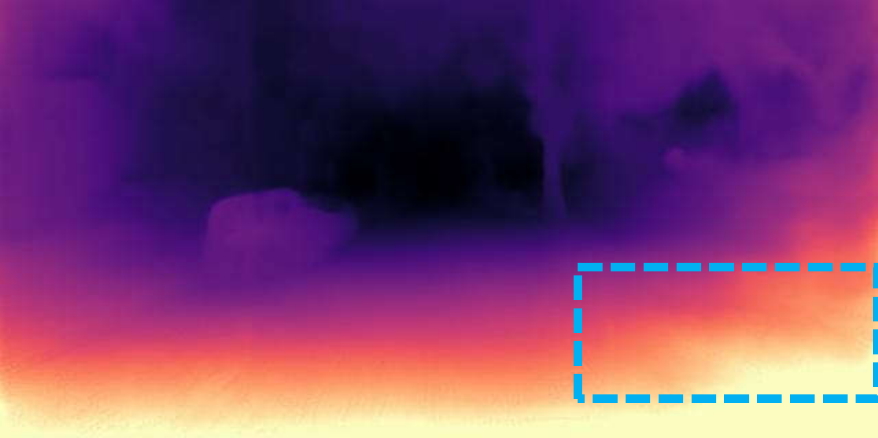} \\

    \vspace{-0.5mm}
    
    {\rotatebox{90}{\hspace{2mm}\scriptsize{BPG}}}&
    \includegraphics[width=\turnheightnew]{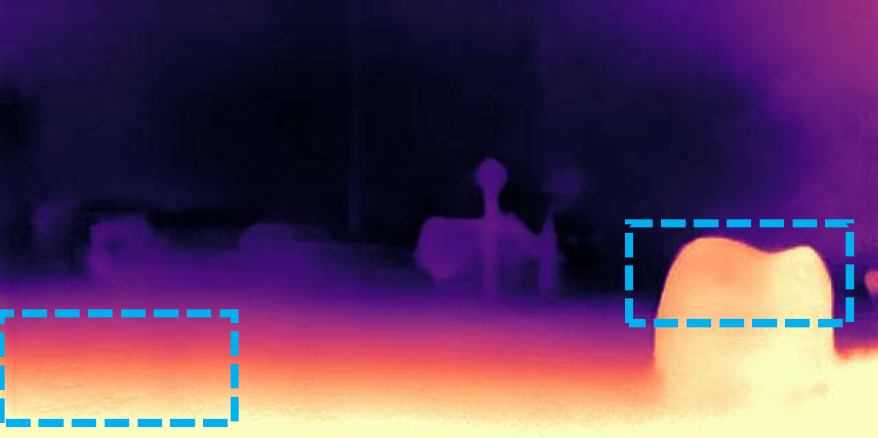} &
    \includegraphics[width=\turnheightnew]{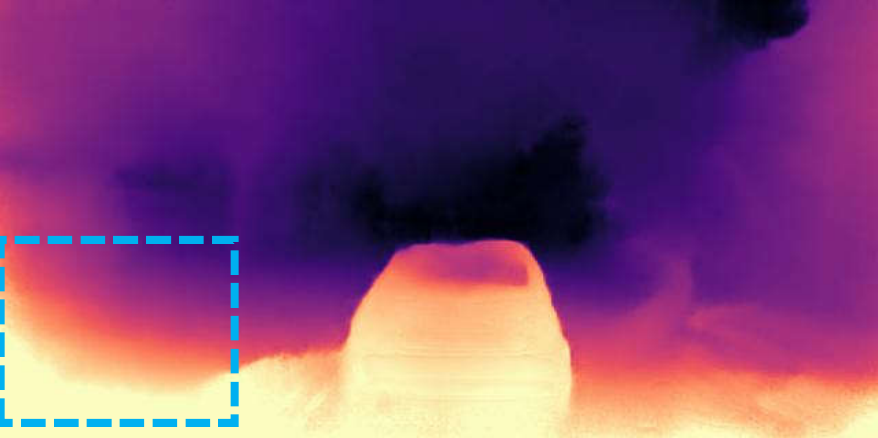} &
    \includegraphics[width=\turnheightnew]{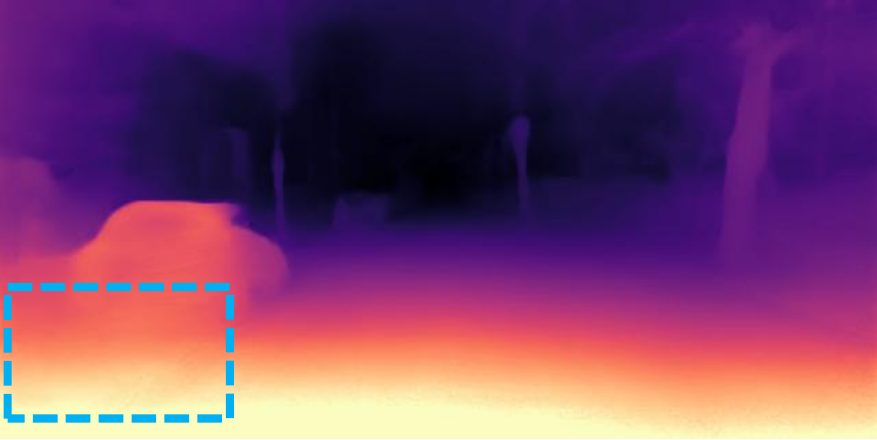} &
    \includegraphics[width=\turnheightnew]{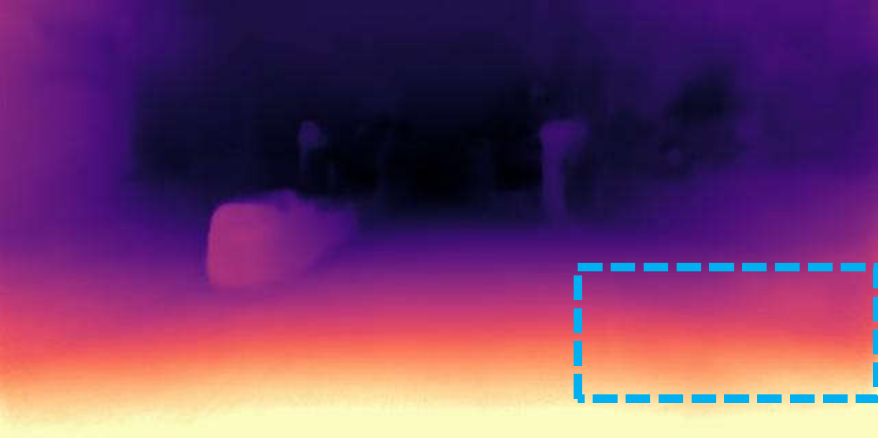} \\

    \vspace{-0.5mm}
    
    \rotatebox{90}{\hspace{2mm}\scriptsize{ING}}&
    \includegraphics[width=\turnheightnew]{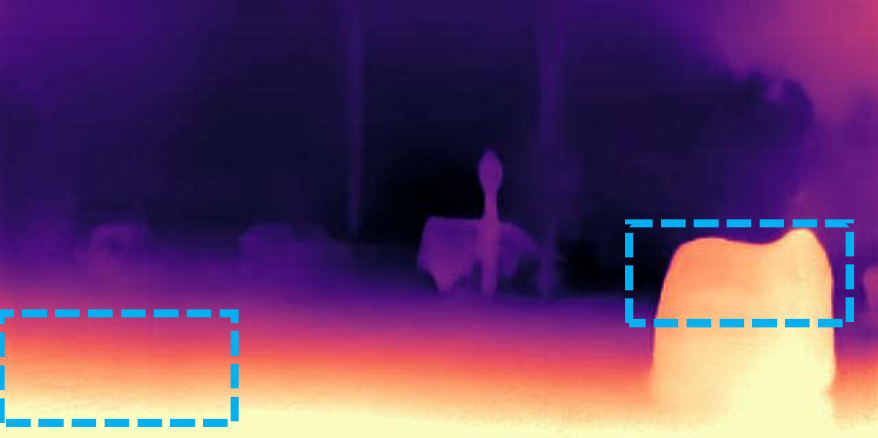} &
    \includegraphics[width=\turnheightnew]{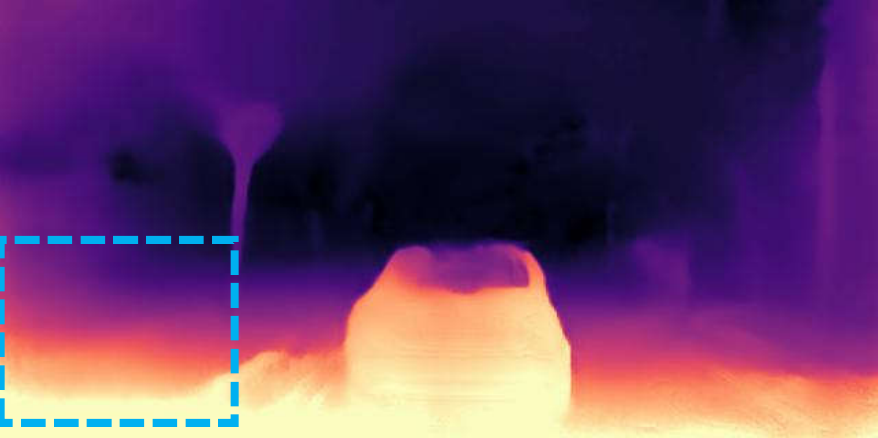} &
    \includegraphics[width=\turnheightnew]{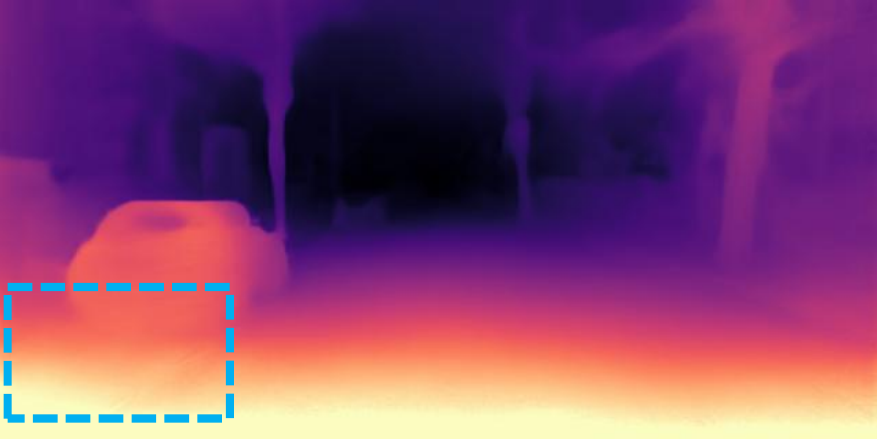} &
    \includegraphics[width=\turnheightnew]{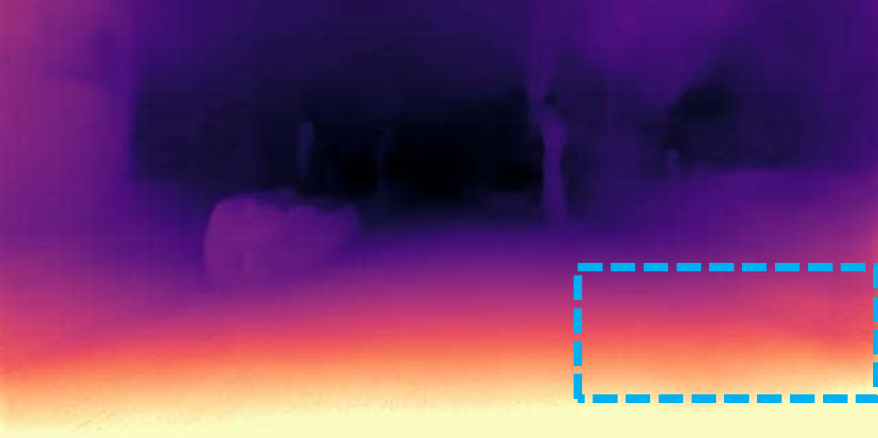} \\

    % \vspace{-0.5mm}
    
    \rotatebox{90}{\hspace{2mm}\scriptsize{Full}}&
    \includegraphics[width=\turnheightnew]{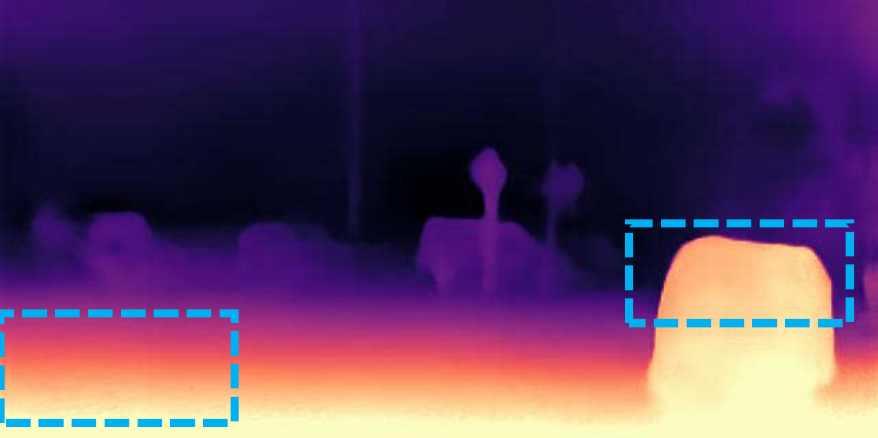} &
    \includegraphics[width=\turnheightnew]{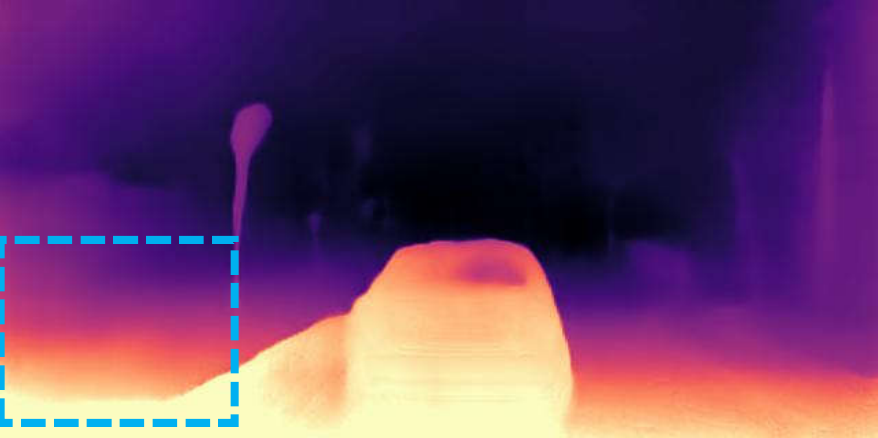} &
    \includegraphics[width=\turnheightnew]{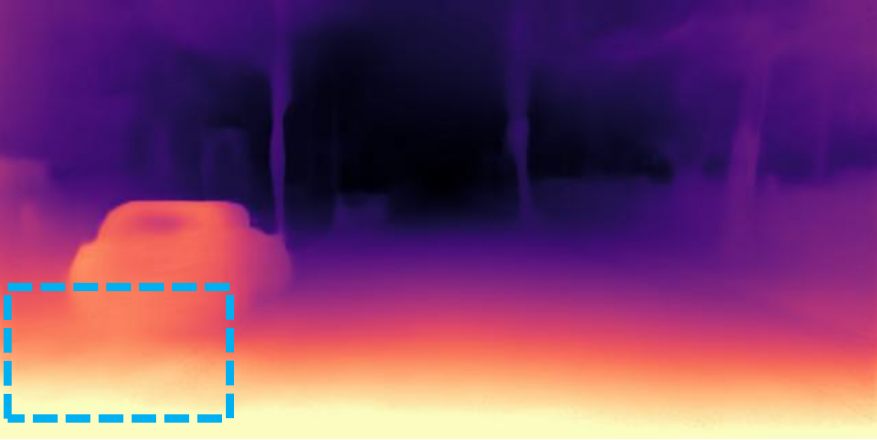} &
    \includegraphics[width=\turnheightnew]{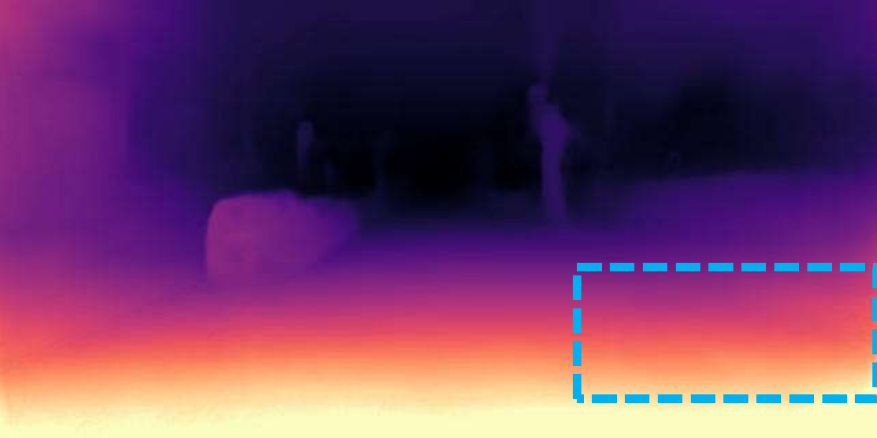} \\

    \end{tabular}
    \vspace{0.5mm}
    \caption{\textbf{Ablation test.} Left: test on nuScenes-Night. Right: test on RobotCar-Night. 
    Baseline, BPG, ING, and Full mean using nothing, Brightness Peak Generator, Image Noise Generator and BPG+ING for data compensation, respectively.} 
    \label{fig:arnd}
\end{figure}

In Fig~\ref{fig:arnd}, the BPG helps the regions where highlight spots frequently occur, such as the cars and  the road, as shown in the cyan dashed boxes. The ING simulates the potential imaging noises in the dark area, thus present better depth prediction at the background, such as the sky. The full model receives the merits of BPG and ING, and thus has the best prediction performance at the mentioned areas. 
More quantitative study is in Sec.~4.4 in the main paper.

\clearpage

\clearpage
% \bibliographystyle{named}
% \bibliography{ijcai24}

\end{document}